\documentclass[11pt,letterpaper]{article}
\usepackage[letterpaper,hmargin=0.875in,vmargin=1in]{geometry}
\usepackage{microtype}
\usepackage{graphicx}
\usepackage{booktabs}
\usepackage{amsmath}
\usepackage{amssymb}
\usepackage{mathtools}
\usepackage{amsthm}
\usepackage{enumitem}
\usepackage{float}
\usepackage{placeins}
\usepackage{algorithm}
\usepackage{algorithmic}
\usepackage[round,authoryear]{natbib}
\usepackage[hidelinks]{hyperref}
\usepackage[capitalize,noabbrev]{cleveref}
\hypersetup{pdftitle={Do Language Models Know Which Constraints Belong Together?},pdfauthor={Junbo Jacob Lian, Kaixuan Chen, Hanzhang Qin, Huiling Chen, Chaoyu Zhang, Chung-Piaw Teo},pdfsubject={}}

\theoremstyle{plain}
\newtheorem{theorem}{Theorem}
\newtheorem{proposition}{Proposition}
\newtheorem{lemma}{Lemma}

\theoremstyle{definition}
\newtheorem{definition}{Definition}
\newtheorem{example}{Example}
\theoremstyle{remark}

\newcommand{\conv}{\operatorname{conv}}
\newcommand{\LP}{\mathrm{LP}}
\newcommand{\IP}{\mathrm{IP}}
\newcommand{\cT}{\mathcal{T}}
\newcommand{\cov}{\mathrm{C}}

\newcommand{\authorcell}[2]{\begin{minipage}[t]{0.31\textwidth}\centering{\large #1}\\[1pt]{\small\href{mailto:#2}{\nolinkurl{#2}}}\end{minipage}}

\title{Do Language Models Know Which Constraints Belong Together?\thanks{Code and data: \href{https://github.com/junbolian/JointDW}{github.com/junbolian/JointDW}.}}
\author{%
\authorcell{Junbo Jacob Lian\textsuperscript{1,2,3}}{jacoblian@u.northwestern.edu}\hspace{0.015\textwidth}%
\authorcell{Kaixuan Chen\textsuperscript{3}}{chenkaixuan77@gmail.com}\hspace{0.015\textwidth}%
\authorcell{Hanzhang Qin\textsuperscript{1}}{hzqin@nus.edu.sg}\\\noalign{\vskip 9pt}
\authorcell{Huiling Chen\textsuperscript{4}}{chenhuiling.jlu@gmail.com}\hspace{0.015\textwidth}%
\authorcell{Chaoyu Zhang\textsuperscript{5}}{cy.zhang@cityu.edu.hk}\hspace{0.015\textwidth}%
\authorcell{Chung-Piaw Teo\textsuperscript{1}}{bizteocp@nus.edu.sg}\\\noalign{\vskip 11pt}
\parbox{0.95\textwidth}{\centering\small\itshape
\textsuperscript{1}Institute of Operations Research and Analytics, National University of Singapore, Singapore\\
\textsuperscript{2}Wenzhou Buyi Pharmacy, Wenzhou, China\\
\textsuperscript{3}McCormick School of Engineering, Northwestern University, Evanston, IL 60208, USA\\
\textsuperscript{4}College of Computer Science and Artificial Intelligence, Wenzhou University, Wenzhou, China\\
\textsuperscript{5}Department of Decision Analytics and Operations, City University of Hong Kong, Hong Kong, China}}
\date{}

\begin{document}
\maketitle
\begin{abstract}
Dantzig--Wolfe reformulation strengthens the linear relaxation of a mixed-integer program by convexifying constraint families, and each candidate costs a column generation run. The choice is hard when families strengthen the bound only together, as in classical lot-sizing, network-design, and facility-location formulations. We prove that when an evaluation convexifies at most $s$ of $k$ families, methods that observe only bounds need of order $k^w$ evaluations to find a hidden group of $w$ families in the worst case. A prediction missing $e$ members reduces this to $\Theta(k^e)$ through a staged search that need not know $e$. The orders extend to several groups, prices, and growing evaluation costs. We study two predictors: a lookahead at an optimal LP dual, which vanishes on groups whose families share no variable, and language models reading the model, verified by the staged search in JointDW. Such groups appear in three of ten MIPLIB-NL models whose bound can be improved and in all 59 benchmark instances. Given the coefficients of constructed models with up to 16 families, one language model names a true group first in 249 of 252 answers. On the benchmark instances, decompositions chosen by language models from the text solve faster than those of GCG's detector.
\end{abstract}

%%%%%%%%%%%%%%%%%%%%%%%%%%%%%%%%%%%%%%%%%%%%%%%%%%%%%%%%%%%%%%%%%%%%%%%%%%%%%%%%%%%%%%%%%%%%%%%%%%%%%%%%%%%%%%%%
\section{Introduction}\label{sec:intro}

Dantzig--Wolfe (DW) reformulation \citep{dantzig1960decomposition} replaces some constraints of a mixed-integer linear program (MILP) by the convex hull of their integer solutions, and column generation solves the resulting relaxation. The bound of this relaxation is never weaker than that of the linear programming (LP) relaxation, and the reformulation is the basis of branch-and-price for lot sizing, vehicle routing, and scheduling \citep{vanderbeck2010reformulation}. The bound depends on which constraints are convexified, and the bound of a candidate is known only after a column generation run. Convexifying more constraints never weakens the bound but makes the pricing problems harder and every evaluation more expensive, so the choice trades bound strength against computation. Modelers make this choice from knowledge of the model class, for example by convexifying the balance and setup rows of a lot-sizing model together. Generic solvers such as GCG instead detect block structure in the matrix and reformulate automatically \citep{gamrath2010experiments,bergner2015automatic,hojny2025scip}. Without the modeler's knowledge, a detector must compare candidates, and a model with $k$ constraint families has $2^k$ of them. How many evaluations does that knowledge replace, and which sources can supply it without a modeler?

\begin{figure}[t]
\centering
\includegraphics[scale=1.25]{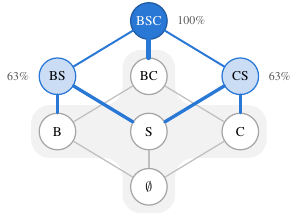}
\caption{Share of the integrality gap closed by every family-level DW relaxation of a lot-sizing instance with two items and two periods, with B, S, and C the balance, setup, and capacity families. Edge width is proportional to the increase of the bound when one family is added, and gray edges add nothing. The shaded sets have gain zero, so a search that adds one family at a time while the bound improves never leaves $\emptyset$. The pairs BS and CS are zero-marginal groups.}
\label{fig:lattice}
\end{figure}

The choice is hard for a specific reason, which \cref{fig:lattice} shows on a lot-sizing instance with two items and two periods. Convexifying the balance rows, the setup rows, or the capacity rows alone leaves the LP bound unchanged, while convexifying the balance and setup rows together closes most of the integrality gap. Together, the two families imply the $(l,S)$ inequalities of single-item lot sizing \citep{barany1984uncapacitated}, which neither implies alone. A search that adds one family at a time while the bound improves therefore returns the LP relaxation, and a score that evaluates each family by itself sees no gain. We call a set of families with this property a zero-marginal group. The classical strong formulations of lot sizing, network design, and facility location rest on such groups, because each of their strong inequalities follows from two families together. What modelers know about these formulations is therefore the information that single-family evaluations cannot reveal. A general account of when families must be convexified together, and of what it costs to find them, has been missing.

Bound evaluations alone find such groups slowly. On binary programs whose DW relaxations we construct exactly, a bound reveals only whether the convexified set contains a hidden group, and every method needs a covering number of evaluations to find it. Each correctly predicted member of the hidden group removes a factor of order $k$ from this cost. For $k=20$ families, a maximum of six families per evaluation, and a hidden group of size three, greedy covering needs 85 evaluations in the worst case, compared with just one for a correct prediction (\cref{fig:cost}(b)). The evaluations a prediction saves thus measure what the predictor knows in units of computation, not by agreement with a reference answer. This holds for any source of predictions, and it makes the choice of a decomposition a learning-augmented search \citep{mitzenmacher2022algorithms} in which the number of missed members measures the prediction error.

Three sources of predictions exist before any column generation run on the instance at hand. Solving each block once at an optimal LP dual gives a lower bound on its gain, and we prove that this lookahead cannot see groups whose families share no variable. Evaluations of similar instances can train a predictor, at the cost of the evaluations that label them. The text of the model is the third source, since a modeler who knows lot sizing knows that balance and setup rows belong together. Language models now read such text when they formulate models \citep{huang2025orlm,ahmaditeshnizi2024optimus,liang2026large,lian2026reloop}, check reformulations \citep{zhai2025equivamap,robbins2026flare}, and configure solvers \citep{lawless2024llms,nie2026milpevo,xu2026agentic}. Whether they know which constraints belong together is open. A correct prediction may come from the meaning of a model or only from familiar names, and its value depends on the search that checks it. We make four contributions.
\begin{itemize}[leftmargin=*,itemsep=1pt,topsep=2pt,parsep=0pt]
\item \textbf{Zero-marginal groups.} We define the family supports of a valid inequality and the zero-marginal groups of constraint families. A group is polyhedral when its gain needs an inequality with a joint support, and LP-geometric when it arises from the geometry of the LP (\cref{sec:structure}). The gain functions of family-level DW relaxations are exactly the nondecreasing set functions with rational values that vanish at $\emptyset$ (\cref{prop:gain}). In the classical formulations of lot sizing, network design, and facility location, as stated in \cref{app:templates}, every family alone leaves the LP relaxation unchanged. Every pair of families with positive gain is therefore a polyhedral group, and each classical inequality that two families imply together has them as a joint support whenever it cuts off a point of the LP relaxation (\cref{prop:classical}).
\item \textbf{The cost of finding them.} When an evaluation may convexify at most $s$ of $k$ families, every method that observes only bounds needs $\Theta(k^w)$ evaluations to find a group of $w$ families, and this lower bound already holds on DW relaxations of stable-set programs. Greedy selection never finds such a group (\cref{thm:cover}). A prediction that misses $e$ members reduces the cost to $\Theta(k^e)$ through one staged search. The search does not need to know $e$, it attains the deterministic lower bound for methods that know $e$ up to a factor $1+o(1)$, and it costs only a lower-order term more than prediction-free search when the prediction is entirely wrong (\cref{thm:pred}). With several groups, the $r$-th best prediction sets the cost (\cref{thm:several}). When an evaluation costs the $d$-th power of the number of convexified families, the search interleaved with a halving search attains the optimal cost $\Theta(k^{\min(e,d)})$ (\cref{thm:costMain}). The search also extends to unknown groups under a price per convexified family, for gains that add up and for gains that overlap when no group contains another, and in each case its order is tight for disjoint groups (\cref{thm:priceSearch,thm:overlap}). Combining known groups is the hard step. When groups of two families share families and their gains add up, the best set under the budget is a densest subgraph. Under a price, one minimum cut finds the best set when the gains add up, and the best set encodes set cover when they overlap (\cref{prop:combine}).
\item \textbf{Two sources of predictions.} The one-pass lookahead at an optimal LP dual provably vanishes on groups whose families share no variable (\cref{prop:lookahead}). JointDW takes the prediction of a language model, verifies it with the staged search, and alternates it with the lookahead order, which delays each of the two orders by at most a factor of two (\cref{sec:method}).
\item \textbf{A computational study.} We report a census of zero-marginal groups in MIPLIB-NL and on 59 standard benchmark instances. We measure five language-model configurations on instances with exact gains, and two of them on models with up to 16 families, against the lookahead, GCG's detector, and a predictor trained on labeled evaluations. We also measure what the chosen sets cost in branch-and-price, including on 76 instances without a joint group, and what a prediction costs in latency. Controlled presentations separate knowledge of a model from familiarity with its names, and we vary reasoning and whether the coefficients are shown (\cref{sec:experiments}). Code and data are public.
\end{itemize}
Zero-marginal groups are common in MIPLIB-NL, and greedy selection misses the best decomposition on most models that have one. On standard benchmark instances, GCG's detector convexifies the block structure of the matrix and not the group that carries the gain. The language models know what the textbooks say about these groups, and on textbook models they know it through the names: once the family names are permuted among the rows, GPT-6.1 Sol names no true group on the templates. On MIPLIB-NL, the strongest language models identify the groups about as often as the lookahead and on partly different models, and reasoning roughly doubles the accuracy of the same model. Given every coefficient of constructed models with 7 to 16 families, one reasoning model names a true group first in almost every answer, on the smaller models also with the names permuted or removed. The evaluations it saves against a covering grow with the number of families. A predictor trained on the gain tables of the same class matches this only after hundreds of labeled evaluations. On most benchmark instances, the best pair is also the fastest set of at most two families for branch-and-price, and the language models choose faster sets than GCG's detector. A prediction pays for its latency only where evaluations are slow. Because the interleaved order bounds the loss from a wrong prediction, a more accurate language model can be adopted as soon as one exists.

%%%%%%%%%%%%%%%%%%%%%%%%%%%%%%%%%%%%%%%%%%%%%%%%%%%%%%%%%%%%%%%%%%%%%%%%%%%%%%%%%%%%%%%%%%%%%%%%%%%%%%%%%%%%%%%%
\section{Related Literature}\label{sec:related}

\paragraph{Strength of DW relaxations and the cost of pricing.}
Branch-and-price solves the DW reformulation of a MILP by column generation at every node of the search tree \citep{barnhart1998branch,lubbecke2005selected}. Geoffrion's theorem \citep{geoffrion1974lagrangean} identifies the bound of a Lagrangian relaxation, and hence of a DW reformulation, with the optimum over the convex hull of the kept constraints intersected with the dualized constraints. \citet{guignard1987lagrangean} showed that intersecting two such hulls can dominate both relaxations. \citet{bastubbe2018computational} computed the bounds of all constraint subsets of small integer programs and found few distinct bounds. \citet{lubbecke2018strength} characterized the DW reformulations of the edge formulation of stable set that are no stronger than its LP relaxation and those that give its integer hull, the setting of our lower-bound constructions. \citet{witt2020polyhedral} studied which valid inequalities DW reformulations imply. What the pricing problem keeps also decides its difficulty. The ng-route relaxation of vehicle routing lets a route revisit customers outside small neighbor sets, and the size of these sets trades the quality of the bound against computing time \citep{baldacci2011new}. In our model, the gain of a set of families measures the bound, and a budget, an evaluation cost, or a price measures the computation (\cref{sec:cost}).

\paragraph{Automatic decomposition.}
Automatic DW reformulation starts from the structure of the constraint matrix. Graph and hypergraph partitioning permute the matrix into arrowhead or bordered block-diagonal form \citep{ferris1998partitioning,wang2013computational,bergner2015automatic}, while community detection finds such forms without requiring a predefined number of blocks \citep{khaniyev2018structure}. GCG detects these structures and solves the reformulation by branch-and-price \citep{gamrath2010experiments}. One of its detectors enumerates subsets of constraint classes, which are the families of this paper \citep{gcg2026consclass}, and its strong score solves each block once at a mixture of LP duals and random duals \citep{gcg2026scores}. These methods judge a decomposition by the shape of its blocks, which does not reveal zero-marginal groups. On the benchmark instances of \cref{sec:experiments}, GCG's detector convexifies a set with no gain on most network-design and facility-location instances. Learned detectors decide from structural features whether and how to decompose \citep{kruber2017learning,basso2020random,basso2023data}. They are trained on instances drawn from a single distribution and see only the constraint matrix, while a language model needs no training instances and also reads the model's names and the description. We train a predictor of this kind on our own gain tables and compare the two (\cref{sec:experiments}). \Cref{prop:lookahead} states which groups a one-pass evaluation at an optimal LP dual cannot see.

\paragraph{Language models for optimization.}
Language models formulate optimization models from text \citep{huang2025orlm,ahmaditeshnizi2024optimus,liang2026large}. Generated formulations are checked by perturbation responses \citep{lian2026reloop}, agreement across model families \citep{lian2026admitor}, solver feedback \citep{zhu2026solid}, and variable mappings, Lean proofs, or SMT solvers \citep{zhai2025equivamap,robbins2026flare,bhattacharyya2026sover}. MIPLIB-NL pairs MIPLIB 2017 instances with natural-language specifications \citep{li2026constructing}. Closer to our question, language models evolve acceleration inequalities and faster formulations \citep{yazdani2026sage,yuan2026formuevo} and cut selection and branching rules \citep{nie2026milpevo}, configure cutting-plane separators \citep{lawless2024llms}, and write constraint handlers that recover global constraints in MIPLIB models \citep{xu2026agentic}. FrontierOR \citep{kong2026frontieror} and \citet{baek2026llms} evaluate algorithm design for OR problems, DecompGym asks coding agents to implement a prescribed decomposition \citep{cheng2026decompgym}, and \citet{le2026provably} learn the multipliers of a fixed Lagrangian relaxation. In these works, the decomposition is absent or given. Language-model agents have also solved open MIPLIB instances and improved the best-known solutions and bounds of others, with the language model searching for exploitable structure, primal solution procedures, and dual certificates and solvers supplying the bounds \citep{huang2026llm4mip}. We study one such structural decision in isolation, the choice of the families to convexify, and we measure the contribution of a language model by its number of misses, which \cref{thm:pred} converts into search cost.

\paragraph{Search with predictions.}
Algorithms with predictions are analyzed by their consistency and robustness \citep{mitzenmacher2022algorithms}. Learning-augmented query policies trade one against the other \citep{erlebach2020learning}, and the cost of graph search with predictions grows with the number of wrong predictions \citep{banerjee2022graph}. With several predictions, online algorithms can compete with the best of them \citep{anand2022online,dinitz2022algorithms}. With several groups, the cost of the staged search is set by the $r$-th best prediction (\cref{thm:several}). Theory revision corrects an initial formula with a number of queries that grows with its distance from the target \citep{goldsmith2004theory}. Under the tractability budget, each miss of a predicted group instead multiplies the cost by a factor of order $k$ (\cref{thm:pred}). Our oracle reveals whether a query contains a hidden group, as in learning a hidden hypergraph with edge-detecting queries \citep{angluin2006learning} and in group testing with tests of bounded size \citep{gandikota2019nearly}. The tractability budget makes covering designs \citep{fort1958minimal,schonheim1964coverings,rodl1985packing,gordon1995new} the exact measure of cost.

%%%%%%%%%%%%%%%%%%%%%%%%%%%%%%%%%%%%%%%%%%%%%%%%%%%%%%%%%%%%%%%%%%%%%%%%%%%%%%%%%%%%%%%%%%%%%%%%%%%%%%%%%%%%%%%%
\section{Family-Level DW Relaxations}\label{sec:structure}

Consider $\min\{cx : x\in X\}$ with
\[
X=\{x\in\mathbb{Z}^{p}\times\mathbb{R}^{q} : A_f x\ge b_f\ \text{for}\ f\in F,\ \ell\le x\le u\},
\]
rational data, finite bounds that are integral for the integer variables, and $X\neq\emptyset$. The rows are partitioned into $k=|F|$ constraint families. In a modeling language, a family is one indexed constraint, such as all balance rows. For $T\subseteq F$, let $X_T$ keep only the rows of the families in $T$, let $\LP$ be the LP relaxation of $X$, and let
\begin{equation}\label{eq:PT}
P_T=\LP\cap\conv(X_T),\qquad z(T)=\min_{x\in P_T}cx .
\end{equation}
The gain of $T$ is $g(T)=z(T)-z(\emptyset)$. By Geoffrion's theorem, $z(T)$ is the bound of the DW reformulation that convexifies the rows of $T$ and keeps those of $F\setminus T$ in the master problem. Each connected component of the rows of $T$ forms one block. Since $X_{T'}\subseteq X_T$ when $T\subseteq T'$, the gain $g$ is nondecreasing, $g(\emptyset)=0$, and $g(F)=z_{\IP}-z_{\LP}$, where $z_{\LP}=z(\emptyset)$ is the LP bound and $z_{\IP}=z(F)$ is the optimal value of the MILP.

\begin{proposition}[gain functions]\label{prop:gain}
A function $g:2^F\to\mathbb R$ is the gain function of a program with $k$ constraint families if and only if it is nondecreasing, takes rational values, and satisfies $g(\emptyset)=0$. Every such function is the gain function of a binary program whose rows form the edge formulation of stable set.
\end{proposition}

\Cref{app:overlap} proves the proposition by writing $g$ as a weighted sum of indicators of its level sets and realizing each indicator by an odd cycle of a stable-set program. Convexifying $F$ solves the MILP itself, and convexifying many families makes the pricing problems computationally difficult. We model tractability by a family $\cT$ of sets of families, closed under taking subsets, with $\cT_s=\{T : |T|\le s\}$ as the main case.

\begin{definition}[family support]\label{def:support}
Let $\alpha x\ge\beta$ be valid for $\conv(X)$. The sets $T$ for which $\alpha x\ge\beta$ is valid for $P_T$ form an up-set that contains $F$. Its minimal elements are the family supports of the inequality, and a family support with at least two families is a joint support.
\end{definition}

The polytope $P_T$ is the intersection of $\LP$ with all valid inequalities that have a family support inside $T$ (\cref{prop:support} in \cref{app:proofs}). Hence $g(T)>0$ exactly when these inequalities together cut off the whole LP-optimal face.

\begin{definition}[zero-marginal group]\label{def:zmg}
Let $B\subseteq F$, and let $W\subseteq F\setminus B$ have at least two families. $W$ is a zero-marginal group at $B$ if $g(B\cup W)>g(B)$ and $g(B\cup S)=g(B)$ for every $S\subsetneq W$; its order is $w=|W|$. Let $Q_W=\bigcap_{S\subsetneq W}P_{B\cup S}$. The group is LP-geometric if $\min\{cx : x\in Q_W\}>z(B)$ and polyhedral otherwise.
\end{definition}

The two kinds differ in what produces the gain. In a polyhedral group, the gain needs an inequality that is not valid for $Q_W$. Every family support of such an inequality inside $B\cup W$ contains all of $W$, because a support that misses some $f\in W$ lies in $B\cup(W\setminus\{f\})$. At $B=\emptyset$, $W$ is therefore a joint support of an inequality that cuts off an LP optimum. In an LP-geometric group, inequalities whose supports each miss a member of $W$ cut off every LP optimum only together. For example, when each of two bound rows fixes one variable, convexifying one row lets the LP shift its weight to the other variable (\cref{ex:geo} in \cref{app:exact}). Both groups in \cref{fig:lattice} are polyhedral.

\paragraph{Classical inequalities as joint supports.}
In lot sizing, the balance and setup rows of an item together imply the $(l,S)$ inequalities \citep{barany1984uncapacitated}, and the capacity and setup rows of a period together imply the flow cover inequalities \citep{padberg1985valid}. In network design, the flow-balance rows summed over one side of a cut bound the flow across the cut from below, and, with the arc-capacity rows and the integrality of the opening variables, this gives the cut-set inequalities \citep{atamturk2002capacitated,raack2011cut}. In capacitated facility location, the capacity rows summed with the assignment rows give a knapsack constraint on the opening variables, whose cover inequalities need both families \citep{aardal1995capacitated}, and the assignment and linking rows together imply the odd-cycle inequalities of \citet{cornuejols1982some}. In these three formulations, as stated in \cref{app:templates}, every family alone leaves the LP relaxation unchanged. As a result, every pair of families with positive gain is a polyhedral zero-marginal group at $\emptyset$, and each of these inequalities that cuts off a point of the LP relaxation has its two families as a joint support (\cref{prop:classical} in \cref{app:classical}). A modeler who knows the class can therefore name a candidate group without evaluating a bound, and so can a language model that has read the same textbooks (\cref{sec:experiments}).

%%%%%%%%%%%%%%%%%%%%%%%%%%%%%%%%%%%%%%%%%%%%%%%%%%%%%%%%%%%%%%%%%%%%%%%%%%%%%%%%%%%%%%%%%%%%%%%%%%%%%%%%%%%%%%%%
\section{The Cost of Choosing a Decomposition}\label{sec:cost}

A decomposition is judged by its bound and by the computation it needs. Convexifying more families never lowers the bound but makes the pricing problems of the final reformulation harder, and every candidate costs a column generation run while the decomposition is chosen. Three models represent these two computational costs, and each examines the evaluation effort required to find the best decomposition. Under the budget, every evaluated and every chosen set has at most $s$ families, and the cost of a method is its number of evaluations (\cref{thm:cover,thm:pred,thm:several}). Under evaluation costs, the evaluation of a set $T$ costs $|T|^d$ instead (\cref{thm:costMain}). Under a price, the chosen set maximizes $g(T)-\lambda|T|$ for a price $\lambda>0$ per convexified family (\cref{prop:combine,thm:priceSearch,thm:overlap}). A prediction of the families that belong together changes only which sets a method evaluates and in which order.

We analyze search in an oracle model. A method observes the program only through $g$: it queries sets $T\in\cT$ one at a time, each chosen after the previous answers, and observes $g(T)$. Its cost is the number of queries until it has queried a set of maximal gain over $\cT$. It need not certify optimality, so the lower bounds hold for this easier task. For a set $W$ of $w$ families, let $g_W(T)=\tfrac12[W\subseteq T]$, and let $N(\cT,w)$ be the least number of members of $\cT$ that together cover every $w$-set contained in some member of $\cT$. For $\cT_s$, this is the covering number $\cov(k,s,w)$, the least number of $s$-subsets of a $k$-set covering all its $w$-subsets.

\begin{theorem}[covering]\label{thm:cover}
Let $2\le w\le s<k$.
\begin{enumerate}[label=(\alph*),leftmargin=*,itemsep=1pt,topsep=2pt]
\item For every $W\subseteq F$ with $|W|=w$, there is a binary program with $k$ constraint families whose gain function is $g_W$. It is the edge formulation of stable set, with objective $\min-\sum_v x_v$, on an odd cycle of length at least $w$ and on $k-w$ further paths, all on disjoint sets of vertices. The edges of the cycle are split into $w$ paths, which form the families of $W$.
\item Every deterministic method needs $N(\cT,w)$ queries in the worst case over $\{g_W\}$, and querying a minimum covering attains this. For $\cT=\cT_s$, every randomized method needs more than $\tfrac12\binom{k}{w}/\binom{s}{w}$ queries in expectation for some $W$.
\item $\binom{k}{w}/\binom{s}{w}\le\cov(k,s,w)\le\bigl(1+\ln\binom{s}{w}\bigr)\binom{k}{w}/\binom{s}{w}$, and $\cov(k,s,w)=(1+o(1))\binom{k}{w}/\binom{s}{w}$ for fixed $s$ and $w$ as $k\to\infty$. A method that does not know $w$ can query minimum coverings for $w'=1,2,\dots$ in turn and pays $\sum_{w'\le w}\cov(k,s,w')=(1+o(1))\cov(k,s,w)$.
\item Adding one family at a time while some addition has a positive marginal gain stops at $\emptyset$ on every $g_W$.
\end{enumerate}
\end{theorem}

Part (a) holds because a forest is bipartite, so convexifying the rows of $T$ changes nothing unless $T$ contains every family of $W$. These families together form the odd cycle, which then contributes the difference of $\tfrac12$ between its LP value and the optimum over its stable-set polytope. On these programs, a bound carries one bit, whether the query contains $W$, so decomposition search under a tractability budget is the search for one hidden hyperedge with edge-detecting queries of bounded size. Without the budget, one query of $F$ attains the maximal gain, and $W$ itself is identified with $O(w\log k)$ queries \citep{angluin2006learning}. Part (b) is an adversary argument, and part (c) follows from counting, from the greedy covering bound of \citet{lovasz1975ratio} and \citet{stein1974two}, and from R\"odl's theorem \citep{rodl1985packing}. \Cref{app:proofs} gives the proofs. The group is visible in the constraint matrix of these programs, so the theorem concerns the information in bound evaluations, and \cref{thm:pred} measures what reading the formulation is worth. Both theorems are worst-case statements over these gain functions. On the gain function of an arbitrary program, the staged search and its alternation with a covering order remain valid search orders, and \cref{sec:experiments} measures their cost.

\begin{theorem}[value of a predicted group]\label{thm:pred}
Let $2\le w\le s<k$, let $g=g_W$ be a gain function of \cref{thm:cover} with hidden group $W$, and let a prediction $P\subseteq F$ with $|P|=w$ miss $e=|W\setminus P|$ members of $W$. For $0\le e'\le w$, an \emph{$e'$-cover} is a family $\mathcal{K}$ of subsets of $P$ with at most $s-e'$ elements each, such that every $(w-e')$-subset of $P$ lies in a member of $\mathcal{K}$. Its weight is $\sum_{A\in\mathcal{K}}\binom{s-|A|}{e'}^{-1}$, and $v_{e'}$ is the least weight of an $e'$-cover. Let $\cov(n,a,t)$ be the least number of sets of at most $a$ elements of an $n$-set that cover its $t$-subsets; it equals $1$ when $t=0$ or $t\le n\le a$ and $0$ when $t>n$, so a stage with $e'>k-w$ has no query.
\begin{enumerate}[label=(\alph*),leftmargin=*,itemsep=1pt,topsep=2pt]
\item \emph{Staged search.} Stage $e'$ takes an $e'$-cover $\mathcal{K}$ and, for each $A\in\mathcal{K}$, a minimum covering of the $e'$-subsets of $F\setminus P$ by sets of at most $s-|A|$ elements, and it queries every $A\cup R$ with $R$ in the covering taken for $A$, with $R=\emptyset$ for $e'=0$. It uses an $e'$-cover with the fewest queries, $\sum_{A\in\mathcal{K}}\cov(k-w,s-|A|,e')$, and the stages run for $e'=0,1,\dots,w$ in turn. Without knowing $e$, this search queries a set that contains $W$ within the number of queries of stages $0$ to $e$.
\item \emph{Lower bounds.} Over the hidden groups with $e$ misses, every deterministic method, even one that knows $e$, needs at least $v_e\binom{k-w}{e}$ queries in the worst case. Every randomized method needs more than $L_e/2$ queries in expectation for some group, where $L_e=\binom{w}{e}\binom{k-w}{e}/\max_{a\in A_e}f_e(a)$ with $A_e=\{a\in\mathbb{Z} : w-e\le a\le\min(w,s-e)\}$ and $f_e(a)=\binom{a}{w-e}\binom{s-a}{e}$. The two bounds satisfy $L_e\le v_e\binom{k-w}{e}$, with equality for $e=w$, where $v_w=1/\binom{s}{w}$, and whenever $e(s+1)\le w(s-w+1)$, where $v_e=1/\binom{s-w}{e}$.
\item \emph{No price for robustness.} For fixed $s$ and $w$ and $k\to\infty$, the cost in (a) is $(1+o(1))\,v_e\binom{k-w}{e}=\Theta(k^e)$, the deterministic lower bound of (b) up to the factor $1+o(1)$. For $e=w$, the cost is at most $\cov(k,s,w)+O(k^{w-1})$.
\end{enumerate}
\end{theorem}

\begin{figure}[t]
\centering
\includegraphics[scale=1.25]{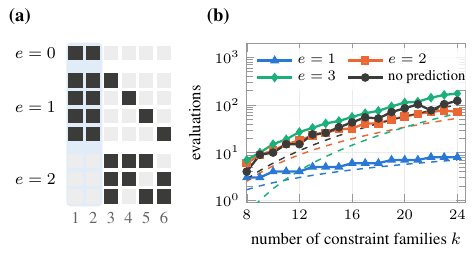}
\caption{(a) Queries of the staged search of \cref{thm:pred}(a), from top to bottom, for $k=6$, $s=3$, $w=2$, and the prediction $P=\{1,2\}$ (shaded columns): one query per row, with its families in dark cells. The queries of stage $e$ contain every hidden group with $e$ members outside $P$, so a correct prediction is confirmed by the first query. (b) Evaluations to find a hidden group of $w=3$ families with $s=6$. Solid lines: for predictions that miss $e$ members, the queries in the stages of the staged search built from greedy coverings and, in each stage, from the cover of $P$ with the fewest queries among those with members of one size; without a prediction, the worst case of a greedy covering. Dashed lines of the same color: the lower bounds $L_e$ of \cref{thm:pred}(b) and $\binom{k}{3}/\binom{6}{3}$ of \cref{thm:cover}.}
\label{fig:cost}
\end{figure}

Every hidden group with $e$ members outside $P$ lies in some query of stage $e$ (\cref{fig:cost}(a)). The deterministic bound of part (b) holds because, for each set $Y$ of $e$ families outside $P$, the queries that contain $Y$ must cover every $(w-e)$-subset of $P$, and a query with $a$ predicted families contains at most $\binom{s-a}{e}$ such sets $Y$. The randomized bound counts the hidden groups that one query can contain, and for $k\ge s+w$, $L_e$ is the value of the linear relaxation of the same covering problem. Part (c) compares the last stage of (a) with (b) through R\"odl's asymptotics. The deterministic cost of a prediction with $e$ misses is therefore $(1+o(1))\,v_e\binom{k-w}{e}$, and each correctly predicted member divides the cost by a factor of order $k$ (\cref{fig:cost}). In learning-augmented algorithms, consistency usually costs robustness \citep{mitzenmacher2022algorithms,erlebach2020learning}. Here, the same search costs one query when the prediction is right, matches the best deterministic method that knows the number of misses up to a factor $1+o(1)$, and costs only a lower-order term more than prediction-free search when the prediction is entirely wrong. For finite $k$, the earlier stages add overhead, for example 56 versus 43 queries at $k=16$ and $e=3$ (\cref{fig:cost}(b)). Interleaving the staged order with a covering order caps the cost at twice the smaller of the two.

\paragraph{Several groups.}
A model can have several zero-marginal groups at $\emptyset$. When each of $m$ pairwise disjoint groups of $w$ families adds $\tfrac12$ to the gain of every set containing it, $r=\min\{m,\lfloor s/w\rfloor\}$ groups fit into one evaluation, and the best sets are those that contain $r$ groups. Disjoint odd cycles, one per group, realize these gain functions as DW relaxations of stable-set programs. A query of positive gain reveals the groups inside it after at most $s$ further evaluations per group, by removing one family at a time.

\begin{theorem}[several groups]\label{thm:several}
Let $m$ pairwise disjoint groups of $w$ families each add $\tfrac12$ to the gain of every set that contains it. Let $m$, $w$, and $s$ be fixed as $k\to\infty$.
\begin{enumerate}[label=(\alph*),leftmargin=*,itemsep=1pt,topsep=2pt]
\item Without predictions, the deterministic cost of a method that knows $m$ and $w$ is $(1+o(1))\binom{k}{w}/\binom{s}{w}$, as in the single-group case, and every randomized method needs $\Omega(k^w)$ evaluations in expectation.
\item Let one predicted set of $w$ families per group miss $e_1,\dots,e_m$ of its members, and let $e_{(r)}$ be the $r$-th smallest of these numbers, with $r=\min\{m,\lfloor s/w\rfloor\}$. The staged search of \cref{thm:pred}, applied stage by stage around all predictions, finds a set of maximal gain within $O(k^{e_{(r)}})$ evaluations without knowing the misses, and within $\cov(k,s,w)+O(k^{w-1})$ evaluations for any predictions.
\item Every method needs $\Omega(k^{e_{(r)}})$ evaluations, in the worst case for deterministic methods and in expectation for randomized methods, even if the method knows the misses.
\end{enumerate}
\end{theorem}

The cost is therefore $\Theta(k^{e_{(r)}})$. It is determined by the $r$ best predictions, regardless of how many members the others miss, and by the worst prediction when all groups fit into one evaluation. \Cref{app:multi} proves the theorem with explicit bounds (\cref{thm:multi}).

\paragraph{Combining known groups.}
Groups can also share families: in \cref{fig:lattice}, the setup rows form one group with the balance rows and another with the capacity rows. The best set must then combine groups, under the budget or under a price. The gains of these two groups are not additive: BS and CS each close 63 percent of the gap, and together they close all of it. We model such overlapping gains as weighted sums over units of the gap, each of which is closed by any one of several groups (\cref{app:overlap}). Every gain function has this form, by the proof of \cref{prop:gain}.

\begin{proposition}[combining known groups]\label{prop:combine}
\begin{enumerate}[label=(\alph*),leftmargin=*,itemsep=1pt,topsep=2pt]
\item Let $g(T)=\tfrac12|E(T)|$ for a graph $(F,E)$ on the families, where $E(T)$ is the set of edges with both ends in $T$. Each edge is a group of two families with gain $\tfrac12$, and the gains add up. Choosing a set of maximal gain over $\cT_s$ is then equivalent to the densest-subgraph problem with $s$ vertices, which is NP-hard when $s$ is part of the input.
\item For gains that are additive over groups, one minimum cut finds a set that maximizes $g(T)-\lambda|T|$. Such a set has the largest gain among the sets of at most as many families, and one parametric minimum cut gives the largest such set for every price.
\item For overlapping gains, maximizing $g(T)-\lambda|T|$ is NP-hard, even when every unit is known and the groups each contain two families, share one family, and are otherwise disjoint.
\end{enumerate}
In (a) and (c), the hardness persists even for binary programs whose bounds are all computed in polynomial time.
\end{proposition}

The densest-subgraph problem is also hard to approximate under standard complexity assumptions \citep{feige2001dense,khot2006ruling,manurangsi2017almost}. Finding the groups costs evaluations, and combining them costs computation, except under a price with additive gains. The prices of (b) solve the budget problem exactly at the budgets where the optimal gain lies on its concave envelope. The groups of (c) share a family as in \cref{fig:lattice}, the structure in which additive gains give the price-optimal set in closed form. \Cref{app:multi,app:cost,app:overlap} prove the three parts (\cref{prop:dks,prop:price,prop:overlapHard}).

\paragraph{Evaluation cost.}
Suppose that the evaluation of a set $T$ costs $|T|^d$ and that $|T|$ is not bounded. One evaluation of $F$ then attains the maximal gain, so the target becomes identifying the group itself. On the gain functions $g_W$ of \cref{thm:cover}, $W$ is the unique maximizer of $g_W(T)-\lambda|T|$ for every price $0<\lambda<1/(2w)$ per convexified family. A method knows $w$, and its cost is the sum of the evaluation costs incurred until it has evaluated $W$.

\begin{theorem}[evaluation cost]\label{thm:costMain}
Let the evaluation of a set $T$ cost $|T|^d$ with $d>0$, and let $w$ and $d$ be fixed as $k\to\infty$. Without a prediction, the optimal cost of evaluating a hidden group $W$ of $w$ families is $\Theta(k^{\min(w,d)})$. With a prediction that misses $e$ members, it is $\Theta(k^{\min(e,d)})$, attained without knowing $e$ by interleaving the staged search with a search that evaluates large sets and halves them. Both orders hold for deterministic and for randomized methods.
\end{theorem}

A prediction thus reduces the optimal cost by a factor of order $k^{w-e}$ when $d\ge w$, as under the budget, by $k^{d-e}$ when $e<d<w$, and by a constant when $d\le e$. \Cref{app:cost} proves the theorem with explicit bounds (\cref{thm:power}).

\paragraph{Unknown groups under a price.}
When the groups are unknown, a staged search around the union of the predictions finds them. For gains that add up over the groups, the price-optimal decomposition is determined by a certificate. The certificate is a collection of groups whose identification fixes the largest maximizer of $g(T)-\lambda|T|$. Let $e^*$ be the largest number of families that the predictions miss in a group of the certificate they cover best. With the number of groups, their sizes, $s$, $d$, and the number of predicted families fixed as $k\to\infty$, the staged search reaches the price-optimal decomposition within $O(k^{e^*})$ evaluations under the budget. Interleaving it with a halving search reaches the same decomposition at a cost of $O(k^{\min(e^*,d)})$  (\cref{thm:priceSearch} in \cref{app:cost}). For overlapping gains of groups of which none contains another, the search queries an optimal set within the same orders, with $e^*$ replaced by the corresponding number $\bar e$ for the collections of groups whose union is optimal (\cref{thm:overlap} in \cref{app:overlap}). This case includes the two groups of \cref{fig:lattice}, which share the setup family. In both cases, the orders are tight for pairwise disjoint groups when the predictions miss at least one family. For disjoint groups with additive gains, the $r$-th best prediction sets the order of the cost under the budget of \cref{thm:several}. Under a price, every group of the certificate must be identified, and the group that the predictions cover worst sets the order.

\paragraph{Side information from LP duals.}
Greedy selection over $\cT_s$ reaches at least a fraction $1-\exp(-\gamma)$ of the best gain when the submodularity ratio $\gamma$ of $g$ is positive \citep{das2011submodular}, and a zero-marginal group at $\emptyset$ of order $2\le w\le s$ forces $\gamma=0$. A cheaper signal than column generation solves each block of $T$ once at an optimal dual $y$ of the LP relaxation, and we write $\psi_y(T)$ for the resulting increase of the bound.

\begin{proposition}[one-pass lookahead]\label{prop:lookahead}
For every optimal dual $y$ of the LP relaxation, $0\le\psi_y(T)\le g(T)$, the function $\psi_y$ is nondecreasing, and $\psi_y(T_1\cup T_2)=\psi_y(T_1)+\psi_y(T_2)$ whenever no variable appears in a row of $T_1$ and in a row of $T_2$. Consequently, $\psi_y$ vanishes on every zero-marginal group at $\emptyset$ whose families pairwise share no variable, every such group is LP-geometric, and every superset of a set receives a value at least as large as the set.
\end{proposition}

A polyhedral group at $\emptyset$ therefore has two families that share a variable, and on such a group the value of $\psi_y$ can depend on the dual. On the instance of \cref{fig:lattice}, $\psi_y$ of capacity with setup is zero at the dual returned by HiGHS and stays below the gain at every optimal dual (\cref{app:exact}). The order of decreasing $\psi_y$ is the prediction of the LP itself, and GCG's strong score uses a yes-or-no version of it \citep{gcg2026scores}. Taking for each set the largest value over several optimal duals gives a stronger order, and both orders serve as baselines in \cref{sec:experiments}. \Cref{lem:lookahead} in \cref{app:proofs} proves the proposition.

%%%%%%%%%%%%%%%%%%%%%%%%%%%%%%%%%%%%%%%%%%%%%%%%%%%%%%%%%%%%%%%%%%%%%%%%%%%%%%%%%%%%%%%%%%%%%%%%%%%%%%%%%%%%%%%%
\section{JointDW}\label{sec:method}

\Cref{prop:lookahead} leaves groups whose families share no variable, and polyhedral groups on which the dual gives no signal, to information that does not come from the LP. JointDW takes it from a language model that reads the text of the model: it asks which families must be convexified together and checks the answer with the staged search of \cref{thm:pred} (\cref{alg:jointdw}). The prediction replaces what bound evaluations cannot do efficiently, and the search limits the cost of a wrong prediction. As in solver-based checks of generated formulations \citep{lian2026admitor, lian2026reloop}, the answer enters the result only through computation: a predicted group counts only after bound evaluations confirm its gain.

\begin{algorithm}[t]
\caption{JointDW}
\label{alg:jointdw}
\footnotesize
\begin{algorithmic}[1]
\REQUIRE MILP with constraint families $F$ and its presentation; budget $s$; time limit $\tau$ per evaluation; $N$ evaluations
\STATE ask the language model which families must be convexified together, given the presentation of the MILP and no bound
\STATE $P\gets$ first joint support in the answer, or the families that its top-ranked reformulation convexifies if it lists none; if $|P|>s$, keep the first $s$ families of $P$ in the order of the families
\STATE $O_1\gets$ staged order of \cref{thm:pred}(a) around $P$, with $w=|P|$
\STATE $O_2\gets$ covering orders for $w'=1,\dots,s$ in turn
\STATE $O\gets$ alternate the entries of $O_1$ and $O_2$, skipping repeats
\FOR{the first $N$ sets $T$ in $O$}
\STATE run column generation at the root node on the DW reformulation that convexifies $T$ with time limit $\tau$, and record its bound, a valid lower bound even if it has not converged
\ENDFOR
\STATE \textbf{return} the set $T$ with the best recorded bound
\end{algorithmic}
\end{algorithm}

\paragraph{Prediction and verification.}
The language model receives the description and the formulation of the model when they exist, and tables of the constraint families and variable families with their sizes, senses, constraint types, and variable types, but no bound, optimum, or solver output (\cref{app:prompt}). Its answer is one JSON object with three parts. For each family, it states whether convexifying the family alone strengthens the bound and whether the family has the integrality property. It lists the joint supports, most important first, each with the class of inequalities it generates. It ranks up to three DW reformulations, each given by the families kept in the master. The anchor $P$ is the first joint support, the predicted group of \cref{thm:pred}, or the families that the top-ranked reformulation convexifies if the answer lists none. In each stage, the staged order takes the cover of $P$ with the fewest queries among those whose members have one size, with greedy coverings of the other families. For $s\le3$, some cover with members of one size attains $v_{e'}$ in every stage. An evaluation that does not finish within $\tau$ keeps its bound as a valid lower bound. The orders are fixed before the first evaluation, so a time limit changes the recorded bounds but not the position at which a set is evaluated. The alternation takes one entry of each order in turn, so the $j$-th entry of either order is reached within $2j$ evaluations. On the gain functions of \cref{thm:cover} with $|P|=|W|$, the cost is therefore at most twice the smaller of the staged cost of \cref{thm:pred}, $O(k^e)$ for a prediction with $e$ misses, and the cost of prediction-free search in \cref{thm:cover}(c). For any anchor, it is at most twice the number of entries of $O_2$ up to its first set of maximal gain, repeated sets included. This bound covers an anchor whose size differs from that of the group, an anchor that is not a group, and a program without a joint support. The order of \cref{alg:jointdw} can also be alternated with the lookahead order of \cref{prop:lookahead}, skipping repeats. We call the result the interleaved order. It delays each of the two orders by at most a factor of two, counted in entries: an entry whose set was evaluated before or exceeds the budget takes its turn without a column generation run.

\paragraph{Implementation.}
For an exported MIPLIB-NL model, the families are defined by row names with their indices removed (\cref{app:miplib}). A set $T$ of families becomes a GCG decomposition whose blocks are the connected components of the rows of $T$. The rows of $F\setminus T$ stay in the master. The census also evaluates each set with families of variables declared as linking variables, which several blocks may share (\cref{app:miplib}). An evaluation runs column generation at the root node under a time limit that serves as the budget, with presolving, separation, primal solution procedures, and propagation disabled, so that its bound comes from the relaxation alone. A bound is discarded if it is nonfinite, lies beyond the reference optimum or a known solution by more than a tolerance. The lookahead solves the same blocks once at an optimal dual of the LP relaxation. On the templates and the rich models, every gain is instead computed exactly from an explicit LP (\cref{app:exact}).

\paragraph{What the prediction is read from.}
A correct prediction can come from knowledge of the model class, from the names alone, or from cues in the statistics. Four presentations of the same model separate these sources: the \emph{original} presentation shows everything, \emph{names only} withholds the description and the formulation, \emph{permuted} also permutes the family names among the families while the statistics stay with their rows, and \emph{anonymized} replaces the names by $F_1,F_2,\dots$ for the constraint families and $V_1,V_2,\dots$ for the variable families. Answers are mapped back to the true families before scoring, so predictions that follow the names move to the wrong families under the permutation, and predictions that follow the statistics survive both changes.

%%%%%%%%%%%%%%%%%%%%%%%%%%%%%%%%%%%%%%%%%%%%%%%%%%%%%%%%%%%%%%%%%%%%%%%%%%%%%%%%%%%%%%%%%%%%%%%%%%%%%%%%%%%%%%%%
\section{Computational Study}\label{sec:experiments}

\begin{table}[t]
\caption{Instance collections. $k$: number of constraint families. Gains are exact, from explicit LPs solved in rational arithmetic, or come from column generation with GCG at the root node for the sets of at most $k-1$ families. On the network-design rich models with 14 families, the exact gains cover the sets of at most two families.}
\label{tab:instances}
\centering
\small
\setlength{\tabcolsep}{5pt}
\begin{tabular}{@{}lccll@{}}
\toprule
Collection & Instances & $k$ & Gains & Measures \\
\midrule
Templates & 15 & 2--6 & exact, every set & accuracy of a prediction \\
MIPLIB-NL & 108 & 2--6 & column generation & census, accuracy, search cost \\
Rich models & 24 & 7--9 & exact, sets of at most three & search cost as $k$ grows \\
 & 60 & 12--16 & exact, sets of at most three & \\
Benchmark instances & 59 & 3 & column generation & solve time \\
Without a joint support & 76 & 2--3 & column generation & false alarms, solve time \\
\bottomrule
\end{tabular}
\end{table}

\paragraph{Instances.}
\Cref{tab:instances} lists the five collections. The templates of \cref{app:templates} have exact gains for every set of families, computed from explicit LPs whose optimal bases are verified in rational arithmetic (\cref{app:exact}). Nine of them are lot-sizing, network-design, and facility-location instances whose groups are the joint supports named in \cref{sec:structure}. Two are generalized assignment instances without a joint support, which measure false alarms, and four are hidden-group programs from \cref{thm:cover}(a), whose presentation carries no information about the group. MIPLIB-NL pairs MIPLIB 2017 instances with natural-language specifications and reconstruction scripts \citep{li2026constructing}, and we keep the exported models with two to six constraint families. GCG on SCIP~10 \citep{hojny2025scip} evaluates every decomposition with at most $k-1$ families at the root node, under a time limit that acts as the tractability budget (\cref{app:miplib}). A zero-marginal group is certain if every zero gain below it comes from a converged evaluation and possible otherwise, and 102 models with a bounded evaluation are scored.

The rich models are connected programs of lot sizing, network design, and facility location that place the classical group among further modeling features, so that a covering of all pairs is large (\cref{app:templates}). A draw in which a single family closes more than half of the best gain of a set of at most three families is rejected, so on every rich model the best such set needs several families together. The benchmark instances are the standard formulations of the same three classes on published data. Each instance has the three families of its class. They comprise the 20 smallest capacitated lot-sizing instances from \citet{trigeiro1989capacitated}, 20 fixed-charge network-design instances from the R set of \citet{crainic2001bundle}, and the 19 capacitated facility-location instances from the OR-Library \citep{beasley1990orlibrary} with a positive LP gap. The generalized assignment and bin-packing instances from the OR-Library have no joint support within the budget and measure false alarms on published data. On these two collections, GCG branch-and-price runs with three seeds and a limit of 1800 seconds to measure the impact of decomposition choices on solve time (\cref{app:templates,app:miplib}). The templates and MIPLIB-NL thus measure whether a prediction is right, the rich models what it saves as the number of families grows, and the benchmark instances what a choice costs in seconds.

\paragraph{Language models.}
Five configurations from four providers answer every template, MIPLIB-NL model, and rich model with seven to nine families three times: Claude Sonnet 5.5 with its default thinking, GPT-6.1 Sol with medium reasoning effort, Grok 4.20 (Reasoning), Grok 4.20 (Non-Reasoning), and DeepSeek-V4-Pro in non-thinking mode. Claude Sonnet 5.5 and GPT-6.1 Sol also answer the rich models with 12 to 16 families, the benchmark instances, and the 76 instances without a joint support, three times each. \Cref{app:prompt} gives the prompt and the settings of each configuration. On the rich models and the benchmark instances, a coefficient variant of the prompt also gives the complete instance with all coefficients.

\paragraph{Measures.}
On an instance with a true joint support, hit@1 records whether the first predicted joint support is true, and recall is the share of true joint supports that the answer lists. Rates average the samples of an instance and then the instances. The lookahead order ranks the sets of at most $k-1$ families by decreasing $\psi_y$ at the dual returned by HiGHS and is scored as an answer with one joint support. The five-dual lookahead order takes for each set the largest value over this dual and four further optimal duals (\cref{app:miplib}). Search cost is the number of evaluations until a set of maximal gain among the sets of at most $s$ families is evaluated. It is averaged over random orders of the families. An order that evaluates no such set on an instance has no cost there and is left out of the mean, and a sample without a cost is left out of the mean of its instance. On MIPLIB-NL, where the recorded gains need not be monotone, this happens for $s=3$ and $s=k-1$, to the covering order on at most two models and to JointDW on at most one. The baselines are a random order, greedy selection, the covering order, which queries greedy coverings of the pairs, triples, and larger sets in turn, the two lookahead orders, and the staged search anchored at GCG's own decomposition. A further baseline is a learned predictor, gradient-boosted trees in the manner of \citet{basso2023data} that score a set from structural features of its decomposition, with or without its lookahead value, and are trained on the gain tables of other instances (\cref{app:results}).

On the benchmark instances and the instances without a joint support, each method contributes one set. For JointDW, this set is the anchor of line~2 of \cref{alg:jointdw} before it is cut to $s$ families. The time statistic of a set is the geometric mean over the three seeds of $t+1$, where $t$ is the branch-and-price time in seconds and an unsolved run is assigned $t=1800$. A set ranks above another if it solves on more seeds. When both solve on the same positive number of seeds, the set with the smaller time statistic ranks higher. When neither solves, the set with the smaller mean final gap ranks higher. The time-optimal set is the first set within the budget in this ranking. The time ratio of a method is the geometric mean over instances of the ratio between its set's time statistic and that of the time-optimal set. A prediction costs the latency of its request. The net time that an order saves against another is the number of evaluations it saves multiplied by the mean time per evaluation, minus this latency. The break-even time of one evaluation is the latency divided by the evaluations saved. Standard errors come from a two-stage bootstrap that resamples the groups of instances with identical prompts and then the instances within them. A difference exceeds sampling variation when its 95 percent bootstrap interval excludes zero (\cref{app:results}).

\begin{table}[t]
\caption{Prediction of joint supports in the original presentation. Templates: the nine lot-sizing, network-design, and facility-location instances. MIPLIB-NL: the 17 scored models with a certain joint group. False alarms: the share of answers that name a joint support on the two generalized assignment instances, which have none. Standard errors, the other presentations, and the five-dual lookahead order: \cref{app:results}.}
\label{tab:llm}
\centering
\small
\setlength{\tabcolsep}{5pt}
\begin{tabular}{@{}lccccc@{}}
\toprule
& \multicolumn{2}{c}{Templates} & \multicolumn{2}{c}{MIPLIB-NL} & False \\
\cmidrule(lr){2-3}\cmidrule(lr){4-5}
Configuration & hit@1 & recall & hit@1 & recall & alarms \\
\midrule
Claude Sonnet 5.5 & 0.70 & 0.74 & 0.57 & 0.56 & 0.17 \\
GPT-6.1 Sol & 1.00 & 0.96 & 0.53 & 0.53 & 0.00 \\
Grok 4.20 (Reasoning) & 0.67 & 0.67 & 0.55 & 0.55 & 0.00 \\
Grok 4.20 (Non-Reasoning) & 0.33 & 0.17 & 0.29 & 0.29 & 0.00 \\
DeepSeek-V4-Pro (Non-Thinking) & 0.30 & 0.26 & 0.27 & 0.37 & 0.33 \\
\midrule
Lookahead order & 0.67 & 0.44 & 0.53 & 0.44 & -- \\
\bottomrule
\end{tabular}
\end{table}

\begin{table}[t]
\caption{Number of evaluations until a best decomposition is evaluated when each evaluation may convexify at most $s$ families, on the instances with a true joint support and a positive gain within the budget, mean $\pm$ standard error. The lowest mean of each column is in bold. Greedy selection stops at the LP relaxation on every template and on most MIPLIB-NL models and is not shown. The five-dual lookahead order and the interleaved orders: \cref{app:results}.}
\label{tab:search}
\centering
\small
\setlength{\tabcolsep}{5pt}
\begin{tabular}{@{}lccc@{}}
\toprule
Order & $s=2$ & $s=3$ & $s=k-1$ \\
\midrule
\multicolumn{4}{@{}l}{\emph{Templates}} \\
Random (expected) & 4.57 $\pm$ 0.77 & 4.56 $\pm$ 0.82 & 3.82 $\pm$ 0.37 \\
Covering & 2.86 $\pm$ 0.61 & 2.33 $\pm$ 0.31 & 1.82 $\pm$ 0.08 \\
Lookahead & \textbf{2.00} $\pm$ 0.76 & 1.92 $\pm$ 0.71 & 1.92 $\pm$ 0.71 \\
JointDW with Claude Sonnet 5.5 & 2.47 $\pm$ 0.67 & 2.11 $\pm$ 0.41 & 1.71 $\pm$ 0.22 \\
JointDW with GPT-6.1 Sol & 2.07 $\pm$ 0.67 & \textbf{1.72} $\pm$ 0.47 & \textbf{1.16} $\pm$ 0.14 \\
JointDW with Grok 4.20 (Reasoning) & 2.48 $\pm$ 0.67 & 2.10 $\pm$ 0.43 & 1.59 $\pm$ 0.18 \\
JointDW with Grok 4.20 (Non-Reasoning) & 2.92 $\pm$ 0.64 & 2.68 $\pm$ 0.48 & 2.16 $\pm$ 0.23 \\
JointDW with DeepSeek-V4-Pro (Non-Thinking) & 3.14 $\pm$ 0.55 & 2.69 $\pm$ 0.31 & 2.14 $\pm$ 0.16 \\
\multicolumn{4}{@{}l}{\emph{MIPLIB-NL}} \\
Random (expected) & 4.82 $\pm$ 0.59 & 5.70 $\pm$ 0.80 & 5.96 $\pm$ 1.02 \\
Covering & 3.08 $\pm$ 0.44 & 2.84 $\pm$ 0.40 & 2.11 $\pm$ 0.11 \\
Lookahead & 2.18 $\pm$ 0.78 & 2.12 $\pm$ 0.78 & 2.29 $\pm$ 0.94 \\
JointDW with Claude Sonnet 5.5 & 2.20 $\pm$ 0.34 & 2.26 $\pm$ 0.36 & 1.91 $\pm$ 0.25 \\
JointDW with GPT-6.1 Sol & \textbf{1.79} $\pm$ 0.34 & \textbf{1.99} $\pm$ 0.44 & \textbf{1.78} $\pm$ 0.31 \\
JointDW with Grok 4.20 (Reasoning) & 2.42 $\pm$ 0.46 & 2.32 $\pm$ 0.40 & 1.90 $\pm$ 0.30 \\
JointDW with Grok 4.20 (Non-Reasoning) & 2.76 $\pm$ 0.48 & 2.75 $\pm$ 0.49 & 2.04 $\pm$ 0.25 \\
JointDW with DeepSeek-V4-Pro (Non-Thinking) & 2.51 $\pm$ 0.34 & 2.51 $\pm$ 0.34 & 2.08 $\pm$ 0.23 \\
\bottomrule
\end{tabular}
\end{table}

\begin{figure}[t]
\centering
\includegraphics{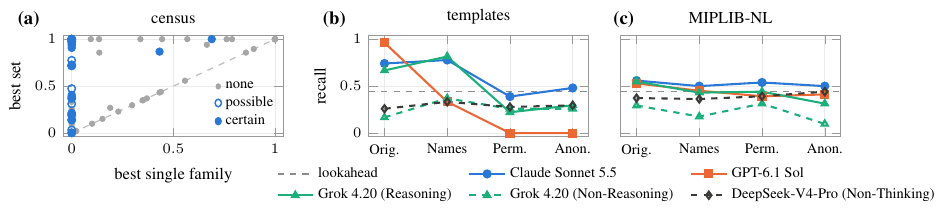}
\caption{(a) Census of MIPLIB-NL: for each model with a positive gain, the largest gain of a single family against the largest gain of a set of families, both as shares of the LP gap; marks on the vertical axis have no family that strengthens the bound alone. Blue marks have a zero-marginal group at $\emptyset$, certain (filled) or possible (open); gray marks have none. (b, c) Recall of joint supports under the four presentations, with the recall of the lookahead order as a dashed line. Solid lines: configurations with reasoning; dashed lines: without.}
\label{fig:results}
\end{figure}

\paragraph{Groups that gain only jointly are common.}
Three of ten MIPLIB-NL models whose bound can be improved have a certain zero-marginal group at $\emptyset$, and another quarter have only a possible one (\cref{app:miplib}). On most of these models, no single family strengthens the bound at all, so their marks in \cref{fig:results}(a) lie on the vertical axis. Greedy selection misses the best decomposition on all but three of the scored models with a true joint support. Every benchmark instance has a certain zero-marginal group at $\emptyset$, the classical pair of its class. On the network-design and facility-location instances, GCG's detector follows the block structure of the matrix and convexifies one block per arc or per facility, a set with no gain on 38 of these 39 instances (\cref{tab:app-bench}). The phenomenon of \cref{fig:lattice} is therefore not confined to textbook instances.

\paragraph{What the language models know.}
On the structured templates, GPT-6.1 Sol names a true joint support first on every instance, and Claude Sonnet 5.5 and Grok 4.20 (Reasoning) do so on about two-thirds of them (\cref{tab:llm}). On MIPLIB-NL, these three configurations are close to each other and to the lookahead order. The two configurations without reasoning fall behind, and for Grok 4.20 the difference between its two variants exceeds sampling variation. GPT-6.1 Sol and both Grok 4.20 configurations never predict a joint support on the assignment templates, which have none, and no configuration finds the hidden group of a program of \cref{thm:cover}(a) more often than chance. Given every coefficient of these programs, Claude Sonnet 5.5 names the hidden group first in every answer, and GPT-6.1 Sol does so on the three programs for which it returns an answer (\cref{app:results}).

The language models and the lookahead make correct predictions on partly different sets of models. On the 17 MIPLIB-NL models with a certain joint group, each configuration with reasoning names the true group first on two or three models where the first set of the lookahead order is wrong. On each of these models, it does so in at least two of its three samples. The lookahead is right on two models where that configuration is wrong. Averaged over the samples, Claude Sonnet 5.5 or the lookahead names a true group first on 0.73 of these models, against 0.57 and 0.53 for each alone (\cref{tab:app-first}). Most misses of the lookahead are the superset effect of \cref{prop:lookahead}, a true group with further families. On the network-design templates, the lookahead at the HiGHS dual is zero on every pair, including the true one, while GPT-6.1 Sol names the true pair first on each of them. This zero is the dependence on the dual that \cref{prop:lookahead} allows for polyhedral groups. With five optimal duals, the lookahead names the true group first on every template, and on MIPLIB-NL its hit rate does not change (\cref{tab:app-five}).

\paragraph{Where the predictions come from.}
On the templates, the predictions follow the names: GPT-6.1 Sol finds no joint support once the names are permuted or anonymized, and permuting the names lowers the recall of Grok 4.20 (Reasoning) and Claude Sonnet 5.5 beyond sampling variation (\cref{fig:results}(b)). MIPLIB-NL names are less standard, and permuting them lowers no recall there beyond sampling variation (\cref{fig:results}(c)). These predictions therefore do not rest on the name attached to each family. Withholding the names and the text together still lowers the recall of both Grok 4.20 variants. Grok 4.20 (Reasoning) names a true group first about twice as often as Grok 4.20 (Non-Reasoning) on both sets of instances, and on MIPLIB-NL its higher hit rate and recall persist when the names are anonymized (\cref{tab:app-reason}).

\paragraph{What the predictions are worth.}
On the templates and on MIPLIB-NL, every order other than the random one needs about three evaluations or fewer on average, because most of these models have at most four families. Among the orders of \cref{tab:search}, JointDW with GPT-6.1 Sol is the cheapest on MIPLIB-NL for every $s$. On MIPLIB-NL, JointDW needs fewer evaluations than the covering order beyond sampling variation for $s=2$ with every configuration except Grok 4.20 (Non-Reasoning), and the interleaved order does so for every configuration and every $s$ (\cref{tab:app-paired-search}). There, every configuration matches the lookahead order at the HiGHS dual within sampling variation, while on the templates the interleaved orders of the two configurations without reasoning cost about half an evaluation more than it. More optimal duals help every configuration. With five of them, the lookahead order needs close to one evaluation for $s=2$ on MIPLIB-NL and exactly one on every template. JointDW with GPT-6.1 Sol matches it within sampling variation, and interleaving brings every configuration within one evaluation of it (\cref{tab:app-five}). GCG's own decompositions, used as anchors of the staged search, are no better than the covering order on the rich models (\cref{app:miplib,tab:rich-search}).

\begin{figure}[t]
\centering
\includegraphics[scale=1.25]{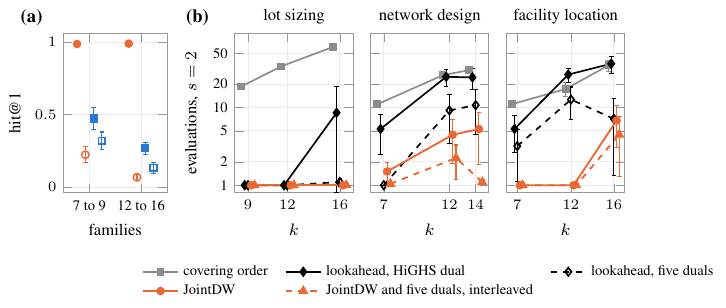}
\caption{Rich models with 7 to 16 families. (a) hit@1 of GPT-6.1 Sol (circles) and Claude Sonnet 5.5 (squares) given the text (open) and every coefficient (filled), on the models with 7 to 9 and with 12 to 16 families. (b) Mean number of evaluations until a best set is evaluated for $s=2$ against the number of families $k$, by class, with JointDW using GPT-6.1 Sol given every coefficient. Bars: one standard error. Every configuration and $s=3$: \cref{tab:rich-search,tab:app-rich-large}.}
\label{fig:rich}
\end{figure}

\paragraph{Rich models.}
On the rich models, a covering of all pairs needs more evaluations as the number of families grows, from 11 with seven families to as many as 60 with 16 (\cref{fig:rich}(b)). The lookahead order at the HiGHS dual finds a best pair with its first set on every lot-sizing model with at most 12 families. On network design and facility location with 12 or more families, it needs about as many evaluations as the covering order, and five optimal duals lower its cost there to between 7 and 13 evaluations. Given the text, no configuration names a true group first in more than a third of its answers (\cref{fig:rich}(a)). On the models with seven to nine families, GPT-6.1 Sol mostly returns no group and states that the coefficients are missing, Claude Sonnet 5.5 often names another classical pair, Grok 4.20 (Reasoning) adds further families to the group, and the two configurations without reasoning name pairs that are not groups.

Given every coefficient, GPT-6.1 Sol names a true group first in 249 of its 252 answers on the rich models, and JointDW then needs between one and seven evaluations on average for pairs in every class and size. Its hit rate does not fall as the number of families grows from 7 to 16 (\cref{tab:rich-pred,tab:app-rich-large}). Interleaved with the five-dual lookahead order, it needs 1.8 evaluations for pairs on the models with 12 to 16 families, fewer than the five-dual order beyond sampling variation. Pooled over these models, no other configuration or order improves on the five-dual order beyond sampling variation (\cref{tab:app-rich-large,tab:app-rich-five}). For sets of three families, the lookahead order alone needs one evaluation on the models with seven to nine families. On those with 12 to 16 families, pooled over the models, the orders with GPT-6.1 Sol given every coefficient do not differ from the five-dual order beyond sampling variation. The lookahead and the learned predictor read the same instance data, the first through the LP and its pricing problems and the second through structural features of each decomposition.

The prediction from the coefficients does not depend on the family names. On the models with seven to nine families, with the family names permuted among the families or replaced by labels, GPT-6.1 Sol still names a true group first in all but one answer. The pair with the classical names after the permutation is never a true group and is never named first. The hit rate of Claude Sonnet 5.5 falls to about a quarter of its value under the permutation (\cref{tab:app-control}).

\begin{figure}[t]
\centering
\includegraphics[scale=1.25]{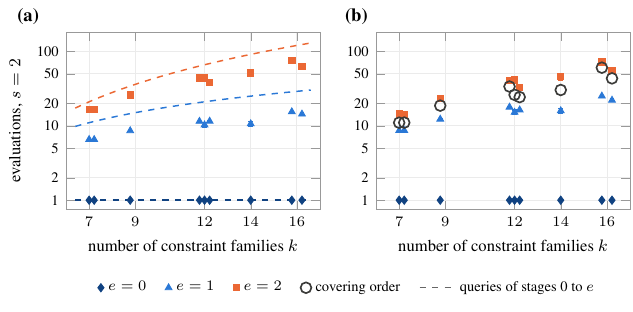}
\caption{Search cost on the rich models for $s=2$, by the number $e$ of members of a best pair that the anchor misses, with one mark per class and number of families $k$. (a) The staged order $O_1$ alone. Dashed lines of the same color: the number of queries of stages $0$ to $e$ of \cref{thm:pred}(a). (b) JointDW and the covering order. Predictions of GPT-6.1 Sol and Claude Sonnet 5.5 under both presentations whose anchor has at most two families. Bars: one standard error.}
\label{fig:misses}
\end{figure}

\paragraph{Search cost and the number of misses.}
\Cref{thm:pred} ties the cost of the staged search to the number $e$ of members that the prediction misses. \Cref{fig:misses} shows this relation on the rich models for pairs, with $e$ calculated on the anchor that the search uses. A prediction without a miss costs one evaluation in every case. With one miss, the staged order stays below the number of queries of its first two stages. JointDW needs about four-fifths as many evaluations as the covering order with seven families, and about half with 16. With both members missed, JointDW needs about 1.3 times the evaluations of the covering order. The bound of \cref{thm:pred}(a) holds for every prediction whose anchor has at least as many families as the best set (\cref{tab:app-misses}). For $s=3$, some best sets have three families while the anchor is a pair. The staged order alone then reaches a best set only in some orders of the families, and JointDW reaches one in every case through $O_2$, at less than three-quarters of the cost of the covering order (\cref{tab:app-larger}).

\paragraph{A predictor trained on labeled evaluations.}
The learned predictor is trained on the gain tables of other instances, so each training instance costs every evaluation in its table. When it is trained on the other seven rich models with seven to nine families of its class, the predictor's cost for $s=2$ is within sampling variation of that of JointDW with GPT-6.1 Sol given the coefficients, and its labels cost 196 to 315 evaluations per class (\cref{tab:app-learned}). Trained instead on the other two classes and on MIPLIB-NL, the predictor names the best pair first on a third of these rich models and needs about five evaluations more than JointDW, beyond sampling variation. With the lookahead value as a feature, the predictor is within sampling variation of JointDW with GPT-6.1 Sol and of the five-dual lookahead order. On MIPLIB-NL, trained on the gain tables of the other scored models, the predictor with the lookahead feature matches the five-dual lookahead order, at more than 800 labeled evaluations per held-out model. The language model needs no labels.

\begin{table}[t]
\caption{Benchmark instances. Left: share of the 59 instances on which the first set of a method is the best pair, for lot sizing (LS), network design (ND), facility location (FL), and all instances. Middle: share of the 59 instances on which it is the time-optimal set, and its time ratio. Right: the same two quantities on the 76 generalized assignment and bin-packing instances, which have no joint support within the budget. JointDW uses the original presentation. Time ratios: geometric mean $\pm$ standard error, with an asterisk when the 95 percent interval of the mean log ratio excludes zero. Per class and paired differences: \cref{tab:app-bench,tab:app-bench-time,tab:app-nojoint}.}
\label{tab:bench}
\centering
\small
\setlength{\tabcolsep}{4pt}
\begin{tabular}{@{}lcccccccc@{}}
\toprule
& \multicolumn{4}{c}{first set is the best pair} & \multicolumn{2}{c}{time-optimal set} & \multicolumn{2}{c}{no joint support} \\
\cmidrule(lr){2-5}\cmidrule(lr){6-7}\cmidrule(l){8-9}
Method & LS & ND & FL & all & hit & time ratio & hit & time ratio \\
\midrule
Lookahead (HiGHS dual) & 1.00 & 0.35 & 0.68 & 0.68 & 0.56 & 2.03 $\pm$ 0.86$^{*}$ & 0.76 & 2.58 $\pm$ 1.16$^{*}$ \\
Lookahead (five duals) & 1.00 & 0.35 & 0.95 & 0.76 & 0.61 & 1.71 $\pm$ 0.62$^{*}$ & 0.79 & 2.56 $\pm$ 1.15$^{*}$ \\
GCG's detector & 1.00 & 0.00 & 0.00 & 0.34 & 0.32 & 2.30 $\pm$ 0.81$^{*}$ & 0.79 & 1.18 $\pm$ 0.13$^{*}$ \\
\multicolumn{9}{@{}l}{\emph{JointDW with}} \\
GPT-6.1 Sol & 1.00 & 0.67 & 1.00 & 0.89 & 0.75 & 1.29 $\pm$ 0.20$^{*}$ & 0.96 & 1.00 $\pm$ 0.00 \\
Claude Sonnet 5.5 & 1.00 & 0.75 & 0.68 & 0.81 & 0.73 & 1.36 $\pm$ 0.24$^{*}$ & 0.75 & 1.23 $\pm$ 0.20 \\
\bottomrule
\end{tabular}
\end{table}

\paragraph{Benchmark instances.}
On the benchmark instances, which have three families each, every method proposes one set (\cref{tab:bench}). On network design, the best pair is flow balance with arc capacity. With either one or with five duals, lookahead names it on about a third of the instances, because these duals give a larger value to flow balance with arc linking or to arc capacity alone. The two language models name it on most instances. On facility location, GPT-6.1 Sol names the best pair, assignment with capacity, on every instance. Pooled over the instances, both language models identify the best pair more often than the detector, beyond sampling variation, while their differences from the two lookahead orders are within sampling variation (\cref{tab:app-bench-paired}). With every coefficient shown instead of the family tables, in prompts of tens to hundreds of thousands of tokens, both configurations name the best pair less often. The first set of GPT-6.1 Sol still gains more of the LP gap than that of the lookahead at the HiGHS dual and that of the detector, beyond sampling variation (\cref{tab:app-bench-data}).

\paragraph{Solve time.}
Within the budget of two families, the best pair is also the time-optimal set on 51 of the 59 instances. Its pricing problems take most of the solve time, yet its time statistic is 3 to 40 times smaller than that of a single family set, depending on the class (\cref{tab:app-tradeoff}). Convexifying all three families makes the pricing problem equivalent to the original program, which SCIP solves in less than a minute on each of these instances. As in the census, the budget excludes this set. The sets of the two language models are the time-optimal set on about three-quarters of the instances. In geometric mean, they take 1.3 and 1.4 times its time, against 1.7 times for the five-dual lookahead and 2.3 times for GCG's detector (\cref{tab:bench}). Both language models choose faster sets than the detector, beyond sampling variation, while their differences from the two lookahead orders remain within sampling variation (\cref{tab:app-bench-time}).

On the generalized assignment and bin-packing instances, the capacity family alone is the time-optimal set on every instance. GPT-6.1 Sol names no joint support there and selects the capacity family in every answer that proposes a set. Its set is therefore time-optimal on 96 percent of these instances, and on bin packing it takes less than half the time of GCG as shipped. Claude Sonnet 5.5 names assignment with capacity in a quarter of its answers, thereby convexifying the program or all but its symmetry-breaking rows. On bin packing, the lookahead orders choose this pair or the assignment family, and their sets take 86 times as long to solve as the time-optimal set (\cref{tab:app-nojoint}).

\paragraph{What a prediction costs.}
One answer of GPT-6.1 Sol to the original prompt takes 32 to 50 seconds and costs one to two cents at list prices. One answer of Claude Sonnet 5.5 takes between 2.6 and 24 minutes and costs about 20 cents (\cref{tab:app-cost}). A prediction therefore saves time only where an evaluation is slow. On MIPLIB-NL, one evaluation in the census setting takes 43 seconds on average. On MIPLIB-NL, JointDW with GPT-6.1 Sol saves 1.3 evaluations relative to the covering order for $s=2$, so its prediction pays for its latency when an evaluation takes more than 28 seconds. Against the five-dual lookahead order, it saves no evaluation beyond sampling variation. On the templates and the rich models, an evaluation takes a tenth of a second, so even saving tens of evaluations does not offset the prediction latency.

The exponent $d$ of \cref{thm:costMain} can be estimated as the slope of the logarithm of the evaluation time against the logarithm of the number of convexified families. Its median is 1.6 over the 41 MIPLIB-NL models with evaluations of at least three sizes, and it lies between 0.02 and 0.50 on the templates and the rich models (\cref{tab:app-growth}). For a group of two families predicted without a miss, MIPLIB-NL thus lies in the range $e<d<w$ of \cref{thm:costMain}.

%%%%%%%%%%%%%%%%%%%%%%%%%%%%%%%%%%%%%%%%%%%%%%%%%%%%%%%%%%%%%%%%%%%%%%%%%%%%%%%%%%%%%%%%%%%%%%%%%%%%%%%%%%%%%%%%
\section{Discussion}\label{sec:discussion}

\paragraph{Implications for automatic decomposition.}
A detector that enumerates subsets of constraint classes and keeps the best bound, as GCG's class-based detector does, pays the covering number of \cref{thm:cover} in the worst case. A detector that follows the block structure of the matrix convexifies, on two of the three benchmark classes, a set without gain on all but one instance. Information from outside the bound evaluations, such as the constraint matrix, an LP dual, or the text of the model, reduces this cost, and \cref{thm:pred} prices it by its number of misses. The LP dual, the text of the model, and labeled evaluations of similar instances fail on different models and have different costs. The dual is the cheapest, and with several optimal duals it is at least as good as any configuration on MIPLIB-NL. On the benchmark network-design instances, however, it values flow balance with arc linking above the best pair. The text with its coefficients is strongest where the dual is degenerate and where many families make a covering large. One request takes from half a minute to more than 20 minutes, so it pays only where evaluations are slow. Labeled evaluations buy accuracy only within the class or collection they come from. Because the alternation delays each of two orders by at most a factor of two, a detector can combine these sources, and a more accurate language model or a better dual can replace a current one at a loss of at most this factor.

\paragraph{Scope.}
The oracle model counts evaluations and not seconds, and its randomized and asymptotic statements use the cardinality budget $s$ or, in \cref{thm:costMain,thm:priceSearch,thm:overlap}, a cost that grows as a power of the number of convexified families. The MIPLIB-NL models with a complete gain table have few families, so the savings of \cref{thm:pred} are measured on the rich models, which have up to 16 families. The rich models are constructed from three problem classes and drawn at random for this study, so a language model can know their class but not their data, and all their groups are connected through shared variables (\cref{prop:lookahead}). The MIPLIB-NL models and the benchmark instances are public. On MIPLIB-NL, the controlled presentations separate knowledge of a class from familiarity with its names. The benchmark instances are small enough that SCIP alone solves each of them in less than a minute, on most of them faster than branch-and-price on the best pair. They measure what a choice of decomposition costs within branch-and-price, not the value of branch-and-price on these classes.

\section{Conclusion}\label{sec:conclusion}

Choosing which constraint families to convexify is hard when families strengthen the bound only together. Bound evaluations then reveal little, and finding a group costs a covering number of evaluations. A prediction with $e$ misses reduces the cost to $\Theta(k^e)$ without an asymptotic price in robustness, so the evaluations a prediction saves measure what the predictor knows. The same orders hold with several groups and with prices, also with overlapping gains when no group contains another, and an evaluation cost equal to the $d$-th power of the number of convexified families turns them into $\Theta(k^{\min(w,d)})$ and $\Theta(k^{\min(e,d)})$. Overlapping gains make the price-optimal combination of known groups NP-hard. Such groups occur in three of ten MIPLIB-NL models whose bound can be improved and in every lot-sizing, network-design, and facility-location benchmark instance we examined. On most of these benchmark instances, the best pair is also the fastest set of at most two families for branch-and-price. Whether a language model knows which constraints belong together depends on what it reads. From the text, language models know the groups that textbooks describe, on textbook models through their names, and the strongest of them identify the groups about as often as the lookahead at an LP dual. Given the coefficients, one of them names a true group first in almost every answer on constructed models with up to 16 families, on the smaller models also with the names permuted or removed, and the evaluations it saves grow with the number of families. None of them needs labeled evaluations. A prediction costs seconds to minutes of latency, so it pays where evaluations are slow. Because the search turns every avoided miss into fewer evaluations and bounds the loss from a wrong prediction, a more accurate language model lowers the cost of choosing a decomposition.

\bibliographystyle{plainnat}
\bibliography{refs}

\newpage
\appendix
\counterwithin{theorem}{section}\counterwithin{proposition}{section}\counterwithin{lemma}{section}\counterwithin{example}{section}
\raggedbottom
\setcounter{topnumber}{4}\setcounter{bottomnumber}{4}\setcounter{totalnumber}{8}
\renewcommand{\topfraction}{0.95}\renewcommand{\bottomfraction}{0.95}\renewcommand{\textfraction}{0.05}\renewcommand{\floatpagefraction}{0.8}

\FloatBarrier
\section{Proofs}\label[appendix]{app:proofs}

Throughout, $\LP=\{x : A_fx\ge b_f\ \text{for}\ f\in F,\ \ell\le x\le u\}$, and for multipliers $y\ge0$ on the rows of $F\setminus T$,
\[
L_T(y)=y\,b_{F\setminus T}+\min\{(c-yA_{F\setminus T})x : x\in X_T\}.
\]

\begin{lemma}[monotonicity]\label{lem:mono}
If $T\subseteq T'$, then $P_{T'}\subseteq P_T$. Hence $g$ is nondecreasing and $0\le g(T)\le z_{\IP}-z_{\LP}$.
\end{lemma}
\begin{proof}
$X_{T'}\subseteq X_T$, so $\conv(X_{T'})\subseteq\conv(X_T)$, and $P_F=\conv(X)$.
\end{proof}

\begin{lemma}[inheritance]\label{lem:inherit}
$z(T)=\max_{y\ge0}L_T(y)$. For $T\subseteq T'$ and every $y\ge0$ on the rows of $F\setminus T$, $L_{T'}(y|_{F\setminus T'})\ge L_T(y)$. Solving the blocks of $T'$ once at optimal multipliers of $T$ therefore gives a valid bound of at least $z(T)$.
\end{lemma}
\begin{proof}
The first statement is Lagrangian duality over the polytope $\conv(X_T)$, on which a linear objective attains its minimum over $X_T$. For $x\in X_{T'}$, the rows of $T'\setminus T$ hold, so $y_{T'\setminus T}(A_{T'\setminus T}x-b_{T'\setminus T})\ge0$. This gives
\[
(c-y_{F\setminus T'}A_{F\setminus T'})x+y_{F\setminus T'}b_{F\setminus T'}\ge(c-yA_{F\setminus T})x+y\,b_{F\setminus T}.
\]
Minimize the left side over $X_{T'}$ and the right side over $X_T\supseteq X_{T'}$.
\end{proof}

\begin{proposition}[support view]\label{prop:support}
For every $T\subseteq F$, $P_T$ is the set of points of $\LP$ that satisfy every inequality $\alpha x\ge\beta$ valid for $\conv(X)$ that has a family support $S\subseteq T$.
\end{proposition}
\begin{proof}
An inequality with a family support $S\subseteq T$ is valid for $P_S\supseteq P_T$ by \cref{lem:mono}. Conversely, $\conv(X_T)$ is the intersection of its valid inequalities. Each of them is valid for $\conv(X)\subseteq\conv(X_T)$ and for $P_T$, so it has a family support inside $T$.
\end{proof}

\paragraph{Polyhedral groups.} Let $W$ be a polyhedral zero-marginal group at $B$. Since $Q_W\subseteq P_B$, $\min\{cx:x\in Q_W\}=z(B)<z(B\cup W)$, so a point $\bar x\in Q_W$ with $c\bar x=z(B)$ does not lie in $P_{B\cup W}$. By \cref{prop:support}, some valid inequality with a family support inside $B\cup W$ is violated by $\bar x$. It has no family support inside any $B\cup S$ with $S\subsetneq W$, since it would then be valid for $P_{B\cup S}\supseteq Q_W$. Every family support of it inside $B\cup W$ therefore contains $W$. At $B=\emptyset$, the point $\bar x$ is an optimum of the LP relaxation, so $W$ is a joint support of an inequality that cuts off an LP optimum.

\paragraph{Interleaving.} Let $O_1$ and $O_2$ be two query orders, and let the interleaved order take their entries alternately, skipping sets already queried. The first $t$ entries of either order appear among the first $2t$ entries of the interleaved order. If $O_i$ queries a set of maximal gain at position $t_i$, the interleaved order does so within $2\min(t_1,t_2)$ queries.

\begin{proof}[Proof of \cref{thm:cover}]
(a) Let the cycle have length $2n+1\ge\max(3,w)$, and let its edges be split into $w$ nonempty paths, one for each family of $W$; each other family is a path on its own vertices. If $W\not\subseteq T$, the rows of $T$ form a forest, whose edge-vertex incidence matrix is totally unimodular, so $\conv(X_T)$ equals the LP relaxation of $X_T$ and $P_T=\LP$. If $W\subseteq T$, the cycle variables range over the stable-set polytope of the cycle, on which $\max\sum_v x_v=n$, against the LP value $(2n+1)/2$. The path components are integral in both cases and share no variable with the cycle, so $g(T)=\tfrac12[W\subseteq T]$.

(b) On $g_W$, a query returns a nonzero value exactly when it contains $W$, and the maximal gain $\tfrac12$ is attained exactly by these queries. Against the answer $0$ to every query, a deterministic method produces a fixed sequence $T_1,T_2,\dots$. If $T_1,\dots,T_{N-1}$ with $N=N(\cT,w)$ leave uncovered some $w$-set $W$ that lies in a member of $\cT$, then on $g_W$ the method has not found a maximizer after $N-1$ queries. Conversely, querying a minimum covering finds every such $W$. For the randomized bound, let $W$ be uniform over the $w$-subsets and $J=\binom{k}{w}/\binom{s}{w}$. The first $j$ queries of a deterministic method contain at most $j\binom{s}{w}$ of the $\binom{k}{w}$ targets, so its number of queries $N_W$ satisfies $\Pr[N_W>j]\ge1-j/J$ and
\[
\mathbb{E}N_W=\sum_{j\ge0}\Pr[N_W>j]\ge\sum_{j=0}^{\lfloor J\rfloor}(1-j/J)\ge(\lfloor J\rfloor+1)/2>J/2.
\]
A randomized method is a distribution over deterministic ones, so its expected cost on the uniform target exceeds $J/2$, and the maximum over $W$ is at least this average.

(c) Each $s$-set contains $\binom{s}{w}$ of the $\binom{k}{w}$ sets of size $w$, which gives the lower bound. The uniform fractional covering has value $\binom{k}{w}/\binom{s}{w}$, and the greedy covering is within a factor $1+\ln\binom{s}{w}$ of the fractional optimum \citep{lovasz1975ratio,stein1974two}. R\"odl's theorem \citep{rodl1985packing} gives the asymptotics. For unknown $w$, the additional terms are $\sum_{w'<w}\cov(k,s,w')=O(k^{w-1})$.

(d) Every single addition has marginal gain $0$ on $g_W$ because $w\ge2$.
\end{proof}

\begin{proof}[Proof of \cref{thm:pred}]
(a) Let $W$ be a hidden group with $|W\setminus P|=e$, and let $\mathcal{K}$ be the $e$-cover of stage $e$. The set $W\cap P$ has $w-e$ elements, so it lies in some $A\in\mathcal{K}$, and $W\setminus P$ lies in some set $R$ of the covering taken for $A$. The query $A\cup R$ has at most $|A|+(s-|A|)=s$ families and contains $W$. Every $0$-cover has $P$ as a member, so the one with the fewest queries is $\{P\}$, and stage $0$ is the query $P$.

(b) As in \cref{thm:cover}(b), a deterministic method that knows $e$ queries a fixed sequence of sets while every answer is $0$, and it has queried a set of maximal gain on every hidden group with $e$ misses only when these queries together contain every such group. Let $\mathcal{Q}$ be a collection of queries with this property. Fix a set $Y$ of $e$ families outside $P$, and let $\mathcal{Q}_Y$ be the queries of $\mathcal{Q}$ that contain $Y$. For every $(w-e)$-subset $X$ of $P$, the set $X\cup Y$ is a hidden group with $e$ misses, and a query that contains it belongs to $\mathcal{Q}_Y$. Every $T\in\mathcal{Q}_Y$ has $|T\cap P|\le s-e$. The sets $T\cap P$ with $T\in\mathcal{Q}_Y$ therefore form an $e$-cover, and $\sum_{T\in\mathcal{Q}_Y}\binom{s-|T\cap P|}{e}^{-1}$ is at least its weight, hence at least $v_e$. A query $T$ contains $\binom{|T\setminus P|}{e}\le\binom{s-|T\cap P|}{e}$ of the sets $Y$. Summing over the $\binom{k-w}{e}$ sets $Y$ gives
\[
v_e\binom{k-w}{e}\;\le\;\sum_{Y}\,\sum_{T\in\mathcal{Q}_Y}\binom{s-|T\cap P|}{e}^{-1}
=\sum_{T\in\mathcal{Q}:\,|T\setminus P|\ge e}\binom{|T\setminus P|}{e}\binom{s-|T\cap P|}{e}^{-1}\;\le\;|\mathcal{Q}|.
\]
For the randomized bound, the hidden groups with $e$ misses number $\binom{w}{e}\binom{k-w}{e}$. A query $T$ with $a=|T\cap P|$ and $|T\setminus P|\le s-a$ contains at most $f_e(a)$ of them, and none if $a\notin A_e$, so \cref{thm:multi}(e) with $m=1$, where $D_1\le\max_{a\in A_e}f_e(a)$, gives more than $L_e/2$. A member of an $e$-cover with $a\in A_e$ elements contains $\binom{a}{w-e}$ of the $\binom{w}{e}$ subsets to be covered and has weight $1/\binom{s-a}{e}$, and a member with fewer than $w-e$ elements contains none of them. The weight of every $e$-cover is therefore at least $\binom{w}{e}/\max_{a\in A_e}f_e(a)$, which gives $L_e\le v_e\binom{k-w}{e}$. For $e=0$, $A_0=\{w\}$ and $v_0=1$. For $e\ge1$, the sequence $f_e$ is log-concave on $w-e\le a\le s-e$ as a product of log-concave sequences, and
\[
\frac{f_e(w)}{f_e(w-1)}=\frac{w\,(s-w+1-e)}{e\,(s-w+1)}\ge1 \iff e(s+1)\le w(s-w+1).
\]
Since $e$ is an integer and $e(s+1)\le w(s-w+1)<(s+1)(s-w+1)$, the condition implies $e\le s-w$, so $w\in A_e$, and the maximum is at $a=w$, where $f_e(w)=\binom{w}{e}\binom{s-w}{e}$. The $e$-cover $\{P\}$ then has the weight $1/\binom{s-w}{e}$ of the bound. For $e=w$, $f_w(a)=\binom{s-a}{w}$ decreases in $a$, the maximum is at $a=0$, and the $w$-cover $\{\emptyset\}$ has weight $1/\binom{s}{w}$.

(c) Let $e'\ge1$, and let $\mathcal{K}$ be an $e'$-cover. For each $A\in\mathcal{K}$, R\"odl's theorem gives $\cov(k-w,s-|A|,e')=(1+o(1))\binom{k-w}{e'}/\binom{s-|A|}{e'}$, so the number of queries of stage $e'$ with $\mathcal{K}$ is $(1+o(1))\binom{k-w}{e'}$ times the weight of $\mathcal{K}$. The $e'$-covers depend only on $(s,w,e')$ and are finite in number, so the stage with the fewest queries has $(1+o(1))\,v_{e'}\binom{k-w}{e'}=\Theta(k^{e'})$ queries, and the sum over stages $0$ to $e$ is $(1+o(1))$ times its last term. For $e=0$, the cost is $1=v_0$. For $e=w$, the $w$-cover $\{\emptyset\}$ gives a stage of $\cov(k-w,s,w)\le\cov(k,s,w)$ queries, and the earlier stages cost $O(k^{w-1})$.
\end{proof}

\paragraph{The linear relaxation.} Let $k\ge s+w$, and let $a$ maximize $f_e$ over $A_e$. Every hidden group with $e$ misses lies in the same number of the queries $T$ with $|T\cap P|=a$ and $|T\setminus P|=s-a$, and each of these queries contains $f_e(a)$ hidden groups. Equal weights on these queries that cover every hidden group once have total weight $L_e$, and no fractional covering of the hidden groups by queries has less, since no query contains more than $f_e(a)$ of them. The bound $v_e\binom{k-w}{e}$ applies to coverings by whole queries.

\paragraph{Finite $k$.} With $w=2$, $s=3$, and $k=20$, an entirely wrong prediction costs at most $1+18+\cov(18,3,2)=73$ queries with minimum coverings, against $\cov(20,3,2)=67$ without a prediction \citep{fort1958minimal}. The values plotted in \cref{fig:cost}(b) use, in each stage, the cover of $P$ with the fewest queries among those whose members have one size, and greedy coverings: at each step, a greedy covering takes the first set in lexicographic order that covers the most uncovered subsets. A greedy covering exceeds a minimum covering by at most a factor that depends only on $s$ and $w$ \citep{lovasz1975ratio,stein1974two}, so the staged search with greedy coverings keeps the order $\Theta(k^e)$ of \cref{thm:pred}(c). For some $k$, the greedy covering is a Steiner system, such as $S(3,6,22)$ with $77=\binom{22}{3}/\binom{6}{3}$ blocks without prediction at $k=22$ and the projective plane $S(2,5,21)$ in stage $2$ at $k=24$, so the plotted values are not monotone in $k$. The value of the staged search for $e$ misses is the number of queries in stages $0$ to $e$, an upper bound on its worst case, since a group can already lie in a query of an earlier stage. For $k=16$ and $e=3$, the staged search needs $56$ queries against $43$ for the greedy covering without prediction. The earlier stages cause this overhead, and the interleaved order of JointDW caps the cost at twice the smaller of the two.

\paragraph{The cover in stage $e$.} A minimum covering of the $(w-e)$-subsets of $P$ by sets of one size $a\in A_e$ has weight $c_e(a)=\cov(w,a,w-e)/\binom{s-a}{e}$. For $(w,s,e)=(4,7,3)$, the condition of \cref{thm:pred}(b) fails, $A_3=\{1,2,3,4\}$, and $c_3(a)$ takes the values $4/20$, $2/10$, $2/4$, and $1/1$ for $a=1,2,3,4$. Keeping all of $P$, which is $a=4$, is therefore five times costlier than the best size, and $v_3=1/5$ equals $\binom{4}{3}/\max_af_3(a)$, so the two bounds of \cref{thm:pred}(b) agree. Members of several sizes can weigh less. For $(w,s,e)=(3,6,2)$, the case of \cref{fig:cost}, the best single size has weight $3/10$. The cover that keeps two predicted families in one member and the third in another has weight $1/\binom{4}{2}+1/\binom{5}{2}=4/15=v_2$, against $\binom{3}{2}/\max_af_2(a)=1/4$ in $L_2$. The deterministic cost is then $(1+o(1))\tfrac{4}{15}\binom{k-3}{2}$, and $v_2\binom{k-3}{2}$ is $16/15$ of $L_2$. Among the 95 triples $(s,w,e)$ with $2\le w\le6$, $w\le s\le10$, and $1\le e<w$, the bound $v_e\binom{k-w}{e}$ exceeds $L_e$ in 16, by a factor of at most $4/3$, and members of one size attain $v_e$ in all but 8.

\begin{lemma}[one-pass lookahead]\label{lem:lookahead}
Let $y$ be an optimal dual solution of the LP relaxation for the rows of $F$, with the bounds $\ell\le x\le u$ kept, and let $\psi_y(T)=L_T(y|_{F\setminus T})-z_{\LP}$. Then $\psi_y(\emptyset)=0$, $\psi_y(T)\le g(T)$, $\psi_y$ is nondecreasing, and $\psi_y(T_1\cup T_2)=\psi_y(T_1)+\psi_y(T_2)$ whenever no variable appears in both a row of $T_1$ and a row of $T_2$. Consequently, $\psi_y(W)=0$ for every zero-marginal group $W$ at $\emptyset$ whose families pairwise share no variable, and every such group is LP-geometric.
\end{lemma}
\begin{proof}
LP duality with the bounds kept gives $L_\emptyset(y)=z_{\LP}$. Weak duality in \cref{lem:inherit} gives $L_T(y|_{F\setminus T})\le z(T)$, and \cref{lem:inherit} gives monotonicity. Write $\bar c=c-yA$. Then $L_T(y|_{F\setminus T})=yb-y_Tb_T+\min\{\bar cx+y_TA_Tx : x\in X_T\}$. If $T_1$ and $T_2$ share no variable, the minimum splits into three parts: one over the variables of $T_1$, one over the variables of $T_2$, and one over the remaining variables at their bounds. Subtracting $L_\emptyset(y)$ leaves the sum of the two increments. For the last claim, $|W|\ge2$ makes every $\{f\}$ a proper subset of $W$, so $0=\psi_y(\emptyset)\le\psi_y(W)=\sum_{f\in W}\psi_y(\{f\})\le\sum_{f\in W}g(\{f\})=0$. The rows of different families of $W$ constrain disjoint sets of variables, and the convex hull of a product of sets is the product of their convex hulls, so $\conv(X_W)=\bigcap_{f\in W}\conv(X_{\{f\}})$. Hence $P_W=\bigcap_{f\in W}P_{\{f\}}\supseteq Q_W\supseteq P_W$, and $\min\{cx:x\in Q_W\}=z(W)>z(\emptyset)$, so $W$ is LP-geometric.
\end{proof}

\paragraph{Greedy selection.} Let $\gamma$ be the submodularity ratio of \citet{das2011submodular} with respect to the set $U$ chosen by greedy selection. It is the minimum of $\sum_{x\in S}[g(L\cup\{x\})-g(L)]/[g(L\cup S)-g(L)]$ over $L\subseteq U$ and disjoint $S$ with $|S|\le s$ and $g(L\cup S)>g(L)$. A zero-marginal group $W$ at $\emptyset$ with $2\le|W|\le s$ gives $\gamma=0$ through $L=\emptyset$ and $S=W$.

\FloatBarrier
\section{Zero-Marginal Groups of Classical Formulations}\label[appendix]{app:classical}

For a set $T$ of families, a \emph{perspective block} of $T$ is a set of rows of $T$ whose variables are one binary variable $y$ and continuous variables $x_1,\dots,x_n$ with bounds $0\le x_j\le u_j$, none of which occurs in another row of $T$. Its rows have the form $Gx\le by$, where $Gx\le0$ and $x\ge0$ imply $x=0$.

\begin{lemma}[families without gain]\label{lem:nogain}
Let the rows of $T\subseteq F$ split into parts on pairwise disjoint sets of variables, each of which either contains no integer variable or is a perspective block of $T$. Then $\conv(X_T)$ is described by the rows of $T$, the bounds, and the inequalities $x_j\le u_jy$ for the variables of each perspective block. Hence $P_T=\LP$ and $g(T)=0$ whenever every point of $\LP$ satisfies these inequalities, in particular when $\LP$ contains a row $x_j\le M_jy$ with $M_j\le u_j$ for each of them.
\end{lemma}
\begin{proof}
The set $X_T$ is the product of the sets that the parts define on their variables with the bounds, and of the box of the variables in no row of $T$. The convex hull of a product is the product of the convex hulls. A part without integer variables defines a polyhedron, and the integer variables of the box have integral bounds. In a perspective block, $y=0$ forces $x=0$, so its set is $\{(0,0)\}\cup(K\times\{1\})$ with the polytope $K=\{x:Gx\le b,\ 0\le x\le u\}$. Its convex hull consists of $(0,0)$ and the points $(yx',y)$ with $x'\in K$ and $0<y\le1$, which is the set $\{(x,y):0\le y\le1,\ Gx\le by,\ 0\le x\le uy\}$. Every point of $\LP$ satisfies the rows of $T$ and the bounds, so $P_T=\LP$ when it also satisfies $x_j\le u_jy$, which follows from $x_j\le M_jy\le u_jy$.
\end{proof}

\begin{lemma}[pairs]\label{lem:pairs}
Let $f\ne f'$ be families with $P_{\{f\}}=P_{\{f'\}}=\LP$, and let $W=\{f,f'\}$. Then $W$ is a family support of every inequality that is valid for $P_W$ and violated by a point of $\LP$. If $g(W)>0$, then $W$ is a polyhedral zero-marginal group at $\emptyset$.
\end{lemma}
\begin{proof}
Such an inequality is valid for $\conv(X)\subseteq P_W$, and it is not valid for $P_S=\LP$ with $S\subsetneq W$, so $W$ is a minimal set of the up-set of \cref{def:support}. If $g(W)>0$, the proper subsets of $W$ have gain $0$, so $W$ is a zero-marginal group at $\emptyset$, and $Q_W=\LP$ has minimum $z(\emptyset)$, so the group is polyhedral.
\end{proof}

\begin{proposition}[classical formulations]\label{prop:classical}
Consider the formulations of \cref{app:templates}.
\begin{enumerate}[label=(\alph*),leftmargin=*,itemsep=1pt,topsep=2pt]
\item \emph{Lot sizing.} $P_{\{B\}}=P_{\{S\}}=P_{\{C\}}=P_{\{B,C\}}=\LP$. For every item $i$, period $l$, and set $\Pi\subseteq\{1,\dots,l\}$, the $(l,S)$ inequality $\sum_{t\in\Pi}x_{it}\le\sum_{t\in\Pi}d_{itl}y_{it}+s_{il}$, where $d_{itl}$ is the demand of item $i$ in periods $t$ to $l$, is valid for $\conv(X_{\{B,S\}})$. For every period $t$ and set $J$ of items with $\sum_{i\in J}M_{it}=C_t+\mu$ and $\mu>0$, the flow cover inequality $\sum_{i\in J}x_{it}+\sum_{i\in J}(M_{it}-\mu)^+(1-y_{it})\le C_t$ is valid for $\conv(X_{\{C,S\}})$.
\item \emph{Network design.} $P_{\{N\}}=P_{\{L\}}=P_{\{U\}}=P_{\{L,U\}}=\LP$. For a set $V$ of nodes, let $\delta^+(V)$ be the set of arcs leaving $V$ and $D_V$ the total demand of the commodities with origin in $V$ and destination outside $V$. The row $\sum_{a\in\delta^+(V)}u_ay_a\ge D_V$ is valid for $X_{\{N,U\}}$, so every inequality valid for its binary solutions, such as the cover inequality $\sum_{a\in\delta^+(V)\setminus R}y_a\ge1$ for $R\subseteq\delta^+(V)$ with $\sum_{a\in R}u_a<D_V$, is valid for $\conv(X_{\{N,U\}})$. With one commodity, $P_{\{N,L\}}=\LP$.
\item \emph{Facility location.} $P_{\{A\}}=P_{\{C\}}=P_{\{L\}}=P_{\{C,L\}}=\LP$. The row $\sum_iC_iy_i\ge\sum_jd_j$ is valid for $X_{\{A,C\}}$, so every inequality valid for its binary solutions, such as the cover inequality $\sum_{i\notin R}y_i\ge1$ for a set $R$ of facilities with $\sum_{i\in R}C_i<\sum_jd_j$, is valid for $\conv(X_{\{A,C\}})$. For an odd number $n\ge3$ of distinct facilities $i_1,\dots,i_n$, distinct customers $j_1,\dots,j_n$, and $i_{n+1}=i_1$, the odd-cycle inequality
\[
\sum_{\ell=1}^{n}y_{i_\ell}+\sum_{\ell=1}^{n}\ \sum_{i\notin\{i_\ell,i_{\ell+1}\}}x_{ij_\ell}\ge\frac{n+1}{2}
\]
is valid for $\conv(X_{\{A,L\}})$.
\end{enumerate}
In each formulation, every pair of families with positive gain is therefore a polyhedral zero-marginal group at $\emptyset$, and each of these inequalities that is violated by a point of $\LP$ has its pair of families as a joint support.
\end{proposition}

\begin{proof}
(a) The rows of $B$ and $C$ contain only the continuous $x$ and $s$, and each row $x_{it}\le M_{it}y_{it}$ is a perspective block of $\{S\}$ in which the bound of $x_{it}$ is $M_{it}$, so \cref{lem:nogain} gives the four equalities. The set $X_{\{B,S\}}$ is contained in the product over the items of the uncapacitated lot-sizing sets, on which the $(l,S)$ inequalities are valid \citep{barany1984uncapacitated}. The set $X_{\{C,S\}}$ is the product over the periods of the single-node fixed-charge sets $\{\sum_ix_{it}\le C_t,\ 0\le x_{it}\le M_{it}y_{it},\ y_{it}\in\{0,1\}\}$ and of a box, and the flow cover inequalities are valid on these sets \citep{padberg1985valid}.

(b) The rows of $N$ contain only flows. For each arc $a$, the rows of $L$ on $a$ form a perspective block of $\{L\}$, with the bounds $x_{qa}\le d_q$. The row of $U$ on $a$, alone or with the rows of $L$ on $a$, forms a perspective block of $\{U\}$ or of $\{L,U\}$. In all three cases, the inequalities $x_{qa}\le d_qy_a$ of \cref{lem:nogain} are rows of $L$. Summing the flow-balance rows of a commodity $q$ with origin in $V$ and destination outside $V$ over the nodes of $V$ gives $\sum_{a\in\delta^+(V)}x_{qa}-\sum_{a\in\delta^-(V)}x_{qa}=d_q$, where $\delta^-(V)$ is the set of arcs entering $V$, so $\sum_{a\in\delta^+(V)}x_{qa}\ge d_q$. The rows of $U$ then give $\sum_{a\in\delta^+(V)}u_ay_a\ge\sum_{a\in\delta^+(V)}\sum_qx_{qa}\ge D_V$ on $X_{\{N,U\}}$, and a binary $y$ that satisfies this row opens an arc outside $R$ whenever $\sum_{a\in R}u_a<D_V$. With one commodity of demand $d$, let $(\bar x,\bar y)$ be a vertex of the relaxation of the rows of $N$ and $L$ with the bounds, and let $A'$ be the set of arcs with $0<\bar y_a<1$. On $A'$, $\bar x_a=d\bar y_a$, since otherwise $\bar y_a$ could move in both directions. Let $Q$ be the polytope of the flows that satisfy the rows of $N$ and $0\le x\le d$. If $\bar x$ were not a vertex of $Q$, then $\bar x\pm\varepsilon v\in Q$ for some $v\ne0$ that vanishes where $\bar x_a\in\{0,d\}$. Setting $y_a=x_a/d$ on $A'$ and $y_a=\bar y_a$ elsewhere would then write $(\bar x,\bar y)$ as the midpoint of two points of the relaxation. So $\bar x$ is a vertex of $Q$. The rows of $N$ form a node-arc incidence matrix with right-hand sides $0$ and $\pm d$, so every vertex of $Q$ has entries in $\{0,d\}$, and $A'=\emptyset$. Every vertex of the relaxation, a polytope, thus lies in $X_{\{N,L\}}$, and $P_{\{N,L\}}=\LP$.

(c) The rows of $A$ contain only the continuous $x$. For each facility $i$, the rows of $L$ on $i$ form a perspective block of $\{L\}$, with the bounds $x_{ij}\le1$. Because $d_j>0$, the row of $C$ on $i$, alone or with the rows of $L$ on $i$, forms a perspective block of $\{C\}$ or of $\{C,L\}$. In all three cases, the inequalities $x_{ij}\le y_i$ of \cref{lem:nogain} are rows of $L$. Summing the capacity rows and the assignment rows weighted by $d_j$ gives $\sum_iC_iy_i\ge\sum_i\sum_jd_jx_{ij}=\sum_jd_j$ on $X_{\{A,C\}}$, and a binary $y$ that satisfies this row opens a facility outside $R$ whenever $\sum_{i\in R}C_i<\sum_jd_j$ \citep{aardal1995capacitated}. For the odd-cycle inequality \citep{cornuejols1982some}, let $O$ be the set of open facilities among $i_1,\dots,i_n$ at a point of $X_{\{A,L\}}$. If $|O|\ge(n+1)/2$, the first sum suffices. Otherwise, since each facility lies in the pairs $\{i_\ell,i_{\ell+1}\}$ of two customers, at least $n-2|O|$ customers $j_\ell$ have both facilities of their pair closed. By the linking and assignment rows, each of these customers contributes $1$ to the second sum, so the left side is at least $n-|O|\ge(n+1)/2$.

Every family of the three formulations has $P_{\{f\}}=\LP$, so \cref{lem:pairs} applies to every pair, and each displayed inequality is valid for $P_W$ of its pair $W$.
\end{proof}

With several commodities, flow balance and arc linking can have positive gain together. Let a network have the arcs $(r,v_i)$ for $i=1,2,3$, with fixed cost $1$, and the arcs $(v_i,t_j)$ for $i\ne j$, with fixed cost $0$, all with capacity $3$. Three commodities of demand $1$ go from $r$ to $t_1$, $t_2$, and $t_3$, and flows have no cost. The node $t_j$ is reached only through the nodes $v_i$ with $i\ne j$, so every integer solution opens at least two arcs $(r,v_i)$. The LP relaxation instead opens each of them to $\tfrac12$ and sends half of commodity $j$ through each $v_i$ with $i\ne j$. Its value is $\tfrac32$, because commodity $j$ gives $\sum_{i\ne j}y_{rv_i}\ge1$. The rows of $U$ are implied by those of $L$, so $z(\{N,L\})=z_{\IP}=2$, and $\{N,L\}$ is a polyhedral zero-marginal group at $\emptyset$ and a joint support of $y_{rv_1}+y_{rv_2}+y_{rv_3}\ge2$.

In generalized assignment, the assignment rows on binary variables form one simplex per job, so $P_{\{A\}}=\LP$, while each capacity row is a binary knapsack whose convex hull can cut off points of the LP relaxation. In both templates, convexifying the capacity family alone closes the whole gap, so no zero-marginal group of order two occurs at $\emptyset$, as \cref{tab:exact} records.

\paragraph{Connection with the search theory.}
The families of each of these groups share variables, as \cref{prop:lookahead} requires of every polyhedral group at $\emptyset$, so on them the value of the one-pass lookahead can depend on the dual. The two groups of a lot-sizing template share the setup family, and the two groups of FL $3\times3$ share the assignment family, which is the pattern of shared families in the second case of \cref{thm:overlap}(d). In each of these templates, the gains of the two groups sum to more than the gap (\cref{tab:exact}), so the gains overlap as on the instance of \cref{fig:lattice}. The templates therefore fall under \cref{thm:overlap} rather than under the additive gains of \cref{thm:priceSearch}.

\FloatBarrier
\section{Several Zero-Marginal Groups}\label[appendix]{app:multi}

A program can have several zero-marginal groups at $\emptyset$. We model them by a collection $\mathcal A=\{W_1,\dots,W_m\}$ of pairwise disjoint sets of $w$ families each, with $2\le w\le s<k$ and $mw\le k$, and the gain function
\begin{equation}\label{eq:gA}
g_{\mathcal A}(T)=\tfrac12\,\bigl|\{i : W_i\subseteq T\}\bigr| .
\end{equation}
Queries are the sets of $\cT_s$, a method chooses each query after the previous answers and observes $g(T)$, and its cost is the number of queries until it has queried a set of maximal gain over $\cT_s$, as in \cref{sec:cost}. Since $g_{\mathcal A}(T)>0$ exactly when $T$ contains some $W_i$, each $W_i$ is a zero-marginal group at $\emptyset$ of order $w$, and the $W_i$ are exactly the inclusion-minimal sets of positive gain. A set in $\cT_s$ contains at most $\lfloor s/w\rfloor$ of the groups, so with $r=\min\{m,\lfloor s/w\rfloor\}$ the maximal gain over $\cT_s$ is $r/2$, attained exactly by the sets that contain $r$ groups. For $0\le e\le w$, let $\kappa_e$ be the number of queries in stages $0$ to $e$ of the staged search of \cref{thm:pred}(a).

\begin{theorem}[several groups]\label{thm:multi}
Let $g=g_{\mathcal A}$ be as in \eqref{eq:gA}.
\begin{enumerate}[label=(\alph*),leftmargin=*,itemsep=2pt,topsep=2pt]
\item \emph{Realizability.} There is a binary program with $k$ constraint families whose gain function is $g_{\mathcal A}$. It is the edge formulation of stable set, with objective $\min-\sum_vx_v$, on $m$ odd cycles and $k-mw$ further paths, all on disjoint sets of vertices; the edges of the $i$-th cycle are split into $w$ paths, which form the families of $W_i$.
\item \emph{Identification.} Let $T$ be a queried set with $g(T)>0$, and let $\nu=2g(T)$. Removing one family at a time identifies the $\nu$ groups contained in $T$ with $\nu|T|-\tfrac12w\nu(\nu-1)\le\nu s$ further queries.
\item \emph{Without predictions.} For a method that knows $m$ and $w$, querying a minimum covering of the $w$-subsets of $F$ by $s$-sets, identifying the groups inside its queries of positive gain by (b), and querying the union of $r$ of them finds a set of maximal gain within $\cov(k,s,w)+ms+1$ queries. Every deterministic method needs at least $\bigl\lceil\binom{k-(m-1)w}{w}/\binom{s}{w}\bigr\rceil$ queries in the worst case to query a set of positive gain, so for fixed $m$, $w$, and $s$, the deterministic cost is $(1+o(1))\binom{k}{w}/\binom{s}{w}$, as for one group. Every randomized method needs more than $\binom{k}{w}/\bigl(2m\binom{s}{w}\bigr)$ queries in expectation for some instance.
\item \emph{Staged search with several predictions.} Let $P_1,\dots,P_m$ be sets of $w$ families, one per group, with unknown numbers of misses $e_i=|W_i\setminus P_i|$, and let $e_{(1)}\le\dots\le e_{(m)}$ be these numbers in increasing order. The stage-synchronized search runs stage $e'$ of \cref{thm:pred}(a) around every $P_i$ before stage $e'+1$ around any of them, identifies by (b) the groups not yet identified inside every stage query of positive gain, and queries the union of $r$ identified groups as soon as it has them. Without knowing any $e_i$, it queries a set of maximal gain within $m\kappa_{e_{(r)}}+ms+1$ queries, and within $m\kappa_{w-1}+\cov(k-w,s,w)+ms+1$ queries for any predictions. The first bound also holds with each $e_i$ replaced by $\min_j|W_i\setminus P_j|\le e_i$, so it does not depend on which prediction is meant for which group. For fixed $m$, $w$, $s$, and $e_{(r)}$, the first bound is $(1+o(1))\,m\,v_{e_{(r)}}\binom{k-w}{e_{(r)}}+O(1)=\Theta(k^{e_{(r)}})$, with $v_e$ as in \cref{thm:pred}, and the second is $\cov(k,s,w)+O(k^{w-1})$, a lower-order term more than the covering of (c).
\item \emph{Lower bound with several predictions.} Let $H=\{i : e_i\ge e_{(r)}\}$, let $V$ be the union of the groups $W_j$ with $j\notin H$, and for $i\in H$ let $p_i=|P_i\setminus V|$, $n_i=|F\setminus(P_i\cup V)|$,
\[
M_i=\binom{p_i}{w-e_i}\binom{n_i}{e_i},\qquad
D_i=\max_{w-e_i\le a\le\min(p_i,s-e_i)}\binom{a}{w-e_i}\binom{\min(n_i,s-a)}{e_i},
\]
and $\eta=\prod_{i\in H}\binom{n_i-(|H|-1)w}{e_i}\big/\binom{n_i}{e_i}$. Let $\theta$ be the share of the tuples $(X_i)_{i\in H}$ of sets $X_i\subseteq P_i\setminus V$ with $|X_i|=w-e_i$ whose sets are pairwise disjoint; $\theta\ge2^{-w|H|}$, and $\theta=1$ when the sets $P_i\setminus V$ are pairwise disjoint. Over the instances with the same predictions, the same numbers of misses, and the same groups $W_j$ with $j\notin H$, every deterministic method, even one that knows all of these, needs at least
\[
\frac{\theta\eta}{\sum_{i\in H}D_i/M_i}\;\ge\;\frac{\theta\eta}{|H|}\,\min_{i\in H}\frac{M_i}{D_i}
\]
queries in the worst case to query a set of maximal gain, and every randomized method needs more than half of this in expectation for some instance. For fixed $m$, $w$, $s$, and misses, $\eta=1-O(1/k)$ and $M_i/D_i=\Theta(k^{e_i})$, so both bounds are $\Omega(k^{e_{(r)}})$, and together with (d) the cost of finding a set of maximal gain is $\Theta(k^{e_{(r)}})$. When all groups fit into one query, $r=m$, and the cost is $\Theta(k^{\max_ie_i})$.
\end{enumerate}
\end{theorem}

For $m=1$, part (e) contains the randomized bound of \cref{thm:pred}(b), part (c) contains the randomized bound of \cref{thm:cover}(b), and the search of (d) queries in the order of the staged search of \cref{thm:pred}(a). The lower bounds in (c) and (e) hold on the programs of (a), and therefore on every class of programs that contains them. The construction in (a), with $k-|\bigcup_iW_i|$ further paths, gives the gain function \eqref{eq:gA} also for groups that share families, since a family can own one path in each of several cycles; parts (b) to (e) use that the groups are disjoint.

\begin{proof}[Proof of (a)]
For each $i$, take an odd cycle $C_i$ with at least $w$ vertices, split its edges into $w$ nonempty paths, and assign them to the families of $W_i$; each family outside $\bigcup_iW_i$ is a path on its own vertices. Every row lies in one component, so $\LP$ and $X_T$, and hence $\conv(X_T)$ and $P_T$, are products of their restrictions to the components, and $z(T)$ is the sum of the optimal values over the components. The paths are bipartite, so their rows give integral relaxations. If $T$ omits a family of $W_i$, the rows of $T$ on $C_i$ form a forest, the restriction of $\conv(X_T)$ to $C_i$ is the relaxation of these rows, and the restriction of $P_T$ to $C_i$ is the LP relaxation of $C_i$, with value $-|C_i|/2$. If $W_i\subseteq T$, the restriction of $P_T$ to $C_i$ is the stable-set polytope of $C_i$, with value $-(|C_i|-1)/2$. Hence $g(T)=\tfrac12|\{i : W_i\subseteq T\}|$. For groups that share a family, that family owns one path in each of the cycles of its groups, the components are still vertex-disjoint, and the same argument applies.
\end{proof}

\begin{proof}[Proof of (b)]
A set has positive gain exactly when it contains a group, so an inclusion-minimal set of positive gain is a group. A pass starts from a set $S$ of positive gain and, for each family $x$ of $S$ in a fixed order, queries the current set without $x$ and removes $x$ when the gain stays positive. The current set keeps a positive gain, and for each retained $x$ the gain without $x$ was zero when $x$ was tested and stays zero by monotonicity, since the set only shrinks afterward. The pass therefore ends at a group, after one query per family of $S$. The first pass starts from $T$ and each later pass from $T$ minus the groups found so far; the groups are disjoint, so the remaining groups in $T$ stay intact, and $\nu=2g(T)$ passes find all of them. Pass $j+1$ starts from $|T|-jw$ families, so the passes cost $\sum_{j=0}^{\nu-1}(|T|-jw)=\nu|T|-\tfrac12w\nu(\nu-1)$ queries. When some groups inside $T$ are already known, the passes start from $T$ minus these groups, and their number is $2g(T)$ minus the number of known groups inside $T$.
\end{proof}

\begin{proof}[Proof of (c)]
Every group lies in some query of the covering, and (b) identifies it after that query, at a cost of at most $s$ queries per group; the union of $r$ identified groups has at most $s$ families and gain $r/2$. This gives the upper bound.

Deterministic methods. Against the answer $0$ to every query, a deterministic method produces a fixed sequence $T_1,T_2,\dots$. After $q$ queries, at most $q\binom{s}{w}$ sets of $w$ families are contained in some query. Choose $W_1$ among the $w$-subsets of $F$ that no query contains, then $W_2$ among those of $F\setminus W_1$, and so on; the $i$-th choice exists when $\binom{k-(i-1)w}{w}>q\binom{s}{w}$, and all these binomial coefficients are at least $\binom{k-(m-1)w}{w}$. Hence, for $q\binom{s}{w}<\binom{k-(m-1)w}{w}$, some instance with pairwise disjoint groups is consistent with the $q$ answers and has no queried set of positive gain, which gives the bound. For fixed $m$, $w$, and $s$, $\binom{k-(m-1)w}{w}=(1-O(1/k))\binom{k}{w}$, and $\cov(k,s,w)=(1+o(1))\binom{k}{w}/\binom{s}{w}$ by \cref{thm:cover}(c), so the two bounds agree up to a factor $1+o(1)$.

Randomized methods. Draw the tuple $(W_1,\dots,W_m)$ uniformly among the tuples of pairwise disjoint sets of $w$ families. Every permutation of $F$ preserves this distribution, so each $W_i$ is uniform over the $w$-subsets of $F$. Fix the random bits of the method. Until the first answer of positive gain, all answers are $0$, so the method makes the queries $T_1,T_2,\dots$ of its sequence against the answer $0$, and a query $T$ contains $W_i$ with probability $\binom{|T|}{w}/\binom{k}{w}\le\binom{s}{w}/\binom{k}{w}$. With $J=\binom{k}{w}/\bigl(m\binom{s}{w}\bigr)$, the union bound over the $m$ groups and the first $q$ queries shows that the number $Q$ of queries until the first positive gain satisfies $\Pr[Q>q]\ge1-q/J$, and as in the proof of \cref{thm:cover}(b), $\mathbb{E}Q\ge\sum_{q=0}^{\lfloor J\rfloor}(1-q/J)>J/2$. Averaging over the random bits gives an instance with expected cost above $J/2$.
\end{proof}

\begin{proof}[Proof of (d)]
For each $i$, let $\hat e_i=\min_j|W_i\setminus P_j|$, so that $\hat e_i\le e_i$, and let $\hat e_{(r)}$ be the $r$-th smallest of these numbers, so that $\hat e_{(r)}\le e_{(r)}$. By \cref{thm:pred}(a), $W_i$ lies in some query of stage $\hat e_i$ around a prediction that attains $\hat e_i$. When stage $\hat e_{(r)}$ has been run around every prediction, every group with $\hat e_i\le\hat e_{(r)}$ has therefore been contained in a stage query of positive gain, and at least $r$ groups have $\hat e_i\le\hat e_{(r)}$. Each stage query of positive gain is followed by the identification of the groups not yet identified inside it, so at least $r$ distinct groups are known by the end of stage $\hat e_{(r)}$. Their union has $rw\le s$ families and gain $r/2$. The stages up to $\hat e_{(r)}$ cost at most $m\kappa_{\hat e_{(r)}}\le m\kappa_{e_{(r)}}$ queries, the identifications at most $s$ queries for each of at most $m$ groups, and the union one query. For the second bound, stages $0$ to $w$ around $P_1$ contain every set of $w$ families, so all groups are identified by the end of stage $w$ around $P_1$. Up to that point, the search makes at most $m\kappa_{w-1}$ queries in earlier stages and the queries of stage $w$ around $P_1$, which has at most the $\cov(k-w,s,w)\le\cov(k,s,w)$ queries of the $w$-cover $\{\emptyset\}$. By \cref{thm:pred}(c), $\kappa_e=(1+o(1))\,v_e\binom{k-w}{e}=\Theta(k^e)$ for fixed $s$, $w$, and $e$, and $ms+1$ is a constant.
\end{proof}

\begin{proof}[Proof of (e)]
At most $r-1$ groups have fewer than $e_{(r)}$ misses, so every set of gain $r/2$ contains a group of $H$. Fix the groups $W_j$ with $j\notin H$, and call the groups of $H$ hidden. A placement assigns to each $i\in H$ a set $X_i\cup Y_i$ with $X_i\subseteq P_i\setminus V$, $|X_i|=w-e_i$, $Y_i\subseteq F\setminus(P_i\cup V)$, and $|Y_i|=e_i$, such that the sets $X_i\cup Y_i$ are pairwise disjoint. Together with the fixed groups, every placement is an instance on which $P_i$ misses exactly $e_i$ members of the $i$-th group for every $i$. Let $\Lambda$ be the set of placements.

Counting placements. The true groups give a placement, with $X_i=W_i\cap P_i$, so at least one tuple $(X_i)_{i\in H}$ is pairwise disjoint, and the pairwise disjoint tuples are a share $\theta$ of all $\prod_{i\in H}\binom{p_i}{w-e_i}$ tuples. Since $p_i\le w$, each factor is at most $2^w$, so $\theta\ge2^{-w|H|}$. For a pairwise disjoint tuple, choose the sets $Y_i$ in a fixed order. The set $Y_i$ must avoid $X_j\cup Y_j$ for the earlier indices $j$ and $X_j$ for the later ones, at most $(|H|-1)w$ families in all, and it is disjoint from $X_i$ because $Y_i\cap P_i=\emptyset$; so it has at least $\binom{n_i-(|H|-1)w}{e_i}$ choices. Hence $|\Lambda|\ge\theta\eta\prod_{i\in H}M_i$.

Eliminating placements. A query $T$ is inconsistent with the placements in which some hidden group lies inside $T$. For $i\in H$, let $a=|T\cap(P_i\setminus V)|$; then $T$ contains at most $\binom{a}{w-e_i}$ choices of $X_i$ and at most $\binom{\min(n_i,s-a)}{e_i}$ choices of $Y_i$, and none unless $w-e_i\le a\le\min(p_i,s-e_i)$, so it contains at most $D_i$ candidates for the $i$-th group. The other coordinates of a placement have at most $M_j$ choices each, so one query eliminates at most $\mu=\sum_{i\in H}D_i\prod_{j\in H,\,j\ne i}M_j$ placements.

The adversary answers every query $T$ with half the number of fixed groups contained in $T$. While some placement has no hidden group inside any queried set, every answer is the gain of $T$ on that instance, and no queried set has gain $r/2$ there. After $q<|\Lambda|/\mu$ queries, such a placement remains, so the method needs at least $|\Lambda|/\mu\ge\theta\eta/\sum_{i\in H}(D_i/M_i)$ queries; the second bound follows from $\sum_{i\in H}D_i/M_i\le|H|\max_{i\in H}D_i/M_i$. For a randomized method, draw the placement uniformly from $\Lambda$ and fix the random bits. Until it queries a set that contains a hidden group, the method receives the answers of the adversary, so its queries form a fixed sequence, and the first $q$ of them contain a hidden group with probability at most $q\mu/|\Lambda|$. As in the proof of (c), the expected number of queries exceeds $|\Lambda|/(2\mu)$, and averaging over the random bits gives an instance with this expected cost.

Finally, the true placement gives $p_i\ge w-e_i$ and $n_i\ge e_i$. For fixed $m$, $w$, $s$, and misses, $n_i\ge k-mw$, so $\eta=1-O(1/k)$, $M_i=\Theta(k^{e_i})$, and $D_i=\Theta(1)$. Every $i\in H$ has $e_i\ge e_{(r)}$, with equality for some $i$, so $\min_{i\in H}M_i/D_i=\Theta(k^{e_{(r)}})$, and $\theta\ge2^{-w|H|}$ is a constant. When $r=m$, $e_{(r)}=\max_ie_i$.
\end{proof}

\begin{proof}[Proof of \cref{thm:several}]
Parts (a), (b), and (c) are the asymptotic forms of parts (c), (d), and (e) of \cref{thm:multi}.
\end{proof}

\paragraph{Groups that share families.}
The construction of \cref{thm:multi}(a) applies also when groups share families. For a graph $G=(F,E)$ on the families, let every edge $uv\in E$ be a group $\{u,v\}$. It is realized by a triangle of its own whose edges are split into a path of one edge, owned by $u$, and a path of two edges, owned by $v$. A family on no edge owns a path of one edge on two further vertices. For $T\subseteq F$, let $E(T)$ be the set of edges with both ends in $T$.

\begin{proposition}[combining groups]\label{prop:dks}
On this program, $g(T)=\tfrac12|E(T)|$ for every $T\subseteq F$, and every bound $z(T)$ is computable in time polynomial in the size of the program. Choosing a set of maximal gain over $\cT_s$ is the densest-subgraph problem with $s$ vertices on $G$. It is NP-hard when $s$ is part of the input. A polynomial-time method that attains a fraction $\alpha$ of the maximal gain on every such program yields a polynomial-time method that attains a fraction $\alpha$ of the maximal number of edges among $s$ vertices on every graph.
\end{proposition}

\begin{proof}
Each triangle is a component, and all its rows are convexified exactly when both of its owners are in $T$, so the argument of (a) gives $g(T)=\tfrac12|E(T)|$. The polytope $P_T$ is the product of the LP relaxations of the paths and of the triangles whose owners are not both in $T$, and of the stable-set polytopes of the other triangles. The stable-set polytope of a triangle on the vertices $a$, $b$, and $c$ is described by nonnegativity and $x_a+x_b+x_c\le1$. Hence $z(T)$ is the value of an LP of polynomial size, and $z(T)=z_{\LP}+\tfrac12|E(T)|$. A set $T$ with $|T|\le s$ has $|E(T)|\le\binom{s}{2}$, with equality exactly when $T$ is a clique of $s$ vertices, so deciding whether the maximal gain over $\cT_s$ reaches $\tfrac12\binom{s}{2}$ is the clique problem, which is NP-complete \citep{karp1972reducibility}. The program has at most $3|E|+2|F|$ variables and is built from $G$ in polynomial time, and a set of gain $\alpha$ times the maximal gain has $\alpha$ times the maximal number of edges.
\end{proof}

For a fixed $s$, enumerating the $O(k^s)$ sets of at most $s$ families solves the problem once the groups are known, so the hardness concerns budgets that grow with $k$. By \cref{prop:dks}, the inapproximability results for the densest-subgraph problem with a given number of vertices \citep{khot2006ruling,manurangsi2017almost} hold also for choosing the best set of families.

\FloatBarrier
\section{Evaluation Cost and Prices}\label[appendix]{app:cost}

Throughout this section, a collection $\mathcal{A}=\{W_1,\dots,W_m\}$ of groups, which may share families and differ in size, with weights $v_i>0$, defines the additive gain $g_{\mathcal{A}}(T)=\sum_{i:\,W_i\subseteq T}v_i$. With rational weights, \cref{lem:realize}, applied to one unit per group whose collection consists of that group alone and whose weight is $v_i$, realizes it as the gain function of a binary program.

\paragraph{Prices.}

\begin{proposition}[prices]\label{prop:price}
Let $g=g_{\mathcal{A}}$, let $c_f\ge0$ for $f\in F$, and let $\Phi(T)=g(T)-\sum_{f\in T}c_f$.
\begin{enumerate}[label=(\alph*),leftmargin=*,itemsep=1pt,topsep=2pt]
\item $g$ is supermodular: $g(A\cup B)+g(A\cap B)\ge g(A)+g(B)$ for all $A,B\subseteq F$.
\item Consider the network with a source, a sink, a node for each group, and a node for each family, and with arcs from the source to group $i$ with capacity $v_i$, from group $i$ to each family of $W_i$ with infinite capacity, and from each family $f$ to the sink with capacity $c_f$. Its minimum cut capacity equals $\sum_iv_i-\max_{T\subseteq F}\Phi(T)$, and the families on the source side of a minimum cut maximize $\Phi$ \citep{rhys1970selection,balinski1970selection,picard1976maximal}.
\item Let $c_f=\lambda$ for every $f$ and write $\Phi_\lambda(T)=g(T)-\lambda|T|$. The maximizers of $\Phi_\lambda$ are closed under union, so a largest maximizer $T^*(\lambda)$ exists; $T^*(\lambda)\subseteq T^*(\lambda')$ whenever $\lambda'\le\lambda$; at most $k+1$ distinct sets $T^*(\lambda)$ occur; and one parametric minimum cut computes all of them \citep{gallo1989fast}.
\item Every maximizer $T$ of $\Phi_\lambda$ maximizes $g$ over the sets of at most $|T|$ families.
\end{enumerate}
\end{proposition}

\begin{proof}
(a) It suffices to prove the inequality for one indicator $[W\subseteq T]$, since nonnegative combinations of supermodular functions are supermodular. If $W\subseteq A\cap B$, both sides equal $2$. If $W\subseteq A$ and $W\not\subseteq B$, the left side is at least $1$ because $W\subseteq A\cup B$, and the right side is $1$; the case $W\subseteq B$ is symmetric. If $W$ lies in neither set, the right side is $0$.

(b) Let $S$ be the source side of a cut of finite capacity. No arc of infinite capacity leaves $S$, so if group $i$ is in $S$, every family of $W_i$ is in $S$. The capacity of the cut is $\sum_{i\notin S}v_i+\sum_{f\in S}c_f=\sum_iv_i-\bigl(\sum_{i\in S}v_i-\sum_{f\in S}c_f\bigr)$. For the set $T$ of families in $S$, the groups in $S$ are contained in $T$, so the bracket is at most $\Phi(T)$, with equality when $S$ contains exactly the groups inside $T$. Hence the minimum capacity is $\sum_iv_i-\max_T\Phi(T)$, and the families of a minimum cut attain the maximum.

(c) Let $\lambda'\le\lambda$, let $T$ maximize $\Phi_\lambda$, and let $T'$ maximize $\Phi_{\lambda'}$. Optimality of $T$ against $T\cap T'$ gives $g(T)-g(T\cap T')\ge\lambda(|T|-|T\cap T'|)$. By (a) and $|T\cup T'|-|T'|=|T|-|T\cap T'|$,
\[
\Phi_{\lambda'}(T\cup T')-\Phi_{\lambda'}(T')\ge g(T)-g(T\cap T')-\lambda'(|T|-|T\cap T'|)\ge(\lambda-\lambda')(|T|-|T\cap T'|)\ge0,
\]
so $T\cup T'$ maximizes $\Phi_{\lambda'}$. With $\lambda'=\lambda$, this shows that the maximizers are closed under union, and with $\lambda'<\lambda$, applied to $T=T^*(\lambda)$ and $T'=T^*(\lambda')$, it gives $T^*(\lambda)\cup T^*(\lambda')=T^*(\lambda')$. The sets $T^*(\lambda)$ form a chain of subsets of $F$, so at most $k+1$ of them are distinct. In the network of (b) with the parameter $\mu=-\lambda$, the source capacities do not depend on $\mu$ and the sink capacities $\lambda=-\mu$ decrease in $\mu$, which is the monotone parametric setting of \citet{gallo1989fast}; run so that it returns the largest source side of a minimum cut, their algorithm gives $T^*(\lambda)$ for all $\lambda$ in the time of one maximum flow.

(d) If $|T'|\le|T|$, then $g(T')\le g(T)+\lambda(|T'|-|T|)\le g(T)$.
\end{proof}

The price is the Lagrangian relaxation of the budget. Let $G(s)=\max\{g(T):|T|\le s\}$. By (d), a maximizer of $\Phi_\lambda$ of size $s$ solves the budget problem with budget $s$, and such a maximizer exists for some $\lambda$ exactly when $(s,G(s))$ lies on the upper concave envelope of $G$, where a line of slope $\lambda$ supports the envelope. The hardness of \cref{prop:dks} concerns the other budgets.

\paragraph{Evaluation cost.} One group $W\subseteq F$ of $w\ge2$ families is hidden, and $w$ is known. A query of a nonempty set $T$ reveals $[W\subseteq T]=2g_W(T)$, for the gain function $g_W$ of \cref{thm:cover}, and costs $\varphi(|T|)$ for a nondecreasing function $\varphi$; the main case is $\varphi(s)=s^d$ with $d>0$. For a price $0<\lambda<1/(2w)$, $W$ is the unique maximizer of $g_W(T)-\lambda|T|$, and a query of a set of $w$ families is answered $1$ exactly when the set is $W$. A method therefore stops once it has queried $W$, and its cost is the total cost of its queries; without this requirement, one query of $F$ would end every search. A prediction is a set $P$ of $w$ families with $e=|W\setminus P|$ misses. For a set $\mathcal{C}$ of candidate groups and a query $T$, let $h_{\mathcal{C}}(T)$ be the number of candidates contained in $T$.

The \emph{halving search} keeps a set $Z\supseteq W$, starting from $Z=F$. While $|Z|>4w$, it splits $Z$ into $2w$ blocks whose sizes differ by at most one, queries $Z\setminus B$ for each block $B$, and replaces $Z$ by the union of the blocks whose query is answered $0$. When $|Z|\le4w$, it queries $Z$ if $|Z|=w$, since then $Z=W$; otherwise, it queries $Z\setminus\{x\}$ for each $x\in Z$, which identifies $W$, and then queries $W$. The \emph{small staged search} runs the staged search of \cref{thm:pred} with queries of at most $2w$ families. Its first query $T$ answered $1$ contains $W$; it then runs a removal pass on $T$, which queries the current set without one family at a time and keeps each removal that is answered $1$, and stops at $W$. The \emph{cost-balanced interleaving} of two searches lets, at each step, the search whose spent cost plus the cost of its next query is smaller make its next query, with ties to the first search, and stops when either search queries $W$.

\begin{lemma}[interleaving]\label{lem:interleave}
If the two searches alone cost $K_1$ and $K_2$ on an instance, their cost-balanced interleaving costs at most $2\min(K_1,K_2)$ on it.
\end{lemma}
\begin{proof}
Each search sees only the answers to its own queries, so it makes the same queries as when it runs alone. While the second search has not stopped, every query of the first search is made when its spent cost plus the cost of that query is at most the spent cost of the second search plus the cost of the next query of the second search, which is at most $K_2$, because the second search makes that query before it stops. So the first search has spent at most $K_2$ as long as the second has not stopped, and symmetrically. If the first search stops first, its spent cost is $K_1$, so $K_1\le K_2$, the second search has spent at most $K_1$, and the total is at most $2K_1=2\min(K_1,K_2)$. The other case is symmetric.
\end{proof}

\begin{lemma}[halving search]\label{lem:halving}
The halving search queries $W$ and costs at most $2w\sum_{t}\varphi\bigl(\lfloor(3/4)^tk\rfloor\bigr)+4w\,\varphi(4w)+\varphi(w)$, where the sum runs over the integers $t\ge0$ with $(3/4)^tk>4w$. For $\varphi(s)=s^d$, this is at most $2wk^d/(1-(3/4)^d)+(4w)^{d+1}+w^d$.
\end{lemma}
\begin{proof}
If $W\subseteq Z$, the query $Z\setminus B$ is answered $0$ exactly when $B$ meets $W$, so the kept blocks contain $W$, and at most $w$ blocks are kept. Each block has at most $\lceil|Z|/(2w)\rceil\le|Z|/(2w)+1$ families, so the new set has at most $|Z|/2+w<3|Z|/4$ families when $|Z|>4w$. By induction, round $t$ starts from at most $(3/4)^tk$ families and makes $2w$ queries of at most that size. In the last phase, $Z\setminus\{x\}$ is answered $0$ exactly when $x\in W$, which identifies $W$ with at most $4w$ queries of fewer than $4w$ families.
\end{proof}

\begin{lemma}[deterministic covering]\label{lem:dcover}
Let the instances be the groups of a set $\mathcal{C}$ of candidates, each a set of $w$ families. Every deterministic method has worst-case cost at least
\[
\bigl(|\mathcal{C}|-1\bigr)\min\Bigl\{\frac{\varphi(|T|)}{h_{\mathcal{C}}(T)} : h_{\mathcal{C}}(T)>0\Bigr\}+\varphi(w).
\]
\end{lemma}
\begin{proof}
The adversary keeps the set $\mathcal{C}_t$ of candidates consistent with the answers so far, starting from $\mathcal{C}$. It answers a query $T$ with $0$ if some candidate of $\mathcal{C}_t$ is not contained in $T$, and removes the candidates contained in $T$; otherwise, it answers $1$ and removes none. Every candidate left in $\mathcal{C}_t$ is consistent with all answers: it is not contained in any query answered $0$, and every query answered $1$ contained all candidates left at that time. If the method queries a candidate $C$ while another candidate $C'$ is left, then $C'\not\subseteq C$ because both have $w$ families, so the answer is $0$, and $C$ is removed. The method therefore queries the true group only after the queries answered $0$ have removed $|\mathcal{C}|-1$ candidates, at a cost of at least $\varphi(|T|)$ for a query that removes at most $h_{\mathcal{C}}(T)$ of them, and then pays $\varphi(w)$ for the last query. The adversary fixes the true group as the last candidate, so this transcript is that of an instance.
\end{proof}

\begin{lemma}[randomized covering]\label{lem:rcover}
Let the instances be the groups of a set $\mathcal{C}$ of candidates, each a set of $w$ families, and suppose every query $T$ either has $\varphi(|T|)\ge\Delta$ or has $\varphi(|T|)\ge\delta\,h_{\mathcal{C}}(T)$. For the uniform distribution on $\mathcal{C}$, every deterministic method has expected cost at least $\min\{\delta|\mathcal{C}|/2,\,\Delta\}$, and every randomized method has expected cost at least this much on some instance.
\end{lemma}
\begin{proof}
Fix a deterministic method, let $T_1,T_2,\dots$ be its queries when every answer is $0$, and call a query large if $\varphi(|T|)\ge\Delta$. Let $Q$ be the first index with $W\subseteq T_Q$; the method makes the queries $T_1,\dots,T_Q$, because it stops only after querying $W$ and its answers before $T_Q$ are $0$. Let $I$ be the index of the first large query in the sequence ($I=\infty$ if there is none), and let $x_i=h_{\mathcal{C}}(T_i)/|\mathcal{C}|$ and $X_i=\sum_{l<i}x_l$. If some candidate lies in no $T_i$, the method never stops on it, and the bound holds; otherwise, $X_I\ge1$ when $I=\infty$. The union bound gives $\Pr[Q\ge i]\ge1-X_i$. Hence the expected cost is at least
\[
\sum_{i<I}\delta\,h_{\mathcal{C}}(T_i)\Pr[Q\ge i]+\Delta\Pr[Q\ge I]\;\ge\;\delta|\mathcal{C}|\sum_{i<I}x_i(1-X_i)^++\Delta(1-Y),
\]
with $Y=\min(X_I,1)$ and the last term absent when $I=\infty$. Since $(1-X_i)^+\ge(1-u)^+$ for $u\in[X_i,X_i+x_i]$, the sum is at least $\int_0^Y(1-u)\,du=Y-Y^2/2\ge Y/2$, and the expected cost is at least $\delta|\mathcal{C}|Y/2+\Delta(1-Y)\ge\min\{\delta|\mathcal{C}|/2,\Delta\}$. A randomized method is a distribution over deterministic methods, so its expected cost under the uniform distribution is at least this bound, and some instance attains at least the average.
\end{proof}

\begin{theorem}[evaluation cost]\label{thm:power}
Let $\varphi(s)=s^d$ with $d>0$ and $w\ge2$. The explicit bounds below hold for every $k>w$ with $k-w\ge e$; the $O$, $\Omega$, and $\Theta$ statements are for fixed $w$ and $d$ as $k\to\infty$.
\begin{enumerate}[label=(\alph*),leftmargin=*,itemsep=1pt,topsep=2pt]
\item \emph{Without a prediction.} Every deterministic method has worst-case cost at least the first of the following bounds, and every randomized method has expected cost at least the second on some instance:
\begin{align*}
&\Bigl(\binom{k}{w}-1\Bigr)w!\min\bigl(w^{d-w},k^{d-w}\bigr)+w^d,\\
&\min\Bigl\{\tfrac12\,w!\min\bigl(w^{d-w},(k/2)^{d-w}\bigr)\binom{k}{w},\,(k/2)^d\Bigr\}.
\end{align*}
Querying the sets of $w$ families one at a time costs at most $\binom{k}{w}w^d$, and the halving search costs $O(k^d)$. Hence the optimal cost is $\Theta(k^{\min(w,d)})$, for deterministic and for randomized methods.
\item \emph{With a prediction.} Let $1\le e\le w$ and $N_e=\binom{w}{e}\binom{k-w}{e}$. On the groups with $e$ misses, every deterministic method that knows $P$ and $e$ has worst-case cost at least the first of the following bounds, and every randomized method has expected cost at least the second on one of them:
\begin{align*}
&(N_e-1)\frac{e!}{\binom{w}{e}}\min\bigl(e^{d-e},(k-w)^{d-e}\bigr)+w^d,\\
&\min\Bigl\{\frac{e!\,N_e}{2\binom{w}{e}}\min\bigl(e^{d-e},((k-w)/2)^{d-e}\bigr),\,\bigl((k-w)/2\bigr)^d\Bigr\}.
\end{align*}
The cost-balanced interleaving of the small staged search and the halving search costs $O(k^{\min(e,d)})$ without knowing $e$. Hence the optimal cost is $\Theta(k^{\min(e,d)})$. For $e=0$, the first query of the staged search is $P=W$, at cost $w^d$.
\item The prediction divides the optimal cost by a factor of order $k^{\min(w,d)-\min(e,d)}$: by $k^{w-e}$ when $d\ge w$, as under the budget of \cref{thm:pred}, by $k^{d-e}$ when $e<d<w$, and by a constant when $d\le e$.
\end{enumerate}
\end{theorem}

\begin{proof}
(a) The candidates are all sets of $w$ families, and $h(T)=\binom{|T|}{w}\le|T|^w/w!$. A query with $h(T)>0$ has $w\le|T|\le k$, so $\varphi(|T|)/h(T)\ge w!\,|T|^{d-w}\ge w!\min(w^{d-w},k^{d-w})$, and \cref{lem:dcover} gives the deterministic bound. For \cref{lem:rcover}, take $\Delta=(k/2)^d$, which every query of at least $k/2$ families costs; a query with $w\le|T|<k/2$ has $\varphi(|T|)\ge w!\min(w^{d-w},(k/2)^{d-w})h(T)$, and a query with $|T|<w$ has $h(T)=0$. With $\binom{k}{w}\ge(k-w+1)^w/w!$, both lower bounds are $\Omega(k^w)$ when $d\ge w$ and $\Omega(k^d)$ when $d<w$. The first upper bound is immediate, and the second is \cref{lem:halving}.

(b) The candidates are the sets $W'$ of $w$ families with $|W'\setminus P|=e$, and there are $N_e$ of them. A query $T$ with $b=|T\setminus P|$ contains at most $\binom{w}{e}\binom{b}{e}\le\binom{w}{e}b^e/e!$ of them, none if $b<e$, and costs at least $b^d$. For $e\le b\le k-w$, this gives $\varphi(|T|)/h(T)\ge\frac{e!}{\binom{w}{e}}b^{d-e}\ge\frac{e!}{\binom{w}{e}}\min(e^{d-e},(k-w)^{d-e})$, and \cref{lem:dcover} gives the deterministic bound. For \cref{lem:rcover}, take $\Delta=((k-w)/2)^d$, which every query with $b\ge(k-w)/2$ costs, and use $b^{d-e}\ge\min(e^{d-e},((k-w)/2)^{d-e})$ for $e\le b<(k-w)/2$. Both bounds are $\Omega(k^e)$ when $d\ge e$ and $\Omega(k^d)$ when $d<e$. For the upper bound, \cref{thm:pred}(a) and (c) with $s=2w$ show that the small staged search reaches a query containing $W$ within $O(k^e)$ queries of cost at most $(2w)^d$; the removal pass then costs at most $2w(2w)^d$, and it queries $W$, because the last kept removal leaves $W$, or $T=W$. The halving search costs $O(k^d)$, and \cref{lem:interleave} gives $O(k^{\min(e,d)})$.

(c) follows from (a) and (b).
\end{proof}

For every nondecreasing $\varphi$, \cref{lem:dcover} gives the lower bound $(\binom{k}{w}-1)\min_{w\le s'\le k}\varphi(s')/\binom{s'}{w}+\varphi(w)$, and the searches give the upper bounds $\binom{k}{w}\varphi(w)$, the bound of \cref{lem:halving}, and $\cov(k,s',w)\,\varphi(s')+s'\varphi(s')$ for every $w\le s'\le k$: query a minimum covering by sets of $s'$ families and run a removal pass on the first query answered $1$. Under the budget of \cref{thm:cover}, $\varphi(s')=1$ for $s'\le s$ and $\varphi(s')=\infty$ otherwise, and the lower bound is $(\binom{k}{w}-1)/\binom{s}{w}+1$. It exceeds the counting bound $\binom{k}{w}/\binom{s}{w}$ of \cref{thm:cover}(c) by $1-1/\binom{s}{w}$, because this model also charges the final query of $W$, and the two agree to first order as $k$ grows.

\begin{proof}[Proof of \cref{thm:costMain}]
Parts (a) and (b) of \cref{thm:power} give the two orders, for deterministic and for randomized methods.
\end{proof}

\paragraph{Searching under a price.}
The groups of $\mathcal{A}$ are now unknown, each has at most $w$ families, and the decomposition is chosen with a price $\lambda>0$ per convexified family. A method knows $w$ and $\lambda$, and a query of $T$ returns $g_{\mathcal{A}}(T)$. Under the budget, every query has $|T|\le s$ with $w\le s<k$, and the cost is the number of queries. Under evaluation costs, a query of $T$ costs $\varphi(|T|)=|T|^d$ with $d>0$. For $\mathcal{D}\subseteq\mathcal{A}$, let $T^*(\mathcal{D})$ be the largest maximizer of $g_{\mathcal{D}}(T)-\lambda|T|$, which exists by \cref{prop:price}(c). A group is \emph{identified} when it belongs, with its weight, to every instance of the class under consideration that is consistent with the answers so far. Unless stated otherwise, this class consists of all collections of groups of at most $w$ families with positive weights. The current decomposition of a method is $T^*(\mathcal{D})$ for the collection $\mathcal{D}$ of identified groups, and the cost of the method is the cost of its queries until its current decomposition is $T^*(\mathcal{A})$. A \emph{certificate} is a collection $\mathcal{B}\subseteq\mathcal{A}$ with $T^*(\mathcal{B})=T^*(\mathcal{A})$. The predictions are sets of families whose union $\bar P$ has a fixed number of families, with $\bar P=\emptyset$ without predictions, and a group $W$ has $\tilde e(W)=|W\setminus\bar P|$ families that no prediction contains. Let $e^*$ be the minimum, over the certificates $\mathcal{B}$, of $\max_{W\in\mathcal{B}}\tilde e(W)$, where the maximum over the empty collection is $0$.

\begin{lemma}[monotone decomposition]\label{lem:monodec}
If $\mathcal{D}\subseteq\mathcal{E}\subseteq\mathcal{A}$, then $T^*(\mathcal{D})\subseteq T^*(\mathcal{E})$. Hence every collection that contains a certificate is a certificate, and the current decomposition of a method never loses a family and is always contained in $T^*(\mathcal{A})$.
\end{lemma}
\begin{proof}
Let $\Phi_{\mathcal{D}}(T)=g_{\mathcal{D}}(T)-\lambda|T|$, $T_1=T^*(\mathcal{D})$, and $T_2=T^*(\mathcal{E})$. Optimality of $T_1$ gives $\Phi_{\mathcal{D}}(T_1)\ge\Phi_{\mathcal{D}}(T_1\cap T_2)$, and \cref{prop:price}(a) for $g_{\mathcal{D}}$, together with $|T_1\cup T_2|+|T_1\cap T_2|=|T_1|+|T_2|$, gives $\Phi_{\mathcal{D}}(T_1\cup T_2)-\Phi_{\mathcal{D}}(T_2)\ge\Phi_{\mathcal{D}}(T_1)-\Phi_{\mathcal{D}}(T_1\cap T_2)\ge0$. The difference $g_{\mathcal{E}}-g_{\mathcal{D}}$ is nondecreasing, so $\Phi_{\mathcal{E}}(T_1\cup T_2)\ge\Phi_{\mathcal{E}}(T_2)$, and $T_1\cup T_2$ maximizes $\Phi_{\mathcal{E}}$ and lies in its largest maximizer $T_2$. This is the monotonicity of maximizers of \citet{topkis1978minimizing}, with $\mathcal{D}$ as the parameter. If a certificate $\mathcal{B}$ lies in $\mathcal{D}$, then $T^*(\mathcal{A})=T^*(\mathcal{B})\subseteq T^*(\mathcal{D})\subseteq T^*(\mathcal{A})$. Every identified group belongs to $\mathcal{A}$, which is consistent with all answers, and stays identified when answers are added, so the identified collection only grows, and the last claim follows.
\end{proof}

\begin{theorem}[searching under a price]\label{thm:priceSearch}
Let $m$, $w$, $s$, $d$, and $|\bar P|$ be fixed, and let $k\to\infty$.
\begin{enumerate}[label=(\alph*),leftmargin=*,itemsep=1pt,topsep=2pt]
\item \emph{Staged search under the budget.} For $q=0,1,\dots,w$ in turn, stage $q$ queries $X\cup R$ for every $X\subseteq\bar P$ with $1\le|X|+q\le w$ and every $R$ in a minimum covering of the $q$-subsets of $F\setminus\bar P$ by sets of at most $s-|X|$ families, with $R=\emptyset$ for $q=0$. After every stage query $T$ with $g(T)>g_{\mathcal{D}}(T)$, where $\mathcal{D}$ is the identified collection, the search queries the nonempty subsets of $T$ of at most $w$ families before its next stage query. Without knowing $e^*$, it reaches $T^*(\mathcal{A})$ within $O(k^{e^*})$ queries, and without predictions within $(1+o(1))\cov(k,s,e^*)$ queries when $e^*\ge1$.
\item \emph{Evaluation cost.} Without the budget, the cost-balanced interleaving of the search of (a) with $s=w$ and the halving search of the proof reaches $T^*(\mathcal{A})$ at cost $O(k^{\min(e^*,d)})$.
\item \emph{Lower bound.} Let $W\in\mathcal{A}$ belong to every certificate, let no family of $W\setminus\bar P$ belong to another group, and let $\tilde e(W)\ge1$. Consider the class of instances obtained from $\mathcal{A}$ by replacing $W\setminus\bar P$ with a set of $\tilde e(W)$ families that lie neither in $\bar P$ nor in another group. On this class, every deterministic method needs $\Omega(k^{\tilde e(W)})$ queries under the budget and cost $\Omega(k^{\min(\tilde e(W),d)})$ under evaluation costs in the worst case, and every randomized method needs as much in expectation on some instance.
\item \emph{Tight cases.} Suppose that the groups are pairwise disjoint, with $a_i=v_i-\lambda|W_i|$, or that they share one family $\hat f\in\bar P$, each has a family besides $\hat f$, and they are otherwise pairwise disjoint, with $a_i=v_i-\lambda(|W_i|-1)$ and $\sum_{i:a_i>0}a_i\ge\lambda$. Then the certificates are the collections that contain $\mathcal{B}^*=\{W_i:a_i\ge0\}$, $e^*=\max_{W\in\mathcal{B}^*}\tilde e(W)$, and for $e^*\ge1$ the orders in (a) and (b) are tight by (c). In the second case with $\sum_{i:a_i>0}a_i<\lambda$, $T^*(\mathcal{A})=\emptyset$, and the cost is zero.
\end{enumerate}
\end{theorem}

Under a price, every group of $\mathcal{B}^*$ must be identified. For disjoint groups with $a_i\ge0$ for every $i$, the largest of the numbers $\tilde e(W_i)$ therefore sets the order of the cost, whereas under the budget of \cref{thm:multi} the $r$-th smallest number of misses does.

\begin{proof}[Proof of \cref{thm:priceSearch}]
(a) For $S\subseteq T$ with $|S|\le w$, every group inside $S$ has at most $w$ families, so $g(S)=\sum_{S'\subseteq S}f(S')$, where $f(S')$ is the weight of $S'$ if $S'\in\mathcal{A}$ and $0$ otherwise, and $f(\emptyset)=g(\emptyset)=0$. M\"obius inversion gives $f(S)=\sum_{S'\subseteq S}(-1)^{|S\setminus S'|}g(S')$, so the answers for the subsets of $T$ with at most $w$ families determine every group inside $T$ and its weight. This holds on every collection of groups of at most $w$ families that is consistent with these answers, so these groups are identified. A stage query $T$ with $g(T)>g_{\mathcal{D}}(T)$ contains a group outside $\mathcal{D}$, so at most $m$ identifications occur, each with at most $\sum_{j=1}^{w}\binom{s}{j}$ queries. Let $\mathcal{B}$ be a certificate with $\max_{W\in\mathcal{B}}\tilde e(W)=e^*$, and let $W\in\mathcal{B}$. In stage $\tilde e(W)$, the set $X=W\cap\bar P$ has $|X|+\tilde e(W)=|W|$, and $\tilde e(W)\le w-|X|\le s-|X|$, so some $R$ of the covering contains $W\setminus\bar P$, and the query $X\cup R$ contains $W$. At that query, $W$ is identified already, or $g(X\cup R)-g_{\mathcal{D}}(X\cup R)\ge v_W>0$, and $W$ is identified. At the end of stage $e^*$, the identified collection therefore contains $\mathcal{B}$, and by \cref{lem:monodec} the current decomposition is $T^*(\mathcal{A})$. The $q$-subsets themselves form a covering, so stage $q$ makes at most $2^{|\bar P|}\binom{k}{q}$ queries, and the stages up to $e^*$ make $O(k^{e^*})$ queries. Without predictions, $X=\emptyset$, stage $q$ is a minimum covering of the $q$-subsets of $F$ by $s$-sets, and $\sum_{q=1}^{e^*}\cov(k,s,q)=(1+o(1))\cov(k,s,e^*)$ by \cref{thm:cover}(c), while the identifications add $O(1)$ queries.

(b) The search of (a) with $s=w$ makes $O(k^{e^*})$ queries of at most $w$ families, at cost $O(k^{e^*})$. The halving search runs passes and uses only the collection $\mathcal{D}_h$ identified from its own answers. A pass queries $F$ and ends the search if $g(F)=g_{\mathcal{D}_h}(F)$, which holds exactly when $\mathcal{D}_h=\mathcal{A}$. Otherwise, it sets $Z=F$ and, while $|Z|>4w$, splits $Z$ into $2w$ blocks whose sizes differ by at most one, and for each block $B$ in turn queries $Z\setminus B$ and replaces $Z$ by $Z\setminus B$ if $g(Z\setminus B)>g_{\mathcal{D}_h}(Z\setminus B)$. The set $Z$ always contains a group outside $\mathcal{D}_h$. Fix such a group $W\subseteq Z$ at the end of a round. Every block kept in the round meets $W$, because, when it was tested, the current set contained $W$ and the current set without the block contained no group outside $\mathcal{D}_h$. Hence at most $w$ blocks are kept, and the round ends with at most $w\lceil|Z|/(2w)\rceil\le|Z|/2+w<3|Z|/4$ families. When $|Z|\le4w$, the pass queries the nonempty subsets of $Z$ of at most $w$ families and identifies the groups inside $Z$ as in (a), at least one of them new. Round $t$ of a pass makes $2w$ queries of at most $(3/4)^tk$ families, so a pass costs at most $k^d+2wk^d/(1-(3/4)^d)+2^{4w}(4w)^d$. At most $m+1$ passes occur, so the halving search reaches $T^*(\mathcal{A})$ at cost $O(k^d)$. In the interleaving, each search computes its identified collection from its own answers and therefore makes the same queries as when it runs alone, while the current decomposition uses the answers of both. When one search has made the queries with which it reaches $T^*(\mathcal{A})$ alone, the groups identified from its answers form a certificate, and by \cref{lem:monodec} so do the groups identified from the answers of both searches. \Cref{lem:interleave}, with stopping read as this event, therefore bounds the cost of the interleaving by twice the smaller of the two costs.

(c) The instances of this class share the groups other than $W$, the weights, and the set $W\cap\bar P$, so the bounds hold also for methods that know all of these. The class lies in the class of all collections, so a group identified relative to the latter is identified relative to the former, and by \cref{lem:monodec} the bounds hold also when identification is relative to all collections. Write $e=\tilde e(W)$, and let $F'$ be the set of families that lie neither in $\bar P$ nor in a group other than $W$. It contains $W\setminus\bar P$ and has at least $k-|\bar P|-(m-1)w$ families. The instances are the collections $\mathcal{A}_Y$ in which $W$ is replaced by $W_Y=(W\cap\bar P)\cup Y$, with the same weight, for the $e$-subsets $Y$ of $F'$. A permutation of $F$ that maps $W\setminus\bar P$ onto $Y$ and fixes every family outside $F'$ maps $\mathcal{A}$ to $\mathcal{A}_Y$, so $W_Y$ belongs to every certificate of $\mathcal{A}_Y$. Until $W_Y$ is identified, the identified collection lies in $\mathcal{A}_Y\setminus\{W_Y\}$ and is no certificate, so the current decomposition differs from $T^*(\mathcal{A}_Y)$. A method for this class yields a method for the single hidden group of \cref{lem:dcover,lem:rcover}, whose candidates are the sets $W_Y$ of $|W|$ families. The new method answers each query $T$ of the given method with $g_{\mathcal{A}\setminus\{W\}}(T)+v_W[W_Y\subseteq T]$, which needs only its own answer $[W_Y\subseteq T]$, and it queries $W_Y$ once the given method has identified it. The two lemmas, with $|W|$ in place of $w$, therefore bound the cost of the given method from below by their bounds minus $\varphi(|W|)$. Under the budget, a query contains at most $\binom{s}{e}$ candidates. With $\varphi=1$ on queries of at most $s$ families, $\delta=1/\binom{s}{e}$, and no large queries, the lemmas give at least $(\binom{|F'|}{e}-1)/\binom{s}{e}$ queries for deterministic methods and $\binom{|F'|}{e}/(2\binom{s}{e})-1$ in expectation for randomized ones, both $\Omega(k^e)$. Under evaluation costs, a query $T$ with $b=|T\cap F'|$ contains at most $\binom{b}{e}\le b^e/e!$ candidates and costs at least $b^d$. The argument of \cref{thm:power}(b), with these $\binom{|F'|}{e}$ candidates and $|F'|$ in place of $k-w$, then gives $\Omega(k^{\min(e,d)})$, and subtracting the constant $\varphi(|W|)$ keeps this order.

(d) Removing from $T$ the families that lie in no group of $\mathcal{D}$ inside $T$ keeps $g_{\mathcal{D}}(T)$ and lowers $\lambda|T|$, so every maximizer of $g_{\mathcal{D}}(T)-\lambda|T|$ is a union of groups of $\mathcal{D}$. For pairwise disjoint groups, the union of the groups $W_i$ with $i$ in a set $I$ has objective $\sum_{i\in I}a_i$. Hence $T^*(\mathcal{D})$ is the union of the groups of $\mathcal{D}$ with $a_i\ge0$, and it equals $T^*(\mathcal{A})$ exactly when $\mathcal{D}$ contains $\mathcal{B}^*$. For groups that share $\hat f$, the sets $W_i\setminus\{\hat f\}$ are nonempty and pairwise disjoint, the union of the groups with $i$ in a nonempty set $I$ has objective $\sum_{i\in I}v_i-\lambda(1+\sum_{i\in I}(|W_i|-1))=\sum_{i\in I}a_i-\lambda$, and $\emptyset$ has objective $0$. If $\sum_{i:a_i>0}a_i<\lambda$, every nonempty union has a negative objective, so $T^*(\mathcal{A})=\emptyset=T^*(\emptyset)$. Otherwise, $T^*(\mathcal{A})$ is the union of the groups with $a_i\ge0$, every collection that contains $\mathcal{B}^*$ has the same largest maximizer, and a collection that omits some $W_j\in\mathcal{B}^*$ has a largest maximizer disjoint from $W_j\setminus\{\hat f\}$, since no other group meets this set. In both cases, the certificates are the collections that contain $\mathcal{B}^*$, so $e^*=\max_{W\in\mathcal{B}^*}\tilde e(W)$, and every $W\in\mathcal{B}^*$ satisfies the assumptions of (c), because its families outside $\bar P$ lie in no other group. Part (c) for a group of $\mathcal{B}^*$ with $\tilde e(W)=e^*$ gives lower bounds of the orders of (a) and (b).
\end{proof}

\FloatBarrier
\section{Groups Whose Gains Overlap}\label[appendix]{app:overlap}

The gains of \cref{app:cost} add up over groups, and those of \cref{fig:lattice} do not: BS and CS each close 63 percent of the gap, and together they close all of it. Such gains are described by a finite set $\mathcal U$ of \emph{units}. Each unit $u$ has a weight $\omega_u>0$ and a nonempty collection $\mathcal G_u$ of nonempty sets of families, and a set $T$ \emph{closes} $u$ if it contains a member of $\mathcal G_u$. The gain of the units is
\begin{equation}\label{eq:gU}
g_{\mathcal U}(T)=\sum_{u\in\mathcal U:\,T\text{ closes }u}\omega_u .
\end{equation}
A unit is a part of the integrality gap that every member of $\mathcal G_u$ closes on its own. When every $\mathcal G_u$ has one member, $g_{\mathcal U}$ is an additive gain $g_{\mathcal A}$ of \cref{app:cost}. With BS and CS as its groups, the gain function of \cref{fig:lattice} has three units. The first, with 26 percent of the gap, is closed by either pair, and the other two, with 37 percent each, are closed by BS and not CS, and by CS and not BS. A \emph{transversal} of $\mathcal G_u$ is a set of families that meets every member of $\mathcal G_u$.

\begin{lemma}[realization]\label{lem:realize}
Let $\mathcal U$ be a finite set of units with rational weights. There is a binary program with $k$ constraint families whose gain function is $g_{\mathcal U}$. It is the edge formulation of stable set on one odd cycle $C_u$ for each unit and one further edge for each family, all on disjoint sets of vertices. Its objective is $\min-\sum_vc_vx_v$, with $c_v=2\omega_u$ for the vertices of $C_u$ and $c_v=1$ for the other vertices. The edges of $C_u$ are split into one set of consecutive edges for each inclusion-minimal transversal of $\mathcal G_u$, and every family of that transversal contains a copy of the rows of these edges. Each family also contains the row of its further edge.
\end{lemma}

\begin{proof}
Let $K_{u,1},\dots,K_{u,n_u}$ be the inclusion-minimal transversals of $\mathcal G_u$, and give $C_u$ an odd number of at least $\max(3,n_u)$ vertices, so that its edges split into $n_u$ nonempty sets of consecutive edges. The $j$-th set is copied into every family of $K_{u,j}$. A set $T$ closes $u$ exactly when it meets every $K_{u,j}$: if $T$ contains a member of $\mathcal G_u$, every transversal meets that member; otherwise, $F\setminus T$ meets every member of $\mathcal G_u$ and contains an inclusion-minimal transversal, which $T$ misses. Every row lies in one component, so, as in the proof of \cref{thm:multi}(a), $z(T)$ is the sum over the components of the optimal values of the restrictions of $P_T$. The rows of a forest, with repeated rows and the bounds, form a totally unimodular system, so their relaxation is the convex hull of its binary points. The further edges therefore contribute the same value to every $z(T)$. If $T$ closes $u$, the rows of $T$ on $C_u$ contain every edge of $C_u$, the restriction of $P_T$ to $C_u$ is the stable-set polytope of $C_u$, and its value is $-\omega_u(|C_u|-1)$. Otherwise, $T$ misses some $K_{u,j}$, the rows of $T$ on $C_u$ miss the $j$-th set of edges and form a forest, and the restriction of $P_T$ to $C_u$ is the LP relaxation of $C_u$, with value $-\omega_u|C_u|$. Hence $g(T)$ is the total weight of the units that $T$ closes.
\end{proof}

\begin{proof}[Proof of \cref{prop:gain}]
The gain function of every program is nondecreasing with $g(\emptyset)=0$ (\cref{sec:structure}). It takes rational values. Fixing the integer variables at each of the finitely many integer points within their bounds leaves a rational polytope, so $\conv(X_T)$ is the convex hull of finitely many rational polytopes. Hence $P_T\supseteq\conv(X)\neq\emptyset$ is a rational polytope, and $z(T)$ is rational. Conversely, let $g$ be nondecreasing with rational values and $g(\emptyset)=0$, let $0<t_1<\dots<t_r$ be its positive values, and let $t_0=0$. Each set $\mathcal L_j=\{T : g(T)\ge t_j\}$ contains every superset of its members and does not contain $\emptyset$, and $g(T)=\sum_{j=1}^r(t_j-t_{j-1})[T\in\mathcal L_j]$. A unit with weight $t_j-t_{j-1}$, whose collection consists of the inclusion-minimal members of $\mathcal L_j$, is closed by $T$ exactly when $T\in\mathcal L_j$. With one such unit for each $j$, $g=g_{\mathcal U}$, and \cref{lem:realize} gives the program.
\end{proof}

By \cref{prop:gain}, every lower bound that holds on a class of nondecreasing set functions with rational values and $g(\emptyset)=0$ holds on the gain functions of binary programs. The program of \cref{lem:realize} can be large, since $\mathcal G_u$ can have many inclusion-minimal transversals; in the programs of \cref{prop:overlapHard}, every unit has two.

\begin{proposition}[prices with overlapping gains]\label{prop:overlapHard}
Let $n\ge1$, let $S_1,\dots,S_{k-1}$ be subsets of $\{1,\dots,n\}$ whose union is $\{1,\dots,n\}$, call a set $I$ of indices a cover if $\bigcup_{i\in I}S_i=\{1,\dots,n\}$, and let $\sigma$ be the least size of a cover. Let $F=\{\hat f,f_1,\dots,f_{k-1}\}$, let each $u\in\{1,\dots,n\}$ be a unit with $\omega_u=1$ and $\mathcal G_u=\bigl\{\{\hat f,f_i\} : u\in S_i\bigr\}$, and let $\lambda=1/(k+1)$. Then
\[
\max_{T\subseteq F}\bigl(g_{\mathcal U}(T)-\lambda|T|\bigr)=n-\lambda(1+\sigma),
\]
and the maximizers are the sets $\{\hat f\}\cup\{f_i : i\in I\}$ for the covers $I$ of size $\sigma$. The program of \cref{lem:realize} for these units has $3n+2k$ vertices, and every bound $z(T)$ is computable in time polynomial in its size. Choosing the price-optimal decomposition is therefore NP-hard for overlapping gains, even when every unit is known and the groups have two families each, share one family, and are otherwise disjoint.
\end{proposition}

\begin{proof}
A set that omits $\hat f$ closes no unit, so its objective is at most $0$. The set $T=\{\hat f\}\cup\{f_i : i\in I\}$ closes the units of $\bigcup_{i\in I}S_i$, and its objective is $|\bigcup_{i\in I}S_i|-\lambda(1+|I|)$, with $0<\lambda(1+|I|)\le\lambda k<1$. If $I$ is a cover, the objective exceeds $n-1$; otherwise, it is less than $n-1$, and every set that omits $\hat f$ has objective at most $0\le n-1$. The maximizers are therefore the sets for the covers of least size, and the maximum is $n-\lambda(1+\sigma)$. Deciding whether a cover of size at most $t$ exists is NP-complete \citep{karp1972reducibility}, and it amounts to deciding whether the maximum is at least $n-\lambda(1+t)$. The inclusion-minimal transversals of $\mathcal G_u$ are $\{\hat f\}$ and $\{f_i : u\in S_i\}$, so $C_u$ is a triangle split into a path of one edge, which $\hat f$ contains, and a path of two edges, which every $f_i$ with $u\in S_i$ contains. The program has $3n$ vertices on the triangles, $2k$ on the further edges, and $n+2\sum_i|S_i|+k$ rows. As in the proof of \cref{prop:dks}, $P_T$ is the product of the LP relaxations of the further edges and of the triangles that $T$ does not close, and of the stable-set polytopes of the triangles that $T$ closes. Each of these stable-set polytopes is described by nonnegativity and one inequality $x_a+x_b+x_c\le1$. Hence $z(T)$ is the value of an LP of polynomial size.
\end{proof}

With additive gains, groups that share $\hat f$ and are otherwise disjoint give the price-optimal set in closed form (\cref{thm:priceSearch}(d)), and every additive gain is maximized under prices by one minimum cut (\cref{prop:price}(b)). With overlapping gains, the same structure encodes set cover.

\paragraph{Searching for groups whose gains overlap.}
The units are now unknown. The members of the collections $\mathcal G_u$ form the set $\mathcal G$ of \emph{groups}. There are $m$ groups, each has at most $w$ families, and no group contains another. The groups are then exactly the inclusion-minimal sets of positive gain, and each group of at least two families is a zero-marginal group at $\emptyset$. A method knows $w$. Under the budget, every query has $|T|\le s$ with $w\le s<k$ and costs $1$, and the optimal sets are the sets of maximal gain over $\cT_s$. Under evaluation costs, a query of $T$ costs $|T|^d$ with $d>0$, and the optimal sets are the maximizers of $g(T)-\lambda|T|$ for a price $\lambda>0$. In both cases, the cost of a method is the cost of its queries until it has queried an optimal set, as in \cref{sec:cost}, and it is zero when $\emptyset$ is optimal. In \cref{thm:priceSearch}, the decomposition is instead computed from the identified groups. Here, this is not possible, because the gain of a union of groups found in different queries depends on how their units overlap, which the answers to the subsets of each query do not reveal. If a family of an optimal set $T$ lies in no group inside $T$, removing it keeps the gain, so the smaller set is again optimal under the budget and has a larger objective under a price. Hence every inclusion-minimal optimal set is a union of groups, and under a price every optimal set is one. A collection $\mathcal B\subseteq\mathcal G$ is \emph{optimal} when its union is an optimal set. With $\bar P$ and $\tilde e$ as in \cref{app:cost}, let $\bar e$ be the minimum, over the optimal collections $\mathcal B$, of $\max_{W\in\mathcal B}\tilde e(W)$, where the maximum over the empty collection is $0$.

\begin{theorem}[searching with overlapping gains]\label{thm:overlap}
Let $m$, $w$, $s$, $d$, and $|\bar P|$ be fixed, and let $k\to\infty$.
\begin{enumerate}[label=(\alph*),leftmargin=*,itemsep=1pt,topsep=2pt]
\item \emph{Staged search.} Make the stage queries of \cref{thm:priceSearch}(a). After every stage query of positive gain, query the nonempty subsets of that query with at most $w$ families; the inclusion-minimal sets of positive gain among them are groups, which are then \emph{found}. After every newly found group, query the unions of found groups that contain it, with at most $s$ families under the budget. Without knowing $\bar e$, this search queries an optimal set within $O(k^{\bar e})$ queries under the budget, and with $s=w$ at cost $O(k^{\bar e})$ under evaluation costs.
\item \emph{Evaluation cost.} The cost-balanced interleaving of the search of (a), with $s=w$, and the halving search of the proof queries an optimal set at cost $O(k^{\min(\bar e,d)})$.
\item \emph{Lower bound.} Let a group $W$ lie in every optimal set, let no family of $W\setminus\bar P$ lie in another group, and let $\tilde e(W)\ge1$. Consider the class of instances obtained by replacing $W\setminus\bar P$, in every $\mathcal G_u$ that contains $W$, with a set of $\tilde e(W)$ families that lie neither in $\bar P$ nor in another group. On this class, every deterministic method needs $\Omega(k^{\tilde e(W)})$ queries under the budget and cost $\Omega(k^{\min(\tilde e(W),d)})$ under evaluation costs in the worst case, and every randomized method needs as much in expectation on some instance.
\item \emph{Tight cases.} Suppose that the groups are pairwise disjoint, or share one family $\hat f\in\bar P$, each have a family besides $\hat f$, and are otherwise pairwise disjoint, and let $\bar e\ge1$. Consider the instances obtained by permuting the families that lie neither in $\bar P$ nor in a group $W$ with $\tilde e(W)<\bar e$. On them, every deterministic method needs $\Omega(k^{\bar e})$ queries under the budget and cost $\Omega(k^{\min(\bar e,d)})$ under evaluation costs in the worst case, and every randomized method needs as much in expectation on some instance. Hence the order of (a) under the budget and the order of (b) are tight. If there is a unique inclusion-minimal optimal set $T^\circ$, every optimal collection contains the groups inside $T^\circ$, and $\bar e$ is the largest $\tilde e(W)$ over these groups.
\end{enumerate}
\end{theorem}

Under the budget, $\bar e$ is set by the optimal collection that the predictions cover best, as the $r$-th best prediction sets the cost in \cref{thm:multi}.

\begin{proof}[Proof of \cref{thm:overlap}]
(a) A subset $S$ of a query $T$ with $|S|\le w$ is an inclusion-minimal set of positive gain exactly when $g(S)>0$ and $g(S\setminus\{x\})=0$ for every $x\in S$, and all these sets are queried. Such a set contains a group, and no proper subset of it contains one, so it equals that group. Conversely, every group inside $T$ is such a set, since it has positive gain and its proper subsets contain no group. Let $\mathcal B$ be an optimal collection with $\max_{W\in\mathcal B}\tilde e(W)=\bar e$; if $\mathcal B$ is empty, $\emptyset$ is optimal, and the cost is $0$. As in the proof of \cref{thm:priceSearch}(a), every $W\in\mathcal B$ lies in some query of stage $\tilde e(W)$, which has positive gain because $W$ does, so $W$ is found by the end of stage $\bar e$. When the last group of $\mathcal B$ is found, the union of $\mathcal B$ is among the unions queried next, since under the budget it has at most $s$ families. Hence the search queries an optimal set by the end of stage $\bar e$. Stage $q$ makes at most $2^{|\bar P|}\binom{k}{q}$ queries, each followed by at most $\sum_{j=1}^{w}\binom{s}{j}$ queries of subsets, and at most $2^m$ unions are queried in all, so the search makes $O(k^{\bar e})$ queries. With $s=w$, every query except the unions has at most $w$ families and every union at most $mw$, so the cost under evaluation costs is also $O(k^{\bar e})$.

(b) The halving search runs passes and uses only the groups $\mathcal D$ found from its own answers. A pass queries $F\setminus H$ for every set $H$ obtained by choosing one family from each group of $\mathcal D$, at most $w^m$ sets, with $H=\emptyset$ when $\mathcal D$ is empty, until one of these queries has positive gain. If none has, the search ends. Otherwise, for the first such $H$, it sets $Z=F\setminus H$. While $|Z|>4w$, it splits $Z$ into $2w$ blocks whose sizes differ by at most one and, for each block $B$ in turn, queries $Z\setminus B$ and replaces $Z$ by $Z\setminus B$ if $g(Z\setminus B)>0$. When $|Z|\le4w$, it queries the nonempty subsets of $Z$ with at most $w$ families, finds the groups inside $Z$ as in (a), queries the unions of found groups that contain a new group, and starts the next pass.

Every group of $\mathcal D$ has a family in $H$, so no subset of $F\setminus H$ contains a group of $\mathcal D$, and a subset of positive gain contains a group outside $\mathcal D$. The set $Z$ keeps a positive gain, so at the end of a round it contains such a group $W$. Every block kept in the round meets $W$, because, when it was tested, the current set contained $W$ and the current set without the block had gain $0$. Hence at most $w$ blocks are kept, and the round ends with at most $w\lceil|Z|/(2w)\rceil\le|Z|/2+w<3|Z|/4$ families. Every pass that does not end the search therefore finds a group outside $\mathcal D$. If a group $W'$ lies outside $\mathcal D$, no group of $\mathcal D$ is contained in $W'$, so choosing for each group of $\mathcal D$ a family outside $W'$ gives a set $H$ with $g(F\setminus H)\ge g(W')>0$. The search therefore ends only after it has found every group, after at most $m+1$ passes, and by then it has queried the union of every collection of groups, among them an optimal collection. Round $t$ of a pass makes $2w$ queries of at most $(3/4)^tk$ families, so a pass costs at most $w^mk^d+2wk^d/(1-(3/4)^d)+2^{4w}(4w)^d$, and the unions cost at most $2^m(mw)^d$ in all. The halving search thus queries an optimal set at cost $O(k^d)$. In the interleaving, each search uses only its own answers and makes the same queries as when it runs alone. \Cref{lem:interleave}, with stopping read as the query of an optimal set, therefore bounds the cost by twice the smaller of $O(k^{\bar e})$ and $O(k^d)$.

(c) Write $e=\tilde e(W)$, and write $\varphi(|T|)$ for the cost of a query of $T$, which is $1$ under the budget and $|T|^d$ under evaluation costs. Let $F'$ be the set of families that lie neither in $\bar P$ nor in a group other than $W$. It contains $W\setminus\bar P$ and has at least $k-|\bar P|-(m-1)w$ families. For an $e$-subset $Y$ of $F'$, let $\mathcal U_Y$ be the instance in which $W$ is replaced by $W_Y=(W\cap\bar P)\cup Y$ in every $\mathcal G_u$, with the same weights. These are the instances of the class. They share everything except $Y$, so the bounds hold also for methods that know all of it. A permutation of $F$ that maps $W\setminus\bar P$ onto $Y$ and fixes every family outside $F'$ maps $\mathcal U$ to $\mathcal U_Y$ and preserves the sizes of sets. The optimal sets of $\mathcal U_Y$ are therefore the images of those of $\mathcal U$, and all of them contain $W_Y$. Let $g_0(T)$ be the total weight of the units $u$ for which $T$ contains a member of $\mathcal G_u$ other than $W$. A set $T$ that does not contain $W_Y$ closes a unit of $\mathcal U_Y$ exactly when it contains a member of $\mathcal G_u$ other than $W$, so $g_{\mathcal U_Y}(T)=g_0(T)$.

Fix a deterministic method, and let $T_1,T_2,\dots$ be its queries when every answer is given by $g_0$. On $\mathcal U_Y$, the method makes these queries up to the first $T_Q$ that contains $W_Y$, and it queries no optimal set before $T_Q$, since every optimal set contains $W_Y$. Call the sets $W_Y$ candidates, and let $h(T)$ be the number of candidates inside $T$. If no index $q$ has every candidate inside one of $T_1,\dots,T_q$, the method never queries an optimal set on some instance. Otherwise, for the least such $q$, the method makes all of $T_1,\dots,T_q$ on the instance whose candidate first lies in $T_q$, so its worst-case cost is at least $\sum_{i\le q}\varphi(|T_i|)\ge\binom{|F'|}{e}\min\{\varphi(|T|)/h(T) : h(T)>0\}$. For randomized methods, the proof of \cref{lem:rcover} applies with $W_Y$ in place of $W$ and the answers of $g_0$ in place of the answer $0$. Under the budget, $h(T)\le\binom{s}{e}$, and every query costs $1$, which gives at least $\binom{|F'|}{e}/\binom{s}{e}$ queries for deterministic methods and $\binom{|F'|}{e}/\bigl(2\binom{s}{e}\bigr)$ in expectation for randomized ones, both $\Omega(k^e)$. Under evaluation costs, a query with $b=|T\cap F'|$ contains at most $\binom{b}{e}\le b^e/e!$ candidates and costs at least $b^d$, and the argument of \cref{thm:power}(b), with $\binom{|F'|}{e}$ candidates and $|F'|$ in place of $k-w$, gives $\Omega(k^{\min(e,d)})$.

(d) Call a group high if $\tilde e(W)\ge\bar e$. Every optimal set contains an inclusion-minimal optimal set, namely an inclusion-minimal one among the optimal sets inside it. The groups inside that set form an optimal collection, which contains a high group by the definition of $\bar e$. Hence every optimal set contains a high group. Let $F''$ be the set of families that lie neither in $\bar P$ nor in a group that is not high. Groups meet at most in $\hat f\in\bar P$, so $F''$ contains $W\setminus\bar P$ for every high group $W$, and it has at least $k-|\bar P|-mw$ families. A permutation of $F$ that fixes every family outside $F''$ fixes $\bar P$ and every group that is not high and preserves the sizes of sets. It maps the instance to one with the same structure and the same $\bar e$, whose optimal sets are the images of the original ones. A query that contains no image of a high group closes a unit exactly when it contains a member of $\mathcal G_u$ that is not high, so its answer does not depend on the permutation. A deterministic method therefore makes a fixed sequence of queries until one of them contains the image of a high group, and it queries no optimal set before that query. For a uniformly random permutation of $F''$, the image of $W\setminus\bar P$ is a uniformly random set of $\tilde e(W)$ families of $F''$. A query $T$ with $b=|T\cap F''|$ thus contains the image of a high group $W$ with probability at most $\binom{b}{\tilde e(W)}/\binom{|F''|}{\tilde e(W)}\le(b/|F''|)^{\bar e}$, and the image of some high group with probability at most $m(b/|F''|)^{\bar e}$. Under the budget, $b\le s$, and as in the proof of \cref{thm:multi}(c), with $J=|F''|^{\bar e}/(ms^{\bar e})$, the expected number of queries exceeds $J/2$, which is $\Omega(k^{\bar e})$ and also bounds the worst case of a deterministic method. Under evaluation costs, the proof of \cref{lem:rcover}, with $m(b_i/|F''|)^{\bar e}$ in place of $h_{\mathcal C}(T_i)/|\mathcal C|$ and $\Delta=(|F''|/2)^d$, gives an expected cost of $\Omega(k^{\min(\bar e,d)})$ by the computation in the proof of \cref{thm:power}(b). Averaging over the random bits of a randomized method gives an instance with at least this expected cost.

If $T^\circ$ is unique, every optimal set contains it, and $T^\circ$ is the union of the set $\mathcal B^\circ$ of groups inside it. For $W\in\mathcal B^\circ$, the set $W$ in the first case and $W\setminus\{\hat f\}$ in the second is nonempty and meets no other group. The union of an optimal collection $\mathcal B$ contains $T^\circ\supseteq W$, so some group of $\mathcal B$ meets this set, and that group is $W$. Hence every optimal collection contains $\mathcal B^\circ$, which is itself optimal, and $\bar e=\max_{W\in\mathcal B^\circ}\tilde e(W)$.
\end{proof}

\begin{proof}[Proof of \cref{prop:combine}]
Part (a) is \cref{prop:dks}, part (b) is \cref{prop:price}(b) to (d), and part (c) is \cref{prop:overlapHard}. \Cref{prop:dks,prop:overlapHard} also state that every bound of their programs is computed in polynomial time.
\end{proof}

\FloatBarrier
\section{Exact Computations and Examples}\label[appendix]{app:exact}

\paragraph{Method.} For every $T$, the polytope $P_T$ is written as one LP on the variables of the MILP and, where needed, auxiliary variables. Each block of $T$ is represented either by the convex hull of its integer points, enumerated over the bounds, or by Balas' extended formulation of the union of the polyhedra of the continuous variables over the integer assignments \citep{balas1985disjunctive}. HiGHS solves the LP, and the optimal basis is recomputed and verified in exact rational arithmetic. The kind of a group is computed from \cref{def:zmg} with one hull formulation for each $P_{W\setminus\{f\}}$ on the same variables.

\begin{example}[LP-geometric groups]\label{ex:geo}
Consider $\min\{-y : y\le x_1+x_2,\ y\le\tfrac12,\ 2x_1\le1,\ 2x_2\le1,\ x\in\{0,1\}^2,\ 0\le y\le1\}$ with four rows as four families, and $z_{\LP}=-\tfrac12$. Convexifying $2x_1\le1$ fixes $x_1=0$, and the LP moves the fractional weight to $x_2$. Each bound row alone therefore has gain $0$, while the two bound rows together have gain $\tfrac12$. Each bound row together with both $y$-rows forms a zero-marginal group of order three with gain $\tfrac14$. The exact computation classifies all three groups as LP-geometric.
\end{example}

\paragraph{Lookahead on the instance of \cref{fig:lattice}.} For the pair of capacity and setup rows, whose gain is $51/10$, the one-pass lookahead is $0$ at the optimal dual returned by HiGHS and about $2.7$ at the optimal dual that maximizes it, in a floating-point computation.

\FloatBarrier
\section{Instances}\label[appendix]{app:templates}

The templates are small instances of five classes with exact gains for every set of families (\cref{tab:exact}). The rich models are larger connected instances of three of these classes with exact gains for every set of at most three families (\cref{tab:rich-inst,tab:rich-large-inst}).

\begin{table}[!htbp]
\caption{Instances with exact ground truth. LP gap: $(z_{\IP}-z_{\LP})/|z_{\IP}|$. Gap closed: the share of the LP gap that convexifying the group closes, computed in exact rational arithmetic and rounded. Every group of order two or three is polyhedral. The generalized assignment instances have no joint support, and their best decomposition convexifies the capacity family alone.}
\label{tab:exact}
\centering
\footnotesize
\setlength{\tabcolsep}{4pt}
\begin{tabular}{@{}llcccc@{}}
\toprule
Class and families & Instance & $k$ & LP gap (\%) & Zero-marginal group & Gap closed (\%) \\
\midrule
Lot sizing & LS $2\times2$ & 3 & 32.4 & BS & 63 \\
B balance, S setup, C capacity & & & & CS & 63 \\
& LS $2\times2$ b & 3 & 34.6 & BS & 60 \\
& & & & CS & 57 \\
& LS $3\times2$ & 3 & 22.0 & BS & 100 \\
& & & & CS & 17 \\
\midrule
Network design & ND a & 3 & 14.7 & NU & 100 \\
N flow balance, L arc linking, U arc capacity & ND b & 3 & 15.2 & NU & 100 \\
& ND c & 3 & 5.9 & NU & 100 \\
\midrule
Facility location & FL $2\times3$ & 3 & 2.4 & AC & 100 \\
A assignment, C capacity, L linking & FL $3\times3$ & 3 & 1.8 & AC & 100 \\
& & & & AL & 25 \\
& FL $2\times4$ & 3 & 1.7 & AC & 100 \\
\midrule
Generalized assignment & GAP $2\times3$ & 2 & 60.0 & none & \\
A assignment, C capacity & GAP $2\times3$ b & 2 & 50.9 & none & \\
\midrule
Hidden group & HG a & 4 & 8.3 & BC & 100 \\
conflict families A to F & HG b & 5 & 8.3 & AD & 100 \\
& HG c & 6 & 7.1 & CE & 100 \\
& HG d & 5 & 16.7 & ACE & 100 \\
\bottomrule
\end{tabular}
\end{table}

\paragraph{Lot sizing.} Items $i$ and periods $t$, continuous production $x_{it}$ and stock $s_{it}$, and binary setups $y_{it}$. The balance family has rows $s_{i,t-1}+x_{it}-s_{it}=d_{it}$. The setup family has rows $x_{it}\le M_{it}y_{it}$, where $M_{it}$ is the minimum of the capacity of period $t$ and the remaining demand of item $i$, and the bounds are $0\le x_{it}\le M_{it}$ and $0\le s_{it}\le\sum_{t'}d_{it'}$. The capacity family has rows $\sum_ix_{it}\le C_t$. The objective is holding plus setup cost. Demands, capacities, holding costs, and setup costs are $((2,3),(2,2))$, $(5,4)$, $(1,1)$, and $(6,6)$ for LS $2\times2$; $((3,2),(1,3))$, $(5,4)$, $(1,2)$, and $(7,5)$ for LS $2\times2$ b; and $((2,1),(1,2),(2,2))$, $(6,4)$, $(1,1,1)$, and $(6,5,7)$ for LS $3\times2$.

\paragraph{Network design.} Two commodities travel on the directed arcs $(1,2)$, $(1,3)$, $(2,3)$, $(2,4)$, and $(3,4)$ of a four-node network. Commodity $q$ has continuous flows $x_{qa}\in[0,d_q]$, and arc $a$ has a binary opening variable $y_a$. The flow-balance family has one conservation row per commodity and node, the arc-linking family has rows $x_{qa}\le d_qy_a$, and the arc-capacity family has rows $\sum_qx_{qa}\le u_ay_a$. The objective is the fixed cost of the opened arcs plus one unit of cost per unit of flow and arc. In ND a, the commodities go from node 1 to node 4 with demand 2 and from node 1 to node 3 with demand 1, every capacity is 2, and the fixed costs are $(4,4,3,4,3)$. ND b has the fixed costs $(6,6,5,6,5)$ and is otherwise equal to ND a. In ND c, the commodities go from node 1 to node 4 with demand 2 and from node 2 to node 4 with demand 1, the capacities are $(3,2,2,2,3)$, and the fixed costs are $(5,4,2,4,5)$.

\paragraph{Facility location.} Continuous $x_{ij}\in[0,1]$ is the share of the demand $d_j$ of customer $j$ served by facility $i$, and binary $y_i$ opens facility $i$. The assignment family has rows $\sum_ix_{ij}=1$, the capacity family has rows $\sum_jd_jx_{ij}\le C_iy_i$, and the linking family has rows $x_{ij}\le y_i$. The objective is the opening cost plus $\sum_{ij}c_{ij}d_jx_{ij}$. Demands, capacities, opening costs, and unit costs are $(2,2,3)$, $(4,4)$, $(5,6)$, and $((1,3,2),(3,1,2))$ for FL $2\times3$; $(2,3,2)$, $(5,4,4)$, $(6,5,5)$, and $((1,2,3),(3,1,2),(2,3,1))$ for FL $3\times3$; and $(3,2,2,1)$, $(5,5)$, $(7,6)$, and $((1,2,3,2),(3,2,1,1))$ for FL $2\times4$.

\paragraph{Generalized assignment.} Binary $x_{ij}$ assigns job $j$ to machine $i$. The assignment family has rows $\sum_ix_{ij}=1$, the capacity family has rows $\sum_ja_{ij}x_{ij}\le b_i$, and the objective is the assignment cost. Weights, capacities, and costs are $((5,5,4),(5,5,4))$, $(9,9)$, and $((1,1,10),(10,10,1))$ for GAP $2\times3$, and $((4,5,3),(5,4,3))$, $(8,7)$, and $((2,1,9),(9,8,1))$ for GAP $2\times3$ b.

\paragraph{Hidden group.} These are the programs of \cref{thm:cover}(a) with the objective $\max\sum_vx_v$. In HG a, with $k=4$, an odd cycle of length 5 is split into two paths, the hidden families B and C. In HG b and HG c, with $k=5$ and $k=6$, a triangle is split into two paths, the hidden families A and D, and C and E. In HG d, with $k=5$, a triangle is split into its three edges, the hidden families A, C, and E. The other families are paths of the same lengths on new vertices. Every family has a name of the form Conflict followed by a letter, and all families have the same one-sentence description.

\paragraph{Rich models.} Each rich model is a single connected program whose families are standard modeling features on shared variables, with data small enough that every set of at most three families can be evaluated exactly with the method of \cref{app:exact}. The lot-sizing models have two products, three periods, six binary setups, and the families Balance, SetupLink, Capacity with setup times and overtime, MinLot, StorageCap, SafetyStock, MaxSetupsPerPeriod, OvertimeLimit, and SetupBudget. The network-design models have two commodities, four nodes and seven arcs or five nodes and eight arcs, one binary per arc, and the families FlowBalance, ArcLink, ArcCapacity, BuildBudget, NodeOutDegree, OppositeArcs, and InArcCover. The facility-location models have three facilities with binary opening variables, three or four customers with split demand, and the families Assign, Capacity, Link, OpenBudget, OpenCount, MinUtilization, and FacilityConflict. Data are drawn at random until eight draws per class are accepted. A draw is rejected if it is infeasible, if its LP gap is below two percent, or if a single family closes more than half of the best gain among sets of at most three families. \Cref{tab:rich-inst} summarizes the accepted instances. Every zero-marginal group is polyhedral and connected through shared variables, and in every instance the best set of at most three families closes the whole LP gap.

\begin{table}[!htbp]
\caption{Rich models: number of families $k$, size, LP gap, the minimal best sets among the sets of at most three families with the number of instances in parentheses, two lot-sizing instances having two such sets, and the number of zero-marginal groups per instance. B: Balance, S: SetupLink, C: Capacity, T: StorageCap, Y: SafetyStock; F: FlowBalance, U: ArcCapacity; A: Assign, P: Capacity.}
\label{tab:rich-inst}
\centering
\footnotesize
\setlength{\tabcolsep}{3pt}
\begin{tabular}{@{}lccccc@{}}
\toprule
Class & $k$ & rows $\times$ columns & LP gap (\%) & best sets & groups \\
\midrule
Lot sizing & 9 & $35\times21$ & 15.2--51.7 & BS (4), BCS (3), BST (1), BSY (2) & 2--4 \\
Network design & 7 & $37\times21$, $43\times24$ & 2.8--30.1 & FU (8) & 1 \\
Facility location & 7 & $21\times12$, $25\times15$ & 5.5--28.5 & AP (8) & 1--2 \\
\bottomrule
\end{tabular}
\end{table}

\paragraph{Rich models with 12 to 16 families.} These models keep the families of the smaller rich models and add standard modeling features on the same variables. Lot sizing adds ProductStorageCap, ProductionCap, and TotalOvertime for $k=12$, and ProductionBudget, WorkforceLimit, HoldingBudget, and NoConsecutiveSetups in addition for $k=16$. Network design adds NodeInDegree, NodeThroughput, CommodityArcCap, MinBuiltArcs, and FlowCostBudget for $k=12$, and ArcConflict and OutArcCover in addition for $k=14$. Facility location adds MaxCustomersPerFacility, DistanceLimit, RegionCover, TotalCapacity, and TransportBudget for $k=12$, and SplitLimit, MinAssignment, FacilityPairCover, and TransportPreference in addition for $k=16$. Data are drawn and accepted as for the smaller models, ten instances per class and number of families. Gains are exact for every set of at most three families, and of at most two on the network-design models with 14 families (\cref{tab:rich-large-inst}). The best pair of these models is not always the classical one. On network design and facility location, pairs with a new family often have the largest gain within the budget, and on five facility-location models no set of at most two families has a positive gain.

\begin{table}[!htbp]
\caption{Rich models with 12 to 16 families: number of families $k$, rows, columns, LP gap, mean number of zero-marginal groups per instance, and number of instances on which some set of at most two families has a positive gain.}
\label{tab:rich-large-inst}
\centering
\footnotesize
\setlength{\tabcolsep}{4pt}
\begin{tabular}{@{}lcccccc@{}}
\toprule
Class & $k$ & rows & columns & LP gap (\%) & groups & positive gain, $s=2$ \\
\midrule
Lot sizing & 12 & 48 & 21 & 26.5--54.5 & 3.4 & 10 \\
 & 16 & 58 & 21 & 34.3--53.2 & 5.6 & 10 \\
Network design & 12 & 61--71 & 21--24 & 7.5--36.1 & 3.5 & 10 \\
 & 14 & 65--77 & 21--24 & 2.5--23.5 & 3.2 & 10 \\
Facility location & 12 & 29--34 & 12--15 & 2.1--11.4 & 2.9 & 7 \\
 & 16 & 42--50 & 12--15 & 2.5--17.0 & 9.7 & 8 \\
\bottomrule
\end{tabular}
\end{table}

\paragraph{Benchmark instances.} The benchmark instances use the published formulation and data of each class, with no added family and no generated parameter (\cref{tab:bench-inst}). Lot sizing uses the capacitated instances with setup times of \citet{trigeiro1989capacitated}, with the families balance, setup linking, and capacity with setup times, and no overtime. The 20 smallest usable instances by items times periods were kept. Network design uses the R instances of multicommodity capacitated fixed-charge network design \citep{crainic2001bundle}, with the families flow balance, arc linking with $x_{qa}\le\min(d_q,u_a)\,y_a$, and arc capacity. Facility location uses the capacitated instances cap41 to cap134 of the OR-Library \citep{beasley1990orlibrary}, with split demand and the families assignment, capacity, and linking $x_{ij}\le y_i$. For the last two classes, every instance whose LP is feasible and whose LP gap against the SCIP optimum is positive was kept, up to the 20 smallest by number of variables; 19 of the 37 facility-location instances have a positive LP gap. Every instance has an optimum proved by SCIP. The ground truth evaluates every set of one or two families with GCG at the root node under the settings of \cref{app:miplib}. Every instance has a certain zero-marginal group at $\emptyset$, and in every instance the best pair is the classical pair of its class: balance with setup linking, flow balance with arc capacity, and assignment with capacity. Because the original presentation shows no coefficients, instances of one class with the same family tables share a prompt: the 59 instances give eight distinct prompts, one for lot sizing, four for network design, and three for facility location.

\begin{table}[!htbp]
\caption{Benchmark instances: number of instances, size, LP gap as $(z_{\IP}-z_{\LP})/|z_{\IP}|$, and the best pair of families, which is a certain zero-marginal group at $\emptyset$ in every instance.}
\label{tab:bench-inst}
\centering
\scriptsize
\setlength{\tabcolsep}{4pt}
\begin{tabular}{@{}llcccl@{}}
\toprule
Class & Data & Instances & rows $\times$ columns & LP gap (\%) & Best pair \\
\midrule
Lot sizing & \citet{trigeiro1989capacitated} & 20 & $195\times270$ & 63.9--74.4 & balance, setup linking \\
Network design & R set \citep{crainic2001bundle} & 20 & $485\times385$ to $1160\times910$ & 0.05--10.6 & flow balance, arc capacity \\
Facility location & OR-Library cap \citep{beasley1990orlibrary} & 19 & $866\times816$ to $2600\times2550$ & 0.006--0.42 & assignment, capacity \\
\bottomrule
\end{tabular}
\end{table}

\paragraph{Instances without a joint support.} The generalized assignment instances are the 60 instances gap1 to gap12 of the OR-Library, with the families assignment and capacity; their published objective is maximized, and we minimize its negative. The bin-packing instances are the 16 of the 20 instances of binpack1, with 120 items and bins of capacity 150, for which SCIP returns a reference value within one hour. They use one binary per item and bin and one per bin, with the number of bins set by first-fit decreasing, and the families assignment, capacity, and symmetry breaking $y_j\ge y_{j+1}$. The reference optimum of an instance is the best value among the SCIP run, the branch-and-price runs, and, for bin packing, the best-known count of the OR-Library file. On every bin-packing instance, this value equals the lower bound $\lceil\sum_iw_i/C\rceil$ and is therefore optimal. No instance has a joint support within the budget of $k-1$ families. On 18 generalized assignment instances, the capacity family alone closes the whole gap. On the others, the rest of the gap is closed only by assignment with capacity, which convexifies the whole program on generalized assignment and all of it but the symmetry-breaking rows on bin packing, so this pair lies outside the budget. On four bin-packing instances, no set within the budget has a positive gain, and these are not scored for naming the set with the best bound. Branch-and-price runs on the sets that some method chooses and on GCG as shipped, and the time-optimal set is the first of the sets within the budget among them.

\FloatBarrier
\section{Prompt and Model Settings}\label[appendix]{app:prompt}

\paragraph{Prompt.} Every request consists of one system message and one user message. The user message defines convexification, a family that strengthens the bound alone, a joint support, and the integrality property, and explains how a reformulation is specified. It then gives the description, the formulation, the model statistics, and the two family tables, with no bound or optimal value. Sections of a description that mention an optimal or best-known value are removed. In the presentations names only, permuted, and anonymized, the description and the formulation are replaced by the word withheld. In the permuted and anonymized presentations, the family names in the tables are permuted among the families or replaced by $F_1,F_2,\dots$ for the constraint families and $V_1,V_2,\dots$ for the variable families. The system message is
\begin{scriptsize}
\begin{verbatim}
You analyze mixed-integer linear programs and design Dantzig-Wolfe reformulations. You answer with one JSON
  object and nothing else.
\end{verbatim}
\end{scriptsize}
and the user message, with each inserted field written as a name in double angle brackets, is
\begin{scriptsize}
\begin{verbatim}
## Task
The mixed-integer linear program (MILP) below is correct as given. You receive its natural-language
  description and formulation when available, and the constraint families and variable families of the solver
  model. Answer three questions about its constraint families, then propose Dantzig-Wolfe (DW) reformulations.

Definitions. Convexifying a set T of constraint families means replacing the LP relaxation by its intersection
  with the convex hull of the integer points that satisfy the constraints of T (with the variable bounds and
  integrality). This is the bound of a DW reformulation whose blocks are the constraints of T.
1. A family "strengthens alone" if convexifying that family by itself gives a bound strictly better than the
  LP relaxation for this model and objective.
2. A "joint support" is a set of two or more families that together strengthen the bound while no proper
  subset of them does: some valid inequalities that cut off LP solutions can only be derived from all of these
  families together.
3. A family "has the integrality property" if convexifying it alone can never improve on the LP relaxation,
  for any objective.

A reformulation is specified by `master_constraint_families` (kept in the master) and
  `linking_variable_families` (variable families copied into each block, with the copies equated in the
  master; leave empty unless intended). Every constraint not in the master is convexified; two constraints are
  in the same block when they share a variable that is not a linking variable; each block is a pricing problem
  solved by a MIP solver at every column-generation iteration, so a block that contains most of the model is
  as hard as the model itself. Proposals are solved at the root node without cutting planes with a time limit
  of <<time_limit>> seconds and scored by the dual bound at the end of the root node.

## Problem description
<<description>>

## Formulation
<<formulation>>

## Model statistics
rows <<num_rows>>, columns <<num_columns>>, objective: <<sense>>

## Constraint families
Columns: family | rows | sense counts | MIPLIB constraint types | row length min/median/max | variable
  families touched (nonzeros) | all coefficients equal to 1
<<constraint_table>>

## Variable families
Columns: family | columns | types (B binary, I integer, C continuous) | columns with nonzero objective
  coefficient
<<variable_table>>

## Output
Return only a JSON object of this form:
{
 "families": {"<constraint family>": {"integrality_property": "has | lacks | unclear", "strengthens_alone":
   "yes | no | unclear"}},
 "joint_supports": [
  {"families": ["...", "..."], "inequality_class": "<name of the valid inequalities they generate, or
    'unknown'>", "reason": "<at most 40 words>"}
 ],
 "assessment": "<at most 100 words>",
 "decompose": true,
 "candidates": [
  {"master_constraint_families": ["..."],
    "linking_variable_families": [],
    "block_problem": "<one pricing problem, in words>",
    "expected_pricing_difficulty": "easy | moderate | hard"}
 ]
}
List every constraint family under "families". List joint supports most important first (an empty list if you
  expect none). Give up to <<k>> candidates, best first, all different. Set "decompose" to false if you expect
  that no reformulation will beat the LP bound within the time limit; still list your best candidates. Use
  family names exactly as listed (a variable family written as `name:B`, `name:I` or `name:C` contains only
  the binary, integer or continuous columns of that name pattern). Every candidate must leave at least one
  constraint family out of the master, and must either keep at least one constraint family in the master or
  name at least one linking variable family.
\end{verbatim}
\end{scriptsize}
The time limit is 120 seconds, and the number of candidates is three.

\paragraph{Prompt variants.} The coefficient variant inserts after the formulation a section headed \texttt{\#\# Model with data}. It holds the complete instance in CPLEX LP format, written by SCIP with the row and column names of the family tables and without the header comment that names the instance. On the rich models with seven to nine families, these prompts have at most about 3300 tokens, and on the benchmark instances between about 12,000 and 230,000 (\cref{tab:app-bench-data}). The two control presentations with coefficients, permuted and anonymized, permute the family names among the families and replace them by $F_1,F_2,\dots$ for the constraint families and $V_1,V_2,\dots$ for the variable families in the family tables and in the model with data alike. Each instance has its own permutation and label map, the same for the two configurations run with these controls. After the permutation, the pair that carries the classical names is a true group in no instance.

\paragraph{Parsing of answers.} Each family named in an answer is matched to a shown name after block suffixes and indices are removed. A name that matches no shown name or more than one is dropped, and an answer whose JSON object does not parse counts as a failed prediction.

\paragraph{Models.} The five configurations are Claude Sonnet 5.5 (Anthropic) with no request option set, so that the default thinking of the endpoint applies, GPT-6.1 Sol (OpenAI) with medium reasoning effort, Grok 4.20 (Reasoning) and Grok 4.20 (Non-Reasoning) (xAI), and DeepSeek-V4-Pro (DeepSeek) in non-thinking mode. Where the endpoint of GPT-6.1 Sol reports a snapshot, it is gpt-6.1-sol-2026-09-29. Every request asks for a JSON object, and no request sets a temperature or a completion limit. The endpoint of Claude Sonnet 5.5 limits a completion, thinking included, to 20{,}000 tokens. Of its 1404 answers on the templates and MIPLIB-NL, 1384 reach this limit, and all of them except one, which does not parse, contain a complete answer object. The three samples of a cell share the prompt and differ only through the sampling of the model. All answers were produced between 30 October and 9 November 2026. The latency of a request is its wall-clock time, and its cost follows from the token counts reported by the endpoint and the list prices on the day of the request. On the templates and MIPLIB-NL, 31 answers do not parse, 1 of Claude Sonnet 5.5 and 30 of DeepSeek-V4-Pro.

\paragraph{Computing environment.} All computations ran on one workstation with a 24-core Intel processor and 48 GB of memory, using SCIP~10 with GCG and HiGHS.

\FloatBarrier
\section{MIPLIB-NL Processing and Solver Settings}\label[appendix]{app:miplib}

Each MIPLIB-NL instance comes with a script that builds the model in a solver interface, from which we export the model to MPS. Of the 223 instances, 205 are kept. The others fail to export, or export a model that is infeasible or whose optimum differs from the reference. Constraint families are the row names with their digit runs removed. Of the 205 models, 57 have fewer than two families, 34 have more than six, and 6 have general constraints, which leaves 108 for the census.

GCG runs on SCIP~10 with one thread at the root node only, with presolving, separation, primal solution procedures, and propagation disabled in the original and in the master problem, so that every bound comes from the relaxation alone. Each evaluation has a time limit of 120 seconds, and an evaluation that repeatedly ends abnormally counts as having no bound. Variable families are defined from the column names in the same way. On the 72 models whose constraint families and variable families number at most six together, each set is also evaluated with every nonempty set of variable families declared as linking variables, which several blocks may then share. The census reports the largest gains and the zero-marginal groups, certain or possible, over all of these choices, and the predictions and the search orders are scored on the evaluations without linking variables. The bounds of instances with an integral objective are rounded before comparison. The LP bound comes from SCIP and, for two models, from the interior-point method of HiGHS with crossover. Without linking variables, convexifying all $k$ families leaves an empty master and solves the MILP itself, so these evaluations use the sets of at most $k-1$ families. The lookahead gives each block 60 seconds and each instance 3600 seconds. The five-dual lookahead adds four optimal duals obtained by maximizing random objectives over the optimal face of the dual, with bounded dual values, and takes for each set the largest of the five values. On 90 of the 102 scored models, at least one of the four duals differs from the HiGHS dual. A recorded bound is discarded when it is not finite, or when it lies beyond the reference optimum, or beyond a solution value that SCIP found, by more than $10^{-4}$ times the largest of the LP gap, the absolute value of the optimum, and one. Of the 3122 family-set evaluations of the census, 46 percent converged, 32 percent stopped at the time limit with a valid bound, and 21 percent returned no bound. Six of the 108 models are not scored. Five of them have no set with a bound among the evaluations without linking variables, and for the sixth, which has no certain group, no answers were collected. GCG's own detected decompositions, used as anchors of the staged search, need 2.3 to 3.5 evaluations on average to reach a best set on the 11 to 13 models where they exist. The branch-and-price runs use GCG with its default settings, one thread, three seeds, and a limit of 1800 seconds on the given reformulation. A run counts as unsolved at the limit if it repeatedly ends abnormally, or if its reported optimum or its final dual bound contradicts the reference optimum by more than the tolerance of the bound validity rule. A feasible solution better than the reference optimum replaces it. On each benchmark instance, every nonempty set of families, GCG as shipped, and SCIP alone run with each seed.

\paragraph{Reference optima.} Gains are reported as shares of the LP gap when MIPLIB-NL gives an optimal value. For four models, our runs contradict this value. For three of them, we use the optimum proved by GCG or SCIP instead, and the fourth is treated as having no reference optimum. Of the 108 models of the census, 75 have a positive gain under some decomposition, and 73 of them are scored. Of the 75, 23 have a certain zero-marginal group at $\emptyset$ and 18 others only a possible one. On 17 of the 23, the group is certain in the evaluations without linking variables, and these are the 17 models with a certain joint group on which predictions are scored (\cref{tab:llm}). On the other six, the group is certain only when a family of variables is declared as linking variables, and it is a possible group otherwise. Of the 98 models with a reference optimum and a positive LP gap, the 69 whose bound improves under some decomposition appear in \cref{fig:results}(a).

\FloatBarrier
\section{Further Results}\label[appendix]{app:results}

\Cref{tab:app-recall} gives recall under every presentation, and \cref{tab:app-paired} the paired differences between presentations. \Cref{tab:app-reason} gives the difference between Grok 4.20 (Reasoning) and Grok 4.20 (Non-Reasoning), \cref{tab:app-paired-search} the paired differences between orders in evaluations, \cref{tab:app-five} the five-dual lookahead order and the orders interleaved with it, and \cref{tab:app-first} the first sets of the language models against those of this order. The later parts of this appendix give the results on the rich models with the coefficient variant and the control presentations, the learned predictor, the benchmark instances and the instances without a joint support, the rich models with 12 to 16 families, the cost of a prediction, and the time of one evaluation. Standard errors and 95 percent percentile intervals come from a two-stage bootstrap with 10{,}000 resamples. It draws groups of instances whose requests are identical in at least one presentation with replacement, and then the instances within each drawn group with replacement: the 15 templates form 11 groups, the 102 scored MIPLIB-NL models 99, the 24 rich models with seven to nine families 16, the 60 with 12 to 16 families 47, the 59 benchmark instances 8, and the 76 instances without a joint support 17. Where every group has one instance, as under the coefficient variant, this is the bootstrap over instances. Every order that depends on the order of the families, including the tie-breaking of the lookahead order, is evaluated under the same 1000 random orders of the families of each instance and averaged. Lookahead values that differ by at most $10^{-9}$ count as ties. 

\begin{table}[!htbp]
\caption{Recall of joint supports by presentation, mean $\pm$ standard error.}
\label{tab:app-recall}
\centering
\small
\begin{tabular}{@{}lcccc@{}}
\toprule
Configuration & Original & Names only & Permuted & Anonymized \\
\midrule
\multicolumn{5}{@{}l}{\emph{Templates}} \\
Claude Sonnet 5.5 & 0.74 $\pm$ 0.07 & 0.78 $\pm$ 0.11 & 0.39 $\pm$ 0.14 & 0.48 $\pm$ 0.14 \\
GPT-6.1 Sol & 0.96 $\pm$ 0.03 & 0.33 $\pm$ 0.18 & 0.00 $\pm$ 0.00 & 0.00 $\pm$ 0.00 \\
Grok 4.20 (Reasoning) & 0.67 $\pm$ 0.08 & 0.81 $\pm$ 0.08 & 0.22 $\pm$ 0.07 & 0.30 $\pm$ 0.13 \\
Grok 4.20 (Non-Reasoning) & 0.17 $\pm$ 0.10 & 0.37 $\pm$ 0.15 & 0.26 $\pm$ 0.08 & 0.26 $\pm$ 0.12 \\
DeepSeek-V4-Pro (Non-Thinking) & 0.26 $\pm$ 0.12 & 0.33 $\pm$ 0.06 & 0.28 $\pm$ 0.06 & 0.30 $\pm$ 0.13 \\
\multicolumn{5}{@{}l}{\emph{MIPLIB-NL}} \\
Claude Sonnet 5.5 & 0.56 $\pm$ 0.10 & 0.50 $\pm$ 0.10 & 0.54 $\pm$ 0.11 & 0.50 $\pm$ 0.08 \\
GPT-6.1 Sol & 0.53 $\pm$ 0.12 & 0.45 $\pm$ 0.12 & 0.39 $\pm$ 0.11 & 0.41 $\pm$ 0.12 \\
Grok 4.20 (Reasoning) & 0.55 $\pm$ 0.11 & 0.43 $\pm$ 0.10 & 0.44 $\pm$ 0.10 & 0.31 $\pm$ 0.08 \\
Grok 4.20 (Non-Reasoning) & 0.29 $\pm$ 0.10 & 0.18 $\pm$ 0.07 & 0.31 $\pm$ 0.09 & 0.10 $\pm$ 0.05 \\
DeepSeek-V4-Pro (Non-Thinking) & 0.37 $\pm$ 0.09 & 0.36 $\pm$ 0.07 & 0.39 $\pm$ 0.10 & 0.44 $\pm$ 0.09 \\
\bottomrule
\end{tabular}
\end{table}

\begin{table}[!htbp]
\caption{Paired differences in recall between presentations, mean $\pm$ standard error. An asterisk marks a difference whose 95 percent bootstrap interval excludes zero.}
\label{tab:app-paired}
\centering
\footnotesize
\begin{tabular}{@{}lccc@{}}
\toprule
Configuration & Names only $-$ permuted & Names only $-$ anonymized & Original $-$ anonymized \\
\midrule
\multicolumn{4}{@{}l}{\emph{Templates}} \\
Claude Sonnet 5.5 & 0.39 $\pm$ 0.18$^{*}$ & 0.30 $\pm$ 0.14$^{*}$ & 0.26 $\pm$ 0.20 \\
GPT-6.1 Sol & 0.33 $\pm$ 0.18$^{*}$ & 0.33 $\pm$ 0.18$^{*}$ & 0.96 $\pm$ 0.03$^{*}$ \\
Grok 4.20 (Reasoning) & 0.59 $\pm$ 0.06$^{*}$ & 0.52 $\pm$ 0.18$^{*}$ & 0.37 $\pm$ 0.15$^{*}$ \\
Grok 4.20 (Non-Reasoning) & 0.11 $\pm$ 0.16 & 0.11 $\pm$ 0.12 & $-$0.09 $\pm$ 0.08 \\
DeepSeek-V4-Pro (Non-Thinking) & 0.06 $\pm$ 0.05 & 0.04 $\pm$ 0.11 & $-$0.04 $\pm$ 0.03 \\
\multicolumn{4}{@{}l}{\emph{MIPLIB-NL}} \\
Claude Sonnet 5.5 & $-$0.04 $\pm$ 0.05 & 0.00 $\pm$ 0.07 & 0.06 $\pm$ 0.07 \\
GPT-6.1 Sol & 0.06 $\pm$ 0.06 & 0.04 $\pm$ 0.06 & 0.12 $\pm$ 0.08 \\
Grok 4.20 (Reasoning) & $-$0.01 $\pm$ 0.04 & 0.12 $\pm$ 0.07 & 0.24 $\pm$ 0.07$^{*}$ \\
Grok 4.20 (Non-Reasoning) & $-$0.14 $\pm$ 0.07$^{*}$ & 0.08 $\pm$ 0.07 & 0.20 $\pm$ 0.09$^{*}$ \\
DeepSeek-V4-Pro (Non-Thinking) & $-$0.03 $\pm$ 0.09 & $-$0.08 $\pm$ 0.07 & $-$0.07 $\pm$ 0.10 \\
\bottomrule
\end{tabular}
\end{table}

\begin{table}[!htbp]
\caption{Grok 4.20 (Reasoning) minus Grok 4.20 (Non-Reasoning), paired over instances, hit@1 and recall, mean $\pm$ standard error. An asterisk marks a difference whose 95 percent bootstrap interval excludes zero.}
\label{tab:app-reason}
\centering
\small
\begin{tabular}{@{}lcc@{}}
\toprule
Presentation & hit@1 & recall \\
\midrule
\multicolumn{3}{@{}l}{\emph{Templates}} \\
Original & 0.33 $\pm$ 0.13$^{*}$ & 0.50 $\pm$ 0.12$^{*}$ \\
Names only & 0.37 $\pm$ 0.24 & 0.44 $\pm$ 0.23 \\
Permuted & $-$0.15 $\pm$ 0.15 & $-$0.04 $\pm$ 0.07 \\
Anonymized & $-$0.11 $\pm$ 0.11 & 0.04 $\pm$ 0.11 \\
\multicolumn{3}{@{}l}{\emph{MIPLIB-NL}} \\
Original & 0.25 $\pm$ 0.11$^{*}$ & 0.25 $\pm$ 0.11$^{*}$ \\
Names only & 0.27 $\pm$ 0.11$^{*}$ & 0.25 $\pm$ 0.10$^{*}$ \\
Permuted & 0.10 $\pm$ 0.13 & 0.13 $\pm$ 0.12 \\
Anonymized & 0.22 $\pm$ 0.11$^{*}$ & 0.22 $\pm$ 0.10$^{*}$ \\
\bottomrule
\end{tabular}
\end{table}

\paragraph{Five optimal duals.} \Cref{tab:app-five} gives the first set, and the search cost of the five-dual lookahead order and of the JointDW orders interleaved with it. On the templates, its first set is a true joint support on every instance, against two-thirds for the HiGHS dual, and it needs one evaluation on every template. On MIPLIB-NL, its first set is unchanged, and its order needs $-1.00\pm0.75$ evaluations relative to the HiGHS dual for $s=2$, within sampling variation. On the rich models, it needs $1.73\pm0.77$ evaluations for $s=2$ against $3.87\pm1.48$ at the HiGHS dual, a difference within sampling variation, and its first set, scored as a joint support, is less often exactly the true group. Against the five-dual order, JointDW with GPT-6.1 Sol does not differ beyond sampling variation on MIPLIB-NL at any $s$, while JointDW with each other configuration needs more evaluations at every $s$. Interleaving with the five-dual order costs every configuration at most about half an evaluation more than the five-dual order on MIPLIB-NL, and on the templates it costs the four configurations other than GPT-6.1 Sol between $0.4$ and $0.9$ evaluations more. On the rich models, interleaving with the five-dual order lowers the cost of interleaving with the HiGHS dual by about 40 percent for $s=2$.

\begin{table}[!htbp]
\caption{Five-dual lookahead. Top: hit@1 and recall of the first set of the lookahead order at the HiGHS dual and with five optimal duals, mean $\pm$ standard error, on the nine structured templates, the 17 MIPLIB-NL models with a certain joint group, and the 24 rich models. Bottom: number of evaluations until a best set is evaluated, by size limit $s$, for the two lookahead orders and for JointDW interleaved with the five-dual order, on the 13 templates with a positive gain for some set, the 17 MIPLIB-NL models, and the 24 rich models; JointDW uses the original presentation, which shows no coefficients. One of these templates has no positive gain for $s=2$, where every order costs one evaluation on it.}
\label{tab:app-five}
\centering
\footnotesize
\setlength{\tabcolsep}{3pt}
\begin{tabular}{@{}lcccccc@{}}
\toprule
& \multicolumn{2}{c}{Templates} & \multicolumn{2}{c}{MIPLIB-NL} & \multicolumn{2}{c}{Rich models} \\
\cmidrule(lr){2-3}\cmidrule(lr){4-5}\cmidrule(l){6-7}
First set & hit@1 & recall & hit@1 & recall & hit@1 & recall \\
\midrule
Lookahead (HiGHS dual) & 0.67 $\pm$ 0.24 & 0.44 $\pm$ 0.17 & 0.53 $\pm$ 0.12 & 0.44 $\pm$ 0.11 & 0.21 $\pm$ 0.10 & 0.19 $\pm$ 0.10 \\
Lookahead (five duals) & 1.00 $\pm$ 0.00 & 0.78 $\pm$ 0.11 & 0.53 $\pm$ 0.12 & 0.44 $\pm$ 0.11 & 0.04 $\pm$ 0.05 & 0.04 $\pm$ 0.05 \\
\bottomrule
\end{tabular}

\medskip
\setlength{\tabcolsep}{2.2pt}
\scriptsize
\resizebox{\textwidth}{!}{%
\begin{tabular}{@{}lcccccccc@{}}
\toprule
& \multicolumn{3}{c}{Templates} & \multicolumn{3}{c}{MIPLIB-NL} & \multicolumn{2}{c}{Rich models} \\
\cmidrule(lr){2-4}\cmidrule(lr){5-7}\cmidrule(l){8-9}
Order & $s=2$ & $3$ & $k-1$ & $2$ & $3$ & $k-1$ & $2$ & $3$ \\
\midrule
Lookahead (HiGHS dual) & 1.92 $\pm$ 0.71 & 1.92 $\pm$ 0.71 & 1.92 $\pm$ 0.71 & 2.18 $\pm$ 0.78 & 2.12 $\pm$ 0.78 & 2.29 $\pm$ 0.94 & 3.87 $\pm$ 1.48 & 1.00 $\pm$ 0.00 \\
Lookahead (five duals) & 1.00 $\pm$ 0.00 & 1.00 $\pm$ 0.00 & 1.00 $\pm$ 0.00 & 1.18 $\pm$ 0.13 & 1.24 $\pm$ 0.16 & 1.65 $\pm$ 0.52 & 1.73 $\pm$ 0.77 & 1.00 $\pm$ 0.00 \\
\multicolumn{9}{@{}l}{\emph{JointDW interleaved with the five-dual order}} \\
Claude Sonnet 5.5 & 1.41 $\pm$ 0.14 & 1.45 $\pm$ 0.14 & 1.50 $\pm$ 0.17 & 1.56 $\pm$ 0.14 & 1.63 $\pm$ 0.13 & 1.72 $\pm$ 0.21 & 2.15 $\pm$ 0.54 & 1.65 $\pm$ 0.07 \\
GPT-6.1 Sol & 1.19 $\pm$ 0.11 & 1.18 $\pm$ 0.11 & 1.10 $\pm$ 0.08 & 1.34 $\pm$ 0.15 & 1.33 $\pm$ 0.10 & 1.33 $\pm$ 0.10 & 2.43 $\pm$ 0.76 & 1.73 $\pm$ 0.06 \\
Grok 4.20 (Reasoning) & 1.40 $\pm$ 0.12 & 1.42 $\pm$ 0.13 & 1.43 $\pm$ 0.14 & 1.51 $\pm$ 0.15 & 1.60 $\pm$ 0.17 & 1.66 $\pm$ 0.22 & 2.11 $\pm$ 0.45 & 1.48 $\pm$ 0.08 \\
Grok 4.20 (Non-Reasoning) & 1.68 $\pm$ 0.16 & 1.74 $\pm$ 0.15 & 1.84 $\pm$ 0.18 & 1.61 $\pm$ 0.16 & 1.69 $\pm$ 0.17 & 1.96 $\pm$ 0.41 & 2.57 $\pm$ 0.60 & 2.00 $\pm$ 0.00 \\
DeepSeek-V4-Pro (Non-Thinking) & 1.81 $\pm$ 0.11 & 1.84 $\pm$ 0.09 & 1.89 $\pm$ 0.13 & 1.66 $\pm$ 0.13 & 1.74 $\pm$ 0.11 & 2.02 $\pm$ 0.37 & 2.58 $\pm$ 0.62 & 1.99 $\pm$ 0.01 \\
\bottomrule
\end{tabular}}
\end{table}

\paragraph{Rich models and control presentations.} \Cref{tab:rich-pred,tab:rich-search} give the predictions and search costs on the rich models with and without coefficients. With every coefficient, JointDW with GPT-6.1 Sol minus the lookahead order at the HiGHS dual is $-2.70\pm1.33$ evaluations for $s=2$ and $1.10\pm0.46$ for $s=3$, both beyond sampling variation. Against the five-dual lookahead order, it is $-0.56\pm0.74$ for $s=2$ and $1.10\pm0.46$ for $s=3$, and the interleaved order is $-0.71\pm0.71$ and $0.18\pm0.08$; JointDW with Claude Sonnet 5.5 needs $3.96\pm0.95$ evaluations more than the five-dual order for $s=2$, and its interleaving $0.18\pm0.32$ more, and the three other configurations need between 7.4 and 10.3 more, which interleaving reduces to less than one (\cref{tab:app-rich-five}). Without coefficients, GPT-6.1 Sol returns no joint support in 49 of its 72 answers, and with them in one. \Cref{tab:app-control} gives the two control presentations with coefficients. For GPT-6.1 Sol, permuting the names changes hit@1 by $0.00\pm0.02$ and lowers recall by $0.07\pm0.03$, beyond sampling variation, and the search cost by less than sampling variation; replacing the names by labels changes neither beyond sampling variation. For Claude Sonnet 5.5, permuting the names lowers hit@1 by $0.36\pm0.08$ and raises the cost of JointDW by $5.06\pm1.16$ evaluations for $s=2$, both beyond sampling variation, and the labels lower hit@1 by $0.19\pm0.09$.

\begin{table}[!htbp]
\caption{Rich models: hit@1 and recall given the text and given every coefficient, mean $\pm$ standard error over 24 instances.}
\label{tab:rich-pred}
\centering
\footnotesize
\setlength{\tabcolsep}{3pt}
\begin{tabular}{@{}lcccc@{}}
\toprule
& \multicolumn{2}{c}{hit@1} & \multicolumn{2}{c}{recall} \\
\cmidrule(lr){2-3}\cmidrule(l){4-5}
Configuration & text & coefficients & text & coefficients \\
\midrule
Claude Sonnet 5.5 & 0.32 $\pm$ 0.06 & 0.47 $\pm$ 0.08 & 0.44 $\pm$ 0.11 & 0.50 $\pm$ 0.06 \\
GPT-6.1 Sol & 0.22 $\pm$ 0.06 & 0.99 $\pm$ 0.01 & 0.24 $\pm$ 0.08 & 0.91 $\pm$ 0.03 \\
Grok 4.20 (Reasoning) & 0.17 $\pm$ 0.07 & 0.11 $\pm$ 0.04 & 0.17 $\pm$ 0.07 & 0.08 $\pm$ 0.03 \\
Grok 4.20 (Non-Reasoning) & 0.00 $\pm$ 0.00 & 0.00 $\pm$ 0.00 & 0.00 $\pm$ 0.00 & 0.00 $\pm$ 0.00 \\
DeepSeek-V4-Pro (Non-Thinking) & 0.00 $\pm$ 0.00 & 0.00 $\pm$ 0.00 & 0.02 $\pm$ 0.02 & 0.01 $\pm$ 0.01 \\
\bottomrule
\end{tabular}
\end{table}

\begin{table}[!htbp]
\caption{Rich models: number of evaluations until a best set is evaluated, mean $\pm$ standard error over 24 instances. The GCG anchor exists on the lot-sizing and facility-location models.}
\label{tab:rich-search}
\centering
\footnotesize
\setlength{\tabcolsep}{2.5pt}
\begin{tabular}{@{}lcccc@{}}
\toprule
& \multicolumn{2}{c}{$s=2$} & \multicolumn{2}{c}{$s=3$} \\
\cmidrule(lr){2-3}\cmidrule(l){4-5}
Order & text & coefficients & text & coefficients \\
\midrule
Random order & \multicolumn{2}{c}{17.33 $\pm$ 1.31} & \multicolumn{2}{c}{17.53 $\pm$ 4.83} \\
Covering order & \multicolumn{2}{c}{13.56 $\pm$ 1.19} & \multicolumn{2}{c}{10.45 $\pm$ 3.30} \\
Lookahead order & \multicolumn{2}{c}{3.87 $\pm$ 1.48} & \multicolumn{2}{c}{1.00 $\pm$ 0.00} \\
GCG anchor (16 instances) & \multicolumn{2}{c}{15.23 $\pm$ 1.72} & \multicolumn{2}{c}{15.80 $\pm$ 4.70} \\
\multicolumn{5}{@{}l}{\emph{JointDW}} \\
Claude Sonnet 5.5 & 8.12 $\pm$ 1.01 & 5.69 $\pm$ 0.69 & 7.47 $\pm$ 1.98 & 4.53 $\pm$ 0.52 \\
GPT-6.1 Sol & 9.21 $\pm$ 0.60 & 1.17 $\pm$ 0.16 & 6.92 $\pm$ 1.07 & 2.10 $\pm$ 0.46 \\
Grok 4.20 (Reasoning) & 8.75 $\pm$ 0.87 & 9.17 $\pm$ 0.46 & 6.34 $\pm$ 1.81 & 6.16 $\pm$ 1.12 \\
Grok 4.20 (Non-Reasoning) & 10.68 $\pm$ 0.64 & 10.27 $\pm$ 0.40 & 12.56 $\pm$ 3.91 & 12.19 $\pm$ 2.87 \\
DeepSeek-V4-Pro (Non-Thinking) & 11.93 $\pm$ 0.93 & 12.00 $\pm$ 0.82 & 10.93 $\pm$ 2.99 & 11.38 $\pm$ 2.51 \\
\multicolumn{5}{@{}l}{\emph{Interleaved with lookahead}} \\
Claude Sonnet 5.5 & 3.46 $\pm$ 0.96 & 3.39 $\pm$ 0.90 & 1.65 $\pm$ 0.07 & 1.63 $\pm$ 0.08 \\
GPT-6.1 Sol & 3.90 $\pm$ 1.17 & 1.01 $\pm$ 0.01 & 1.73 $\pm$ 0.06 & 1.18 $\pm$ 0.08 \\
Grok 4.20 (Reasoning) & 3.53 $\pm$ 0.96 & 3.36 $\pm$ 0.78 & 1.48 $\pm$ 0.08 & 1.58 $\pm$ 0.06 \\
Grok 4.20 (Non-Reasoning) & 4.40 $\pm$ 1.26 & 4.30 $\pm$ 1.05 & 2.00 $\pm$ 0.00 & 2.00 $\pm$ 0.00 \\
DeepSeek-V4-Pro (Non-Thinking) & 4.27 $\pm$ 1.17 & 4.29 $\pm$ 1.05 & 1.99 $\pm$ 0.01 & 1.94 $\pm$ 0.02 \\
\bottomrule
\end{tabular}
\end{table}

\begin{table}[!htbp]
\caption{Control presentations with every coefficient on the 24 rich models. Top: hit@1 and recall, mean $\pm$ standard error, when the family names are shown as given, permuted among the families, or replaced by labels, pooled and by class. Bottom: number of evaluations until a best set is evaluated for JointDW and for its interleaving with the lookahead order at the HiGHS dual, pooled over the 24 instances.}
\label{tab:app-control}
\centering
\footnotesize
\setlength{\tabcolsep}{3pt}
\begin{tabular}{@{}lcccccc@{}}
\toprule
& \multicolumn{2}{c}{names as given} & \multicolumn{2}{c}{names permuted} & \multicolumn{2}{c}{names anonymized} \\
\cmidrule(lr){2-3}\cmidrule(lr){4-5}\cmidrule(l){6-7}
Configuration & hit@1 & recall & hit@1 & recall & hit@1 & recall \\
\midrule
\multicolumn{7}{@{}l}{\emph{Pooled}} \\
GPT-6.1 Sol & 0.99 $\pm$ 0.01 & 0.91 $\pm$ 0.03 & 0.99 $\pm$ 0.01 & 0.83 $\pm$ 0.04 & 1.00 $\pm$ 0.00 & 0.86 $\pm$ 0.04 \\
Claude Sonnet 5.5 & 0.47 $\pm$ 0.08 & 0.50 $\pm$ 0.06 & 0.11 $\pm$ 0.04 & 0.15 $\pm$ 0.05 & 0.28 $\pm$ 0.06 & 0.40 $\pm$ 0.05 \\
\multicolumn{7}{@{}l}{\emph{Lot sizing}} \\
GPT-6.1 Sol & 1.00 $\pm$ 0.00 & 0.84 $\pm$ 0.06 & 1.00 $\pm$ 0.00 & 0.71 $\pm$ 0.07 & 1.00 $\pm$ 0.00 & 0.75 $\pm$ 0.07 \\
Claude Sonnet 5.5 & 0.83 $\pm$ 0.06 & 0.33 $\pm$ 0.06 & 0.21 $\pm$ 0.08 & 0.11 $\pm$ 0.04 & 0.42 $\pm$ 0.11 & 0.21 $\pm$ 0.06 \\
\multicolumn{7}{@{}l}{\emph{Network design}} \\
GPT-6.1 Sol & 0.96 $\pm$ 0.04 & 0.96 $\pm$ 0.04 & 0.96 $\pm$ 0.04 & 0.96 $\pm$ 0.04 & 1.00 $\pm$ 0.00 & 1.00 $\pm$ 0.00 \\
Claude Sonnet 5.5 & 0.17 $\pm$ 0.08 & 0.71 $\pm$ 0.07 & 0.04 $\pm$ 0.04 & 0.21 $\pm$ 0.06 & 0.00 $\pm$ 0.00 & 0.50 $\pm$ 0.10 \\
\multicolumn{7}{@{}l}{\emph{Facility location}} \\
GPT-6.1 Sol & 1.00 $\pm$ 0.00 & 0.92 $\pm$ 0.04 & 1.00 $\pm$ 0.00 & 0.83 $\pm$ 0.08 & 1.00 $\pm$ 0.00 & 0.83 $\pm$ 0.08 \\
Claude Sonnet 5.5 & 0.42 $\pm$ 0.13 & 0.48 $\pm$ 0.10 & 0.08 $\pm$ 0.05 & 0.15 $\pm$ 0.11 & 0.42 $\pm$ 0.08 & 0.48 $\pm$ 0.07 \\
\bottomrule
\end{tabular}

\medskip
\begin{tabular}{@{}lccc@{}}
\toprule
Order & names as given & names permuted & names anonymized \\
\midrule
\multicolumn{4}{@{}l}{\emph{$s=2$}} \\
JointDW with GPT-6.1 Sol & 1.17 $\pm$ 0.16 & 1.27 $\pm$ 0.18 & 1.00 $\pm$ 0.00 \\
Interleaved order with GPT-6.1 Sol & 1.01 $\pm$ 0.01 & 1.03 $\pm$ 0.02 & 1.00 $\pm$ 0.00 \\
JointDW with Claude Sonnet 5.5 & 5.69 $\pm$ 0.70 & 10.75 $\pm$ 0.79 & 8.45 $\pm$ 0.63 \\
Interleaved order with Claude Sonnet 5.5 & 3.39 $\pm$ 0.90 & 4.29 $\pm$ 1.12 & 3.25 $\pm$ 0.82 \\
\multicolumn{4}{@{}l}{\emph{$s=3$}} \\
JointDW with GPT-6.1 Sol & 2.10 $\pm$ 0.46 & 2.14 $\pm$ 0.46 & 1.99 $\pm$ 0.46 \\
Interleaved order with GPT-6.1 Sol & 1.18 $\pm$ 0.08 & 1.19 $\pm$ 0.08 & 1.17 $\pm$ 0.08 \\
JointDW with Claude Sonnet 5.5 & 4.53 $\pm$ 0.51 & 10.18 $\pm$ 2.04 & 7.56 $\pm$ 1.31 \\
Interleaved order with Claude Sonnet 5.5 & 1.62 $\pm$ 0.08 & 1.86 $\pm$ 0.05 & 1.72 $\pm$ 0.06 \\
\bottomrule
\end{tabular}
\end{table}

\begin{table}[!htbp]
\caption{Rich models with every coefficient: paired differences in the number of evaluations until a best set is evaluated, JointDW and its interleaving with the five-dual lookahead order minus the five-dual lookahead order, for $s=2$ and $s=3$, pooled over the 24 instances and by class, with every coefficient shown, with the family names as given and, for two configurations, permuted or anonymized. Mean $\pm$ standard error of the cluster bootstrap over prompt groups, with an asterisk when the 95 percent interval excludes zero.}
\label{tab:app-rich-five}
\centering
\footnotesize
\setlength{\tabcolsep}{4pt}
\begin{tabular}{@{}llcccc@{}}
\toprule
& & \multicolumn{2}{c}{JointDW minus five-dual} & \multicolumn{2}{c}{Interleaved minus five-dual} \\
\cmidrule(lr){3-4}\cmidrule(l){5-6}
Configuration & names & $s=2$ & $s=3$ & $s=2$ & $s=3$ \\
\midrule
\multicolumn{6}{@{}l}{\emph{Pooled}} \\
Claude Sonnet 5.5 & as given & 3.96 $\pm$ 0.95$^{*}$ & 3.53 $\pm$ 0.51$^{*}$ & 0.18 $\pm$ 0.32 & 0.62 $\pm$ 0.08$^{*}$ \\
 & permuted & 9.02 $\pm$ 1.06$^{*}$ & 9.18 $\pm$ 2.04$^{*}$ & 0.76 $\pm$ 0.10$^{*}$ & 0.86 $\pm$ 0.05$^{*}$ \\
 & anonymized & 6.72 $\pm$ 1.10$^{*}$ & 6.56 $\pm$ 1.31$^{*}$ & 0.17 $\pm$ 0.54 & 0.72 $\pm$ 0.06$^{*}$ \\
GPT-6.1 Sol & as given & $-$0.56 $\pm$ 0.74 & 1.10 $\pm$ 0.46$^{*}$ & $-$0.71 $\pm$ 0.71 & 0.18 $\pm$ 0.08$^{*}$ \\
 & permuted & $-$0.46 $\pm$ 0.75 & 1.14 $\pm$ 0.46$^{*}$ & $-$0.70 $\pm$ 0.71 & 0.19 $\pm$ 0.08$^{*}$ \\
 & anonymized & $-$0.73 $\pm$ 0.71 & 0.99 $\pm$ 0.46$^{*}$ & $-$0.73 $\pm$ 0.71 & 0.17 $\pm$ 0.08$^{*}$ \\
Grok 4.20 (Reasoning) & as given & 7.44 $\pm$ 0.91$^{*}$ & 5.16 $\pm$ 1.12$^{*}$ & 0.51 $\pm$ 0.24 & 0.58 $\pm$ 0.06$^{*}$ \\
Grok 4.20 (Non-Reasoning) & as given & 8.55 $\pm$ 0.87$^{*}$ & 11.19 $\pm$ 2.87$^{*}$ & 0.84 $\pm$ 0.16$^{*}$ & 1.00 $\pm$ 0.00$^{*}$ \\
DeepSeek-V4-Pro (Non-Thinking) & as given & 10.27 $\pm$ 1.15$^{*}$ & 10.38 $\pm$ 2.52$^{*}$ & 0.84 $\pm$ 0.13$^{*}$ & 0.94 $\pm$ 0.02$^{*}$ \\
\multicolumn{6}{@{}l}{\emph{Lot sizing}} \\
Claude Sonnet 5.5 & as given & 1.43 $\pm$ 0.51$^{*}$ & 3.39 $\pm$ 1.32$^{*}$ & 0.11 $\pm$ 0.04$^{*}$ & 0.46 $\pm$ 0.17$^{*}$ \\
 & permuted & 11.53 $\pm$ 2.09$^{*}$ & 17.52 $\pm$ 4.95$^{*}$ & 0.73 $\pm$ 0.09$^{*}$ & 0.79 $\pm$ 0.13$^{*}$ \\
 & anonymized & 7.79 $\pm$ 1.34$^{*}$ & 11.36 $\pm$ 3.28$^{*}$ & 0.54 $\pm$ 0.10$^{*}$ & 0.62 $\pm$ 0.12$^{*}$ \\
GPT-6.1 Sol & as given & 0.00 $\pm$ 0.00 & 2.97 $\pm$ 1.09$^{*}$ & 0.00 $\pm$ 0.00 & 0.50 $\pm$ 0.18$^{*}$ \\
 & permuted & 0.47 $\pm$ 0.44 & 3.23 $\pm$ 1.03$^{*}$ & 0.04 $\pm$ 0.04 & 0.54 $\pm$ 0.17$^{*}$ \\
 & anonymized & 0.00 $\pm$ 0.00 & 2.97 $\pm$ 1.09$^{*}$ & 0.00 $\pm$ 0.00 & 0.50 $\pm$ 0.18$^{*}$ \\
Grok 4.20 (Reasoning) & as given & 9.38 $\pm$ 0.82$^{*}$ & 9.12 $\pm$ 2.80$^{*}$ & 0.67 $\pm$ 0.05$^{*}$ & 0.62 $\pm$ 0.14$^{*}$ \\
Grok 4.20 (Non-Reasoning) & as given & 11.85 $\pm$ 0.27$^{*}$ & 24.32 $\pm$ 6.50$^{*}$ & 1.00 $\pm$ 0.00$^{*}$ & 1.00 $\pm$ 0.00$^{*}$ \\
DeepSeek-V4-Pro (Non-Thinking) & as given & 16.47 $\pm$ 0.46$^{*}$ & 22.79 $\pm$ 5.30$^{*}$ & 1.00 $\pm$ 0.00$^{*}$ & 1.00 $\pm$ 0.00$^{*}$ \\
\multicolumn{6}{@{}l}{\emph{Network design}} \\
Claude Sonnet 5.5 & as given & 6.81 $\pm$ 0.69$^{*}$ & 3.96 $\pm$ 0.41$^{*}$ & 0.83 $\pm$ 0.08$^{*}$ & 0.83 $\pm$ 0.08$^{*}$ \\
 & permuted & 9.58 $\pm$ 0.48$^{*}$ & 5.68 $\pm$ 0.28$^{*}$ & 0.96 $\pm$ 0.04$^{*}$ & 0.96 $\pm$ 0.04$^{*}$ \\
 & anonymized & 9.36 $\pm$ 0.50$^{*}$ & 5.39 $\pm$ 0.35$^{*}$ & 0.99 $\pm$ 0.01$^{*}$ & 0.96 $\pm$ 0.04$^{*}$ \\
GPT-6.1 Sol & as given & 0.50 $\pm$ 0.47 & 0.32 $\pm$ 0.30 & 0.04 $\pm$ 0.04 & 0.04 $\pm$ 0.04 \\
 & permuted & 0.33 $\pm$ 0.30 & 0.18 $\pm$ 0.17 & 0.04 $\pm$ 0.04 & 0.04 $\pm$ 0.04 \\
 & anonymized & 0.00 $\pm$ 0.00 & 0.00 $\pm$ 0.00 & 0.00 $\pm$ 0.00 & 0.00 $\pm$ 0.00 \\
Grok 4.20 (Reasoning) & as given & 6.95 $\pm$ 0.43$^{*}$ & 2.64 $\pm$ 0.40$^{*}$ & 0.77 $\pm$ 0.05$^{*}$ & 0.46 $\pm$ 0.08$^{*}$ \\
Grok 4.20 (Non-Reasoning) & as given & 8.32 $\pm$ 0.31$^{*}$ & 4.85 $\pm$ 0.22$^{*}$ & 1.00 $\pm$ 0.00$^{*}$ & 1.00 $\pm$ 0.00$^{*}$ \\
DeepSeek-V4-Pro (Non-Thinking) & as given & 7.98 $\pm$ 0.33$^{*}$ & 3.95 $\pm$ 0.31$^{*}$ & 0.95 $\pm$ 0.02$^{*}$ & 0.86 $\pm$ 0.05$^{*}$ \\
\multicolumn{6}{@{}l}{\emph{Facility location}} \\
Claude Sonnet 5.5 & as given & 3.65 $\pm$ 2.38 & 3.25 $\pm$ 0.69$^{*}$ & $-$0.40 $\pm$ 0.92 & 0.58 $\pm$ 0.13$^{*}$ \\
 & permuted & 5.97 $\pm$ 1.89$^{*}$ & 4.34 $\pm$ 0.33$^{*}$ & 0.60 $\pm$ 0.27$^{*}$ & 0.83 $\pm$ 0.05$^{*}$ \\
 & anonymized & 3.01 $\pm$ 2.48 & 2.92 $\pm$ 0.36$^{*}$ & $-$1.02 $\pm$ 1.56 & 0.58 $\pm$ 0.08$^{*}$ \\
GPT-6.1 Sol & as given & $-$2.18 $\pm$ 2.07 & 0.00 $\pm$ 0.00 & $-$2.18 $\pm$ 2.07 & 0.00 $\pm$ 0.00 \\
 & permuted & $-$2.18 $\pm$ 2.07 & 0.00 $\pm$ 0.00 & $-$2.18 $\pm$ 2.07 & 0.00 $\pm$ 0.00 \\
 & anonymized & $-$2.18 $\pm$ 2.07 & 0.00 $\pm$ 0.00 & $-$2.18 $\pm$ 2.07 & 0.00 $\pm$ 0.00 \\
Grok 4.20 (Reasoning) & as given & 5.99 $\pm$ 2.42$^{*}$ & 3.72 $\pm$ 0.57$^{*}$ & 0.10 $\pm$ 0.71 & 0.67 $\pm$ 0.09$^{*}$ \\
Grok 4.20 (Non-Reasoning) & as given & 5.48 $\pm$ 2.05$^{*}$ & 4.40 $\pm$ 0.03$^{*}$ & 0.52 $\pm$ 0.46 & 1.00 $\pm$ 0.00$^{*}$ \\
DeepSeek-V4-Pro (Non-Thinking) & as given & 6.37 $\pm$ 2.08$^{*}$ & 4.39 $\pm$ 0.09$^{*}$ & 0.58 $\pm$ 0.39 & 0.97 $\pm$ 0.01$^{*}$ \\
\bottomrule
\end{tabular}
\end{table}

\paragraph{Learned predictor.} The learned predictor is a gradient-boosted regression \citep{chen2016xgboost} of the gain share of a set of families, the gain divided by the best gain of the instance, from static features of the decomposition that convexifies the set in the manner of \citet{basso2023data}: numbers and sizes of the blocks, their densities, the size of the border, statistics of the right-hand sides and the objective coefficients, the share of integer variables, the number of variables shared by the families of the set, and, in the variant with lookahead, the lookahead value of the set at the HiGHS dual. The hyperparameters are those of \citet{basso2023data}: depth 13, learning rate 0.1, and 100 rounds. The sets of at most $s$ families are ordered by decreasing predicted share, and the order is scored like the lookahead order: as an answer, as a search order, as the anchor of the staged search, and interleaved with the five-dual lookahead order. The labeling cost of a split is the number of evaluations in the gain tables of its training instances, which is what the labels cost. Three splits are reported in \cref{tab:app-learned}: within a class, trained on the other instances of the class of the test instance; leave one class out, trained on the templates and rich models of the other two classes and on all scored MIPLIB-NL models; and MIPLIB-NL leave one out, trained on the other scored MIPLIB-NL models. \Cref{tab:app-learned-size} gives the cost against the number of training instances of the same class, with 20 random draws of the training set for each size. Against JointDW with GPT-6.1 Sol given the coefficients, the structural order within a class differs by $-0.06\pm0.19$ evaluations for $s=2$ and $-0.20\pm0.21$ for $s=3$, and across classes by $4.81\pm1.43$ and $1.09\pm1.68$; with the lookahead feature, the differences across classes are $0.17\pm0.32$ and $-0.68\pm0.38$. On MIPLIB-NL, where the structural order is scored on the 88 and 89 scored models with a best set of at most $s$ families, it differs from the five-dual lookahead order by $0.60\pm0.26$ evaluations for $s=2$ and by $0.05\pm0.12$ with the lookahead feature; against JointDW with GPT-6.1 Sol, on the 84 and 82 models where both are scored, the structural order differs by $-0.10\pm0.22$ for $s=2$ and $-0.70\pm0.37$ for $s=3$, and the order with the lookahead feature by $-0.70\pm0.18$, beyond sampling variation, and $-0.41\pm0.41$. The five features of largest gain importance are listed with the released data.

\begin{table}[!htbp]
\caption{Learned predictor: hit@1 of its first set and of the five-dual lookahead order, number of evaluations until a best set is evaluated for the learned order and for the staged search anchored at its first set, mean $\pm$ standard error, the mean labeling cost of the training set in evaluations, and the number of test instances, by split and variant, for $s=2$ and $s=3$. Templates: the instances with a positive gain at $s$ for the within-class split, and the nine structured templates for the leave-one-class-out split.}
\label{tab:app-learned}
\centering
\footnotesize
\setlength{\tabcolsep}{3pt}
\begin{tabular}{@{}lccccccc@{}}
\toprule
& & \multicolumn{2}{c}{hit@1} & \multicolumn{2}{c}{evaluations} & & \\
\cmidrule(lr){3-4}\cmidrule(lr){5-6}
Instances & $s$ & learned & five duals & learned order & staged & labeling cost & $n$ \\
\midrule
\multicolumn{8}{@{}l}{\emph{Within class, structural features}} \\
Templates & $2$ & 0.81 $\pm$ 0.11 & 0.92 $\pm$ 0.08 & 1.25 $\pm$ 0.20 & 1.21 $\pm$ 0.15 & 19.7 & 14 \\
Templates & $3$ & 0.86 $\pm$ 0.10 & 1.00 $\pm$ 0.00 & 1.33 $\pm$ 0.28 & 1.19 $\pm$ 0.14 & 28.5 & 15 \\
Rich models & $2$ & 0.96 $\pm$ 0.03 & 0.96 $\pm$ 0.04 & 1.11 $\pm$ 0.05 & 2.38 $\pm$ 0.74 & 235.7 & 24 \\
Rich models & $3$ & 0.42 $\pm$ 0.13 & 0.04 $\pm$ 0.05 & 1.90 $\pm$ 0.59 & 3.23 $\pm$ 1.36 & 595.0 & 24 \\
\multicolumn{8}{@{}l}{\emph{Within class, structural and lookahead features}} \\
Templates & $2$ & 0.81 $\pm$ 0.11 & 0.92 $\pm$ 0.08 & 1.25 $\pm$ 0.20 & 1.21 $\pm$ 0.15 & 19.7 & 14 \\
Templates & $3$ & 0.86 $\pm$ 0.10 & 1.00 $\pm$ 0.00 & 1.33 $\pm$ 0.28 & 1.19 $\pm$ 0.14 & 28.5 & 15 \\
Rich models & $2$ & 0.96 $\pm$ 0.04 & 0.96 $\pm$ 0.04 & 1.04 $\pm$ 0.04 & 1.32 $\pm$ 0.34 & 235.7 & 24 \\
Rich models & $3$ & 0.33 $\pm$ 0.12 & 0.04 $\pm$ 0.05 & 1.04 $\pm$ 0.05 & 1.65 $\pm$ 0.76 & 595.0 & 24 \\
\multicolumn{8}{@{}l}{\emph{Leave one class out, structural features}} \\
Templates & $2$ & 0.44 $\pm$ 0.23 & 1.00 $\pm$ 0.00 & 2.78 $\pm$ 0.79 & 1.94 $\pm$ 0.30 & 1393.7 & 9 \\
Templates & $3$ & 0.89 $\pm$ 0.12 & 1.00 $\pm$ 0.00 & 1.22 $\pm$ 0.25 & 1.17 $\pm$ 0.18 & 2607.0 & 9 \\
Rich models & $2$ & 0.33 $\pm$ 0.14 & 0.96 $\pm$ 0.04 & 5.98 $\pm$ 1.38 & 9.63 $\pm$ 2.32 & 1393.7 & 24 \\
Rich models & $3$ & 0.04 $\pm$ 0.04 & 0.04 $\pm$ 0.05 & 3.19 $\pm$ 1.98 & 4.79 $\pm$ 1.39 & 2607.0 & 24 \\
\multicolumn{8}{@{}l}{\emph{Leave one class out, structural and lookahead features}} \\
Templates & $2$ & 0.67 $\pm$ 0.24 & 1.00 $\pm$ 0.00 & 2.33 $\pm$ 0.95 & 1.50 $\pm$ 0.36 & 1393.7 & 9 \\
Templates & $3$ & 0.44 $\pm$ 0.22 & 1.00 $\pm$ 0.00 & 2.78 $\pm$ 0.85 & 1.83 $\pm$ 0.33 & 2607.0 & 9 \\
Rich models & $2$ & 0.96 $\pm$ 0.04 & 0.96 $\pm$ 0.04 & 1.33 $\pm$ 0.26 & 2.25 $\pm$ 0.89 & 1393.7 & 24 \\
Rich models & $3$ & 0.00 $\pm$ 0.00 & 0.04 $\pm$ 0.05 & 1.42 $\pm$ 0.29 & 3.67 $\pm$ 1.68 & 2607.0 & 24 \\
\multicolumn{8}{@{}l}{\emph{MIPLIB-NL, leave one out, structural features}} \\
MIPLIB-NL & $2$ & 0.56 $\pm$ 0.12 & 0.65 $\pm$ 0.12 & 1.81 $\pm$ 0.25 & 1.93 $\pm$ 0.21 & 811.0 & 88 \\
MIPLIB-NL & $3$ & 0.47 $\pm$ 0.12 & 0.59 $\pm$ 0.12 & 1.90 $\pm$ 0.42 & 1.64 $\pm$ 0.19 & 1199.1 & 89 \\
\multicolumn{8}{@{}l}{\emph{MIPLIB-NL, leave one out, structural and lookahead features}} \\
MIPLIB-NL & $2$ & 0.65 $\pm$ 0.12 & 0.65 $\pm$ 0.12 & 1.26 $\pm$ 0.11 & 1.30 $\pm$ 0.14 & 811.0 & 88 \\
MIPLIB-NL & $3$ & 0.47 $\pm$ 0.12 & 0.59 $\pm$ 0.12 & 2.04 $\pm$ 0.51 & 1.70 $\pm$ 0.24 & 1199.1 & 89 \\
\bottomrule
\end{tabular}
\end{table}

\begin{table}[!htbp]
\caption{Learned predictor against the number of training instances of the same class on the rich models: number of evaluations until a best set is evaluated for the learned order, mean $\pm$ standard error over 20 random draws of the training set, with the labeling cost of each training set in evaluations.}
\label{tab:app-learned-size}
\centering
\footnotesize
\setlength{\tabcolsep}{3pt}
\begin{tabular}{@{}llcccccc@{}}
\toprule
& & \multicolumn{4}{c}{training instances} & \\
\cmidrule(lr){3-6}
Class & features, $s$ & 1 & 2 & 4 & 7 & labeling cost \\
\midrule
Facility location & structural, $2$ & 6.05 $\pm$ 1.32 & 3.30 $\pm$ 1.14 & 1.11 $\pm$ 0.03 & 1.05 $\pm$ 0.03 & 28, 56, 112, 196 \\
& structural, $3$ & 1.60 $\pm$ 0.14 & 1.52 $\pm$ 0.17 & 1.10 $\pm$ 0.07 & 1.00 $\pm$ 0.00 & 63, 126, 252, 441 \\
& structural and lookahead, $2$ & 11.16 $\pm$ 1.15 & 6.78 $\pm$ 1.27 & 4.40 $\pm$ 1.21 & 1.10 $\pm$ 0.07 & 28, 56, 112, 196 \\
& structural and lookahead, $3$ & 1.06 $\pm$ 0.03 & 1.05 $\pm$ 0.04 & 1.00 $\pm$ 0.00 & 1.00 $\pm$ 0.00 & 63, 126, 252, 441 \\
Lot sizing & structural, $2$ & 1.40 $\pm$ 0.02 & 1.31 $\pm$ 0.02 & 1.25 $\pm$ 0.02 & 1.34 $\pm$ 0.06 & 45, 90, 180, 315 \\
& structural, $3$ & 3.60 $\pm$ 0.28 & 4.05 $\pm$ 0.32 & 4.58 $\pm$ 0.53 & 3.05 $\pm$ 0.70 & 129, 258, 516, 903 \\
& structural and lookahead, $2$ & 1.19 $\pm$ 0.05 & 1.01 $\pm$ 0.01 & 1.00 $\pm$ 0.00 & 1.00 $\pm$ 0.00 & 45, 90, 180, 315 \\
& structural and lookahead, $3$ & 1.57 $\pm$ 0.17 & 1.45 $\pm$ 0.13 & 1.82 $\pm$ 0.24 & 1.10 $\pm$ 0.07 & 129, 258, 516, 903 \\
Network design & structural, $2$ & 5.39 $\pm$ 1.11 & 3.84 $\pm$ 1.13 & 1.00 $\pm$ 0.00 & 1.00 $\pm$ 0.00 & 28, 56, 112, 196 \\
& structural, $3$ & 1.00 $\pm$ 0.00 & 1.00 $\pm$ 0.00 & 1.00 $\pm$ 0.00 & 1.00 $\pm$ 0.00 & 63, 126, 252, 441 \\
& structural and lookahead, $2$ & 5.39 $\pm$ 1.11 & 3.84 $\pm$ 1.13 & 1.00 $\pm$ 0.00 & 1.00 $\pm$ 0.00 & 28, 56, 112, 196 \\
& structural and lookahead, $3$ & 1.00 $\pm$ 0.00 & 1.00 $\pm$ 0.00 & 1.00 $\pm$ 0.00 & 1.00 $\pm$ 0.00 & 63, 126, 252, 441 \\
\bottomrule
\end{tabular}
\end{table}

\paragraph{Benchmark instances.} \Cref{tab:app-bench} gives, per class and method, the share of instances on which the first set is the best pair, its gain as a share of the LP gap, and the number of instances on which this gain is zero. It also gives the share on which the first set is the time-optimal set, strictly or within a tolerance, and its time ratio. A set lies within the tolerance if it solves on as many seeds as the time-optimal set and its time statistic is at most 1.1 times that of the time-optimal set. On lot sizing, every method chooses the best pair, and GCG as shipped, whose detector chooses the same pair, solves these instances in about 12 percent less time than our runs of this pair. GCG's detector convexifies capacity with linking on facility location and arc linking with arc capacity on 18 of the 20 network-design instances; on the other two, its decomposition matches no set of families. \Cref{tab:app-bench-paired} gives the paired differences in hit@1 and in the gain of the first set, and \cref{tab:app-bench-time} the paired differences in the time-optimal hit and in the logarithm of the time ratio. \Cref{tab:app-tradeoff} gives, per class, the share of solved runs, the time statistic, the share of the solve time spent in pricing, and the number of nodes for the best pair, the other pairs, the single families, and the set of all three families. On facility location, assignment alone or assignment with linking is faster than the best pair on 7 of the 19 instances. On three lot-sizing instances, no set of at most two families solves on any seed, so every such set lies within the tolerance, and the time-optimal set is the one with the smallest final gap: the best pair on two of these instances and setup linking alone on the third. The set of all three families leaves an empty master, so its pricing problem is the program itself. With this set, branch-and-price solves every benchmark instance on every seed, in 1.1 seconds at the median, and SCIP alone solves each instance in less than a minute.

\begin{table}[!htbp]
\caption{Benchmark instances, per class and method: share of instances on which the first set is the best pair, gain of the first set as a share of the LP gap, number of instances on which this gain is zero, share on which the first set is the time-optimal set, strictly and within the tolerance (as many solved seeds and at most 1.1 times its time statistic), and time ratio. Mean $\pm$ standard error of the two-stage bootstrap; time ratios as geometric means, with an asterisk when the 95 percent interval of the mean log ratio excludes zero. The best pair is a reference. JointDW uses the original presentation with three samples per instance, averaged within an instance.}
\label{tab:app-bench}
\centering
\scriptsize
\setlength{\tabcolsep}{3pt}
\begin{tabular}{@{}lcccccc@{}}
\toprule
& \multicolumn{3}{c}{first set and the best pair} & \multicolumn{3}{c}{first set and the time-optimal set} \\
\cmidrule(lr){2-4}\cmidrule(l){5-7}
Method & is best & gain & zero gain & is time-optimal & within 10\% & time ratio \\
\midrule
\multicolumn{7}{@{}l}{\emph{Lot sizing, 20 instances}} \\
Best pair & 1.00 $\pm$ 0.00 & 0.964 $\pm$ 0.005 & 0 & 0.95 $\pm$ 0.05 & 1.00 $\pm$ 0.00 & 1.000 $\pm$ 0.000 \\
Lookahead (HiGHS dual) & 1.00 $\pm$ 0.00 & 0.964 $\pm$ 0.005 & 0 & 0.95 $\pm$ 0.05 & 1.00 $\pm$ 0.00 & 1.000 $\pm$ 0.000 \\
Lookahead (five duals) & 1.00 $\pm$ 0.00 & 0.964 $\pm$ 0.005 & 0 & 0.95 $\pm$ 0.05 & 1.00 $\pm$ 0.00 & 1.000 $\pm$ 0.000 \\
GCG's detector & 1.00 $\pm$ 0.00 & 0.964 $\pm$ 0.005 & 0 & 0.95 $\pm$ 0.05 & 1.00 $\pm$ 0.00 & 0.882 $\pm$ 0.040$^{*}$ \\
JointDW with GPT-6.1 Sol & 1.00 $\pm$ 0.00 & 0.964 $\pm$ 0.005 & 0 & 0.95 $\pm$ 0.05 & 1.00 $\pm$ 0.00 & 1.000 $\pm$ 0.000 \\
JointDW with Claude Sonnet 5.5 & 1.00 $\pm$ 0.00 & 0.964 $\pm$ 0.005 & 0 & 0.95 $\pm$ 0.05 & 1.00 $\pm$ 0.00 & 1.000 $\pm$ 0.000 \\
\multicolumn{7}{@{}l}{\emph{Network design, 20 instances}} \\
Best pair & 1.00 $\pm$ 0.00 & 1.000 $\pm$ 0.000 & 0 & 1.00 $\pm$ 0.00 & 1.00 $\pm$ 0.00 & 1.000 $\pm$ 0.000 \\
Lookahead (HiGHS dual) & 0.35 $\pm$ 0.15 & 0.628 $\pm$ 0.098 & 4 & 0.30 $\pm$ 0.15 & 0.35 $\pm$ 0.15 & 4.740 $\pm$ 3.091$^{*}$ \\
Lookahead (five duals) & 0.35 $\pm$ 0.15 & 0.642 $\pm$ 0.089 & 3 & 0.30 $\pm$ 0.15 & 0.35 $\pm$ 0.15 & 4.370 $\pm$ 2.791$^{*}$ \\
GCG's detector & 0.00 $\pm$ 0.00 & 0.003 $\pm$ 0.005 & 19 & 0.00 $\pm$ 0.00 & 0.00 $\pm$ 0.00 & 4.062 $\pm$ 0.770$^{*}$ \\
JointDW with GPT-6.1 Sol & 0.67 $\pm$ 0.16 & 0.807 $\pm$ 0.102 & 1 & 0.65 $\pm$ 0.15 & 0.73 $\pm$ 0.12 & 1.921 $\pm$ 0.540$^{*}$ \\
JointDW with Claude Sonnet 5.5 & 0.75 $\pm$ 0.15 & 0.827 $\pm$ 0.126 & 1 & 0.75 $\pm$ 0.15 & 0.80 $\pm$ 0.12 & 1.769 $\pm$ 0.597 \\
\multicolumn{7}{@{}l}{\emph{Facility location, 19 instances}} \\
Best pair & 1.00 $\pm$ 0.00 & 1.000 $\pm$ 0.000 & 0 & 0.63 $\pm$ 0.15 & 0.68 $\pm$ 0.15 & 1.099 $\pm$ 0.068$^{*}$ \\
Lookahead (HiGHS dual) & 0.68 $\pm$ 0.24 & 0.684 $\pm$ 0.242 & 6 & 0.42 $\pm$ 0.19 & 0.47 $\pm$ 0.21 & 1.743 $\pm$ 0.952$^{*}$ \\
Lookahead (five duals) & 0.95 $\pm$ 0.07 & 0.947 $\pm$ 0.072 & 1 & 0.58 $\pm$ 0.16 & 0.63 $\pm$ 0.15 & 1.121 $\pm$ 0.068$^{*}$ \\
GCG's detector & 0.00 $\pm$ 0.00 & 0.000 $\pm$ 0.000 & 19 & 0.00 $\pm$ 0.00 & 0.00 $\pm$ 0.00 & 3.469 $\pm$ 0.432$^{*}$ \\
JointDW with GPT-6.1 Sol & 1.00 $\pm$ 0.00 & 1.000 $\pm$ 0.000 & 0 & 0.63 $\pm$ 0.15 & 0.68 $\pm$ 0.15 & 1.099 $\pm$ 0.068$^{*}$ \\
JointDW with Claude Sonnet 5.5 & 0.68 $\pm$ 0.02 & 0.684 $\pm$ 0.024 & 6 & 0.47 $\pm$ 0.10 & 0.53 $\pm$ 0.13 & 1.431 $\pm$ 0.197$^{*}$ \\
\multicolumn{7}{@{}l}{\emph{All instances, 59 instances}} \\
Best pair & 1.00 $\pm$ 0.00 & 0.988 $\pm$ 0.009 & 0 & 0.86 $\pm$ 0.08 & 0.90 $\pm$ 0.08 & 1.031 $\pm$ 0.029 \\
Lookahead (HiGHS dual) & 0.68 $\pm$ 0.16 & 0.760 $\pm$ 0.115 & 10 & 0.56 $\pm$ 0.16 & 0.61 $\pm$ 0.17 & 2.027 $\pm$ 0.864$^{*}$ \\
Lookahead (five duals) & 0.76 $\pm$ 0.13 & 0.849 $\pm$ 0.073 & 4 & 0.61 $\pm$ 0.15 & 0.66 $\pm$ 0.15 & 1.711 $\pm$ 0.622$^{*}$ \\
GCG's detector & 0.34 $\pm$ 0.24 & 0.328 $\pm$ 0.228 & 38 & 0.32 $\pm$ 0.22 & 0.34 $\pm$ 0.23 & 2.300 $\pm$ 0.810$^{*}$ \\
JointDW with GPT-6.1 Sol & 0.89 $\pm$ 0.08 & 0.922 $\pm$ 0.045 & 1 & 0.75 $\pm$ 0.10 & 0.81 $\pm$ 0.10 & 1.286 $\pm$ 0.203$^{*}$ \\
JointDW with Claude Sonnet 5.5 & 0.81 $\pm$ 0.09 & 0.827 $\pm$ 0.072 & 7 & 0.73 $\pm$ 0.11 & 0.78 $\pm$ 0.11 & 1.362 $\pm$ 0.237$^{*}$ \\
\bottomrule
\end{tabular}
\end{table}

\begin{table}[!htbp]
\caption{Benchmark instances: paired differences in the share of instances on which the first set is the time-optimal set (all instances) and in the logarithm of the time ratio (per class and all instances), JointDW under the original presentation minus a baseline, and the five-dual lookahead minus GCG's detector. Mean $\pm$ standard error of the two-stage bootstrap, with an asterisk when the 95 percent interval excludes zero. Negative log ratios favor the first method.}
\label{tab:app-bench-time}
\centering
\scriptsize
\setlength{\tabcolsep}{3pt}
\resizebox{\textwidth}{!}{%
\begin{tabular}{@{}llccccc@{}}
\toprule
& & \multicolumn{4}{c}{difference in the log time ratio} & time-optimal \\
\cmidrule(lr){3-6}
Method & minus & LS & ND & FL & all & all \\
\midrule
JointDW with GPT-6.1 Sol & Lookahead (HiGHS dual) & 0.00 $\pm$ 0.00 & $-$0.90 $\pm$ 0.82 & $-$0.46 $\pm$ 0.56 & $-$0.45 $\pm$ 0.39 & 0.19 $\pm$ 0.12 \\
 & Lookahead (five duals) & 0.00 $\pm$ 0.00 & $-$0.82 $\pm$ 0.83 & $-$0.02 $\pm$ 0.03 & $-$0.29 $\pm$ 0.34 & 0.14 $\pm$ 0.11 \\
 & GCG's detector & 0.13 $\pm$ 0.05$^{*}$ & $-$0.75 $\pm$ 0.34$^{*}$ & $-$1.15 $\pm$ 0.12$^{*}$ & $-$0.58 $\pm$ 0.29$^{*}$ & 0.42 $\pm$ 0.17$^{*}$ \\
JointDW with Claude Sonnet 5.5 & Lookahead (HiGHS dual) & 0.00 $\pm$ 0.00 & $-$0.99 $\pm$ 0.63 & $-$0.20 $\pm$ 0.53 & $-$0.40 $\pm$ 0.36 & 0.17 $\pm$ 0.15 \\
 & Lookahead (five duals) & 0.00 $\pm$ 0.00 & $-$0.90 $\pm$ 0.59 & 0.24 $\pm$ 0.10$^{*}$ & $-$0.23 $\pm$ 0.29 & 0.12 $\pm$ 0.15 \\
 & GCG's detector & 0.13 $\pm$ 0.05$^{*}$ & $-$0.83 $\pm$ 0.34$^{*}$ & $-$0.89 $\pm$ 0.13$^{*}$ & $-$0.52 $\pm$ 0.27$^{*}$ & 0.41 $\pm$ 0.17$^{*}$ \\
Lookahead (five duals) & GCG's detector & 0.13 $\pm$ 0.05$^{*}$ & 0.07 $\pm$ 0.53 & $-$1.13 $\pm$ 0.12$^{*}$ & $-$0.30 $\pm$ 0.30 & 0.29 $\pm$ 0.13$^{*}$ \\
\bottomrule
\end{tabular}}
\end{table}

\begin{table}[!htbp]
\caption{Benchmark instances: paired differences in hit@1 and in the gain of the first set as a share of the LP gap, JointDW under the original presentation minus a baseline, and the five-dual lookahead minus GCG's detector, on network design (ND, 20 instances), facility location (FL, 19), and all 59 instances; on lot sizing, every difference is zero. Mean $\pm$ standard error of the two-stage bootstrap, with an asterisk when the 95 percent interval excludes zero.}
\label{tab:app-bench-paired}
\centering
\scriptsize
\setlength{\tabcolsep}{3pt}
\resizebox{\textwidth}{!}{%
\begin{tabular}{@{}llcccccc@{}}
\toprule
& & \multicolumn{3}{c}{difference in hit@1} & \multicolumn{3}{c}{difference in gain} \\
\cmidrule(lr){3-5}\cmidrule(l){6-8}
Method & minus & ND & FL & all & ND & FL & all \\
\midrule
JointDW with GPT-6.1 Sol & Lookahead (HiGHS dual) & 0.32 $\pm$ 0.24 & 0.32 $\pm$ 0.24 & 0.21 $\pm$ 0.14 & 0.18 $\pm$ 0.15 & 0.32 $\pm$ 0.24 & 0.16 $\pm$ 0.11 \\
 & Lookahead (five duals) & 0.32 $\pm$ 0.24 & 0.05 $\pm$ 0.07 & 0.12 $\pm$ 0.11 & 0.17 $\pm$ 0.14 & 0.05 $\pm$ 0.07 & 0.07 $\pm$ 0.06 \\
 & GCG's detector & 0.67 $\pm$ 0.16$^{*}$ & 1.00 $\pm$ 0.00$^{*}$ & 0.55 $\pm$ 0.21$^{*}$ & 0.80 $\pm$ 0.11$^{*}$ & 1.00 $\pm$ 0.00$^{*}$ & 0.59 $\pm$ 0.22$^{*}$ \\
JointDW with Claude Sonnet 5.5 & Lookahead (HiGHS dual) & 0.40 $\pm$ 0.25 & 0.00 $\pm$ 0.26 & 0.14 $\pm$ 0.15 & 0.20 $\pm$ 0.17 & 0.00 $\pm$ 0.26 & 0.07 $\pm$ 0.11 \\
 & Lookahead (five duals) & 0.40 $\pm$ 0.25 & $-$0.26 $\pm$ 0.10$^{*}$ & 0.05 $\pm$ 0.14 & 0.19 $\pm$ 0.16 & $-$0.26 $\pm$ 0.10$^{*}$ & $-$0.02 $\pm$ 0.10 \\
 & GCG's detector & 0.75 $\pm$ 0.15$^{*}$ & 0.68 $\pm$ 0.02$^{*}$ & 0.47 $\pm$ 0.18$^{*}$ & 0.82 $\pm$ 0.13$^{*}$ & 0.68 $\pm$ 0.02$^{*}$ & 0.50 $\pm$ 0.19$^{*}$ \\
Lookahead (five duals) & GCG's detector & 0.35 $\pm$ 0.15$^{*}$ & 0.95 $\pm$ 0.07$^{*}$ & 0.42 $\pm$ 0.18$^{*}$ & 0.64 $\pm$ 0.09$^{*}$ & 0.95 $\pm$ 0.07$^{*}$ & 0.52 $\pm$ 0.20$^{*}$ \\
\bottomrule
\end{tabular}}
\end{table}

\begin{table}[!htbp]
\caption{Benchmark instances, branch-and-price on the sets of at most two families and on all three, per class: number of sets, share of runs solved within 1800 seconds, mean time statistic in seconds, mean share of the solve time spent in pricing, and mean number of nodes. The set of all three families leaves an empty master; its pricing share and nodes are not comparable and are omitted.}
\label{tab:app-tradeoff}
\centering
\scriptsize
\setlength{\tabcolsep}{4pt}
\begin{tabular}{@{}lccccc@{}}
\toprule
Sets & number & solved & time statistic (s) & pricing share & nodes \\
\midrule
\multicolumn{6}{@{}l}{\emph{Lot sizing, 20 instances}} \\
Best pair & 20 & 0.85 & 571.6 & 0.78 & 13913 \\
Other pairs & 40 & 0.00 & 1801.0 & 0.25 & 12269 \\
Single families & 60 & 0.11 & 1665.1 & 0.17 & 23232 \\
All three families & 20 & 1.00 & 6.7 & -- & -- \\
\multicolumn{6}{@{}l}{\emph{Network design, 20 instances}} \\
Best pair & 20 & 1.00 & 3.7 & 0.96 & 1 \\
Other pairs & 40 & 0.87 & 288.6 & 0.82 & 131 \\
Single families & 60 & 1.00 & 26.8 & 0.67 & 153 \\
All three families & 20 & 1.00 & 2.8 & -- & -- \\
\multicolumn{6}{@{}l}{\emph{Facility location, 19 instances}} \\
Best pair & 19 & 1.00 & 2.4 & 0.93 & 1 \\
Other pairs & 38 & 0.99 & 7.4 & 0.92 & 3 \\
Single families & 57 & 0.94 & 101.9 & 0.79 & 3 \\
All three families & 19 & 1.00 & 1.3 & -- & -- \\
\bottomrule
\end{tabular}
\end{table}

\paragraph{Benchmark instances with every coefficient shown.} \Cref{tab:app-bench-data} gives JointDW with GPT-6.1 Sol and Claude Sonnet 5.5 on the benchmark instances under the coefficient variant of the prompt, the complete instance in LP format as for the rich models, with three samples per instance. Every instance then has its own prompt, so the bootstrap resamples instances. GPT-6.1 Sol returns a joint support in every answer, and Claude Sonnet 5.5 in 159 of 177; three answers of Claude Sonnet 5.5 name a set of three families, which the gain tables of these instances, over sets of at most two families, score as a miss without gain. Relative to the original presentation, hit@1 is lower by $0.11\pm0.04$ for GPT-6.1 Sol and by $0.25\pm0.05$ for Claude Sonnet 5.5, both beyond sampling variation and mostly on network design, where hit@1 is 0.40 and 0.17 against 0.67 and 0.75 under the original presentation. The logarithm of the time ratio is higher by $0.47\pm0.17$ and $0.65\pm0.18$, also beyond sampling variation. Against the baselines, the first set of GPT-6.1 Sol gains more of the LP gap than that of the lookahead at the HiGHS dual, by $0.105\pm0.051$, and than that of the detector, by $0.54\pm0.06$, and differs from that of the five-dual lookahead by $0.02\pm0.04$. The first set of Claude Sonnet 5.5 gains more than that of the detector and less than that of the five-dual lookahead, beyond sampling variation.

\begin{table}[!htbp]
\caption{Benchmark instances with every coefficient shown, JointDW with three samples per instance, averaged within an instance. Top: hit@1, gain of the first set as a share of the LP gap, share of instances on which the first set is the time-optimal set, time ratio, and number of answers without a joint support. Bottom: paired differences, JointDW with every coefficient shown minus a baseline or minus the same configuration in the original presentation, in hit@1, in the gain, and in the logarithm of the time ratio. Mean $\pm$ standard error of the bootstrap over instances, with an asterisk when the 95 percent interval excludes zero (time ratio: of the mean log ratio).}
\label{tab:app-bench-data}
\centering
\scriptsize
\setlength{\tabcolsep}{3pt}
\begin{tabular}{@{}lccccc@{}}
\toprule
JointDW with & hit@1 & gain of first set & time-optimal & time ratio & no joint support \\
\midrule
\multicolumn{6}{@{}l}{\emph{Lot sizing, 20 instances}} \\
GPT-6.1 Sol & 1.00 $\pm$ 0.00 & 0.964 $\pm$ 0.005 & 0.95 $\pm$ 0.05 & 1.000 $\pm$ 0.000 & 0 \\
Claude Sonnet 5.5 & 0.95 $\pm$ 0.03 & 0.934 $\pm$ 0.020 & 0.90 $\pm$ 0.05 & 0.994 $\pm$ 0.076 & 1 \\
\multicolumn{6}{@{}l}{\emph{Network design, 20 instances}} \\
GPT-6.1 Sol & 0.40 $\pm$ 0.07 & 0.688 $\pm$ 0.049 & 0.37 $\pm$ 0.07 & 7.658 $\pm$ 3.665$^{*}$ & 0 \\
Claude Sonnet 5.5 & 0.17 $\pm$ 0.06 & 0.514 $\pm$ 0.065 & 0.13 $\pm$ 0.04 & 10.777 $\pm$ 5.153$^{*}$ & 0 \\
\multicolumn{6}{@{}l}{\emph{Facility location, 19 instances}} \\
GPT-6.1 Sol & 0.95 $\pm$ 0.03 & 0.947 $\pm$ 0.028 & 0.65 $\pm$ 0.11 & 1.097 $\pm$ 0.041$^{*}$ & 0 \\
Claude Sonnet 5.5 & 0.58 $\pm$ 0.06 & 0.596 $\pm$ 0.062 & 0.37 $\pm$ 0.08 & 1.641 $\pm$ 0.167$^{*}$ & 17 \\
\multicolumn{6}{@{}l}{\emph{All instances, 59 instances}} \\
GPT-6.1 Sol & 0.78 $\pm$ 0.04 & 0.865 $\pm$ 0.025 & 0.66 $\pm$ 0.05 & 2.054 $\pm$ 0.419$^{*}$ & 0 \\
Claude Sonnet 5.5 & 0.56 $\pm$ 0.05 & 0.683 $\pm$ 0.039 & 0.47 $\pm$ 0.05 & 2.620 $\pm$ 0.567$^{*}$ & 18 \\
\bottomrule
\end{tabular}

\medskip
\begin{tabular}{@{}llccc@{}}
\toprule
& & \multicolumn{3}{c}{difference in} \\
\cmidrule(l){3-5}
JointDW with & minus & hit@1 & gain & log time ratio \\
\midrule
\multicolumn{5}{@{}l}{\emph{Network design, 20 instances}} \\
GPT-6.1 Sol & Lookahead (HiGHS dual) & 0.05 $\pm$ 0.15 & 0.059 $\pm$ 0.099 & 0.48 $\pm$ 0.45 \\
 & Lookahead (five duals) & 0.05 $\pm$ 0.15 & 0.046 $\pm$ 0.096 & 0.56 $\pm$ 0.56 \\
 & GCG's detector & 0.40 $\pm$ 0.07$^{*}$ & 0.685 $\pm$ 0.049$^{*}$ & 0.63 $\pm$ 0.39 \\
 & Original presentation & $-$0.27 $\pm$ 0.10$^{*}$ & $-$0.120 $\pm$ 0.052$^{*}$ & 1.38 $\pm$ 0.45$^{*}$ \\
Claude Sonnet 5.5 & Lookahead (HiGHS dual) & $-$0.18 $\pm$ 0.10 & $-$0.114 $\pm$ 0.065 & 0.82 $\pm$ 0.38$^{*}$ \\
 & Lookahead (five duals) & $-$0.18 $\pm$ 0.11 & $-$0.127 $\pm$ 0.062$^{*}$ & 0.90 $\pm$ 0.47$^{*}$ \\
 & GCG's detector & 0.17 $\pm$ 0.06$^{*}$ & 0.511 $\pm$ 0.066$^{*}$ & 0.98 $\pm$ 0.38$^{*}$ \\
 & Original presentation & $-$0.58 $\pm$ 0.08$^{*}$ & $-$0.313 $\pm$ 0.071$^{*}$ & 1.81 $\pm$ 0.40$^{*}$ \\
\multicolumn{5}{@{}l}{\emph{Facility location, 19 instances}} \\
GPT-6.1 Sol & Lookahead (HiGHS dual) & 0.26 $\pm$ 0.11$^{*}$ & 0.263 $\pm$ 0.110$^{*}$ & $-$0.46 $\pm$ 0.38$^{*}$ \\
 & Lookahead (five duals) & 0.00 $\pm$ 0.06 & 0.000 $\pm$ 0.060 & $-$0.02 $\pm$ 0.02$^{*}$ \\
 & GCG's detector & 0.95 $\pm$ 0.03$^{*}$ & 0.947 $\pm$ 0.028$^{*}$ & $-$1.15 $\pm$ 0.12$^{*}$ \\
 & Original presentation & $-$0.05 $\pm$ 0.03 & $-$0.053 $\pm$ 0.028 & 0.00 $\pm$ 0.00 \\
Claude Sonnet 5.5 & Lookahead (HiGHS dual) & $-$0.11 $\pm$ 0.14 & $-$0.088 $\pm$ 0.139 & $-$0.06 $\pm$ 0.39 \\
 & Lookahead (five duals) & $-$0.37 $\pm$ 0.08$^{*}$ & $-$0.351 $\pm$ 0.082$^{*}$ & 0.38 $\pm$ 0.11$^{*}$ \\
 & GCG's detector & 0.58 $\pm$ 0.06$^{*}$ & 0.596 $\pm$ 0.062$^{*}$ & $-$0.75 $\pm$ 0.13$^{*}$ \\
 & Original presentation & $-$0.11 $\pm$ 0.07 & $-$0.088 $\pm$ 0.063 & 0.14 $\pm$ 0.08 \\
\multicolumn{5}{@{}l}{\emph{All instances, 59 instances}} \\
GPT-6.1 Sol & Lookahead (HiGHS dual) & 0.10 $\pm$ 0.06 & 0.105 $\pm$ 0.051$^{*}$ & 0.01 $\pm$ 0.20 \\
 & Lookahead (five duals) & 0.02 $\pm$ 0.06 & 0.016 $\pm$ 0.038 & 0.18 $\pm$ 0.19 \\
 & GCG's detector & 0.44 $\pm$ 0.06$^{*}$ & 0.537 $\pm$ 0.055$^{*}$ & $-$0.11 $\pm$ 0.17 \\
 & Original presentation & $-$0.11 $\pm$ 0.04$^{*}$ & $-$0.058 $\pm$ 0.021$^{*}$ & 0.47 $\pm$ 0.17$^{*}$ \\
Claude Sonnet 5.5 & Lookahead (HiGHS dual) & $-$0.11 $\pm$ 0.06$^{*}$ & $-$0.077 $\pm$ 0.051 & 0.26 $\pm$ 0.19 \\
 & Lookahead (five duals) & $-$0.20 $\pm$ 0.05$^{*}$ & $-$0.166 $\pm$ 0.039$^{*}$ & 0.43 $\pm$ 0.17$^{*}$ \\
 & GCG's detector & 0.23 $\pm$ 0.05$^{*}$ & 0.355 $\pm$ 0.048$^{*}$ & 0.13 $\pm$ 0.17 \\
 & Original presentation & $-$0.25 $\pm$ 0.05$^{*}$ & $-$0.145 $\pm$ 0.036$^{*}$ & 0.65 $\pm$ 0.18$^{*}$ \\
\bottomrule
\end{tabular}
\end{table}

\paragraph{Instances without a joint support.} \Cref{tab:app-nojoint} gives the methods on the generalized assignment and bin-packing instances. The last time ratio is taken against GCG as shipped, whose detector convexifies the capacity rows on generalized assignment and the assignment rows with the symmetry-breaking rows on bin packing. On generalized assignment, GPT-6.1 Sol proposes no reformulation in 10 of its 180 answers, which count as misses, and chooses the capacity family in all others. The 13 false alarms of Claude Sonnet 5.5 are answers that name a joint support on the 18 instances where the capacity family alone closes the whole gap. Its other answers that name a joint support name assignment with capacity, a set outside the budget, which keeps its measured time in the time ratios.

\begin{table}[!htbp]
\caption{Instances without a joint support, per class and method: hit@1 against the set with the best bound (bin packing: the 12 instances where some set within the budget has a positive gain) and against the time-optimal set, time ratio to the time-optimal set and to GCG as shipped, false alarms (answers that name a joint support on an instance where the capacity family alone closes the whole gap), and samples or instances whose set lies outside the budget. Mean $\pm$ standard error of the two-stage bootstrap; time ratios as geometric means, with an asterisk when the 95 percent interval of the mean log ratio excludes zero.}
\label{tab:app-nojoint}
\centering
\scriptsize
\setlength{\tabcolsep}{3pt}
\begin{tabular}{@{}lcccccc@{}}
\toprule
& \multicolumn{2}{c}{hit@1 against} & \multicolumn{2}{c}{time ratio to} & false & outside \\
\cmidrule(lr){2-3}\cmidrule(lr){4-5}
Method & best bound & time-optimal & time-optimal & GCG as shipped & alarms & budget \\
\midrule
\multicolumn{7}{@{}l}{\emph{Generalized assignment, 60 instances}} \\
Best set & 1.00 $\pm$ 0.00 & 1.00 $\pm$ 0.00 & 1.00 $\pm$ 0.00 & 1.01 $\pm$ 0.02 & -- & 0 \\
Lookahead (HiGHS dual) & 0.97 $\pm$ 0.03 & 0.97 $\pm$ 0.03 & 1.01 $\pm$ 0.01 & 1.02 $\pm$ 0.03 & -- & 0 \\
Lookahead (five duals) & 1.00 $\pm$ 0.00 & 1.00 $\pm$ 0.00 & 1.00 $\pm$ 0.00 & 1.01 $\pm$ 0.02 & -- & 0 \\
GCG's detector & 1.00 $\pm$ 0.00 & 1.00 $\pm$ 0.00 & 0.99 $\pm$ 0.02 & 1.00 $\pm$ 0.00 & -- & 0 \\
JointDW with GPT-6.1 Sol & 0.94 $\pm$ 0.05 & 0.94 $\pm$ 0.05 & 1.00 $\pm$ 0.00 & 1.01 $\pm$ 0.02 & 0 & 0 \\
JointDW with Claude Sonnet 5.5 & 0.75 $\pm$ 0.06 & 0.75 $\pm$ 0.06 & 0.99 $\pm$ 0.01 & 1.00 $\pm$ 0.02 & 13 & 45 \\
\multicolumn{7}{@{}l}{\emph{Bin packing, 16 instances}} \\
Best set & 1.00 $\pm$ 0.00 & 0.75 $\pm$ 0.18 & 3.12 $\pm$ 2.67 & 1.36 $\pm$ 1.12 & -- & 0 \\
Lookahead (HiGHS dual) & 0.00 $\pm$ 0.00 & 0.00 $\pm$ 0.00 & 86.16 $\pm$ 24.54$^{*}$ & 37.53 $\pm$ 12.13$^{*}$ & -- & 3 \\
Lookahead (five duals) & 0.00 $\pm$ 0.00 & 0.00 $\pm$ 0.00 & 86.16 $\pm$ 24.54$^{*}$ & 37.53 $\pm$ 12.13$^{*}$ & -- & 15 \\
GCG's detector & 0.00 $\pm$ 0.00 & 0.00 $\pm$ 0.00 & 2.30 $\pm$ 0.77$^{*}$ & 1.00 $\pm$ 0.00 & -- & 0 \\
JointDW with GPT-6.1 Sol & 1.00 $\pm$ 0.00 & 1.00 $\pm$ 0.00 & 1.00 $\pm$ 0.00 & 0.44 $\pm$ 0.15$^{*}$ & 0 & 0 \\
JointDW with Claude Sonnet 5.5 & 0.92 $\pm$ 0.09 & 0.77 $\pm$ 0.16 & 2.80 $\pm$ 2.16 & 1.22 $\pm$ 0.90 & 0 & 11 \\
\multicolumn{7}{@{}l}{\emph{All instances, 76 instances}} \\
Best set & 1.00 $\pm$ 0.00 & 0.95 $\pm$ 0.04 & 1.27 $\pm$ 0.23 & 1.07 $\pm$ 0.17 & -- & 0 \\
Lookahead (HiGHS dual) & 0.81 $\pm$ 0.10 & 0.76 $\pm$ 0.10 & 2.58 $\pm$ 1.16$^{*}$ & 2.18 $\pm$ 0.81$^{*}$ & -- & 3 \\
Lookahead (five duals) & 0.83 $\pm$ 0.10 & 0.79 $\pm$ 0.10 & 2.56 $\pm$ 1.15$^{*}$ & 2.16 $\pm$ 0.81$^{*}$ & -- & 15 \\
GCG's detector & 0.83 $\pm$ 0.10 & 0.79 $\pm$ 0.10 & 1.18 $\pm$ 0.13$^{*}$ & 1.00 $\pm$ 0.00 & -- & 0 \\
JointDW with GPT-6.1 Sol & 0.95 $\pm$ 0.04 & 0.96 $\pm$ 0.04 & 1.00 $\pm$ 0.00 & 0.84 $\pm$ 0.09$^{*}$ & 0 & 0 \\
JointDW with Claude Sonnet 5.5 & 0.78 $\pm$ 0.05 & 0.75 $\pm$ 0.05 & 1.23 $\pm$ 0.20 & 1.04 $\pm$ 0.14 & 13 & 56 \\
\bottomrule
\end{tabular}
\end{table}

\paragraph{Rich models with 12 to 16 families.} \Cref{tab:app-rich-large} gives hit@1, recall, and the search cost on these models by number of families. GPT-6.1 Sol given every coefficient names a true group first in 178 of its 180 answers, and its recall falls from 0.85 with 12 families to 0.56 with 16, where the models have more joint supports. Interleaved with the five-dual order, JointDW with GPT-6.1 Sol given every coefficient needs $5.16\pm1.75$ evaluations fewer than the five-dual order for $s=2$, beyond sampling variation, and JointDW alone $3.71\pm2.11$ fewer, within it. For $s=3$, the differences are $4.98\pm4.83$ fewer and $0.04\pm5.54$ more, within sampling variation. Pooled over the models, every JointDW order saves evaluations against the covering order beyond sampling variation for $s=2$ and $s=3$. The savings of JointDW with GPT-6.1 Sol given every coefficient grow from $23.5\pm1.9$ evaluations for $s=2$ with 12 families to $43.7\pm4.5$ with 16. GCG's decomposition by its default score exists on 15 of the 60 models, and as an anchor of the staged search it needs $0.90\pm0.09$ evaluations more than the covering order for $s=2$.

\begin{table}[!htbp]
\caption{Rich models with 12 to 16 families, pooled over the classes. Top: hit@1 and recall given the text and given every coefficient. Bottom: number of evaluations until a best set is evaluated for $s=2$ and $s=3$. Columns 12: the 30 models with 12 families; 16: the 20 lot-sizing and facility-location models with 16 families; all: the 60 models, including the network-design models with 14 families, which enter only for $s=2$. Mean $\pm$ standard error of the two-stage bootstrap.}
\label{tab:app-rich-large}
\centering
\scriptsize
\setlength{\tabcolsep}{3pt}
\begin{tabular}{@{}lcccccc@{}}
\toprule
& \multicolumn{3}{c}{hit@1} & \multicolumn{3}{c}{recall} \\
\cmidrule(lr){2-4}\cmidrule(l){5-7}
Configuration & 12 & 16 & all & 12 & 16 & all \\
\midrule
GPT-6.1 Sol, text & 0.09 $\pm$ 0.04 & 0.07 $\pm$ 0.03 & 0.07 $\pm$ 0.02 & 0.06 $\pm$ 0.03 & 0.02 $\pm$ 0.01 & 0.04 $\pm$ 0.02 \\
GPT-6.1 Sol, coefficients & 0.99 $\pm$ 0.01 & 1.00 $\pm$ 0.00 & 0.99 $\pm$ 0.01 & 0.85 $\pm$ 0.03 & 0.56 $\pm$ 0.06 & 0.75 $\pm$ 0.03 \\
Claude Sonnet 5.5, text & 0.18 $\pm$ 0.06 & 0.10 $\pm$ 0.05 & 0.13 $\pm$ 0.04 & 0.11 $\pm$ 0.03 & 0.03 $\pm$ 0.01 & 0.08 $\pm$ 0.02 \\
Claude Sonnet 5.5, coefficients & 0.28 $\pm$ 0.06 & 0.37 $\pm$ 0.08 & 0.27 $\pm$ 0.04 & 0.14 $\pm$ 0.03 & 0.07 $\pm$ 0.02 & 0.11 $\pm$ 0.02 \\
\bottomrule
\end{tabular}

\medskip
\begin{tabular}{@{}lcccccc@{}}
\toprule
& \multicolumn{3}{c}{$s=2$} & \multicolumn{3}{c}{$s=3$} \\
\cmidrule(lr){2-4}\cmidrule(l){5-7}
Order & 12 & 16 & all & 12 & 16 & all \\
\midrule
Covering order & 25.66 $\pm$ 2.35 & 47.58 $\pm$ 4.96 & 33.73 $\pm$ 2.91 & 44.92 $\pm$ 8.49 & 97.27 $\pm$ 18.58 & 65.86 $\pm$ 9.87 \\
Lookahead (HiGHS dual) & 17.45 $\pm$ 4.68 & 22.56 $\pm$ 8.30 & 20.31 $\pm$ 3.86 & 23.32 $\pm$ 11.02 & 32.57 $\pm$ 23.17 & 27.02 $\pm$ 11.07 \\
Lookahead (five duals) & 7.62 $\pm$ 3.12 & 4.12 $\pm$ 3.24 & 6.97 $\pm$ 2.17 & 2.63 $\pm$ 1.25 & 14.36 $\pm$ 13.66 & 7.32 $\pm$ 5.40 \\
\multicolumn{7}{@{}l}{\emph{JointDW}} \\
GPT-6.1 Sol, text & 20.58 $\pm$ 2.02 & 39.37 $\pm$ 4.66 & 28.36 $\pm$ 2.63 & 36.54 $\pm$ 6.49 & 86.92 $\pm$ 15.57 & 56.69 $\pm$ 8.19 \\
GPT-6.1 Sol, coefficients & 2.15 $\pm$ 0.90 & 3.90 $\pm$ 1.99 & 3.26 $\pm$ 1.01 & 8.84 $\pm$ 3.42 & 5.15 $\pm$ 1.17 & 7.37 $\pm$ 2.11 \\
Claude Sonnet 5.5, text & 13.90 $\pm$ 1.49 & 25.26 $\pm$ 3.15 & 18.70 $\pm$ 1.70 & 20.89 $\pm$ 4.28 & 55.80 $\pm$ 15.71 & 34.85 $\pm$ 7.12 \\
Claude Sonnet 5.5, coefficients & 11.34 $\pm$ 1.30 & 16.77 $\pm$ 2.77 & 14.46 $\pm$ 1.31 & 14.99 $\pm$ 3.30 & 38.18 $\pm$ 9.25 & 24.27 $\pm$ 4.52 \\
\multicolumn{7}{@{}l}{\emph{JointDW interleaved with the five-dual order}} \\
GPT-6.1 Sol, text & 7.30 $\pm$ 2.65 & 4.73 $\pm$ 3.06 & 6.88 $\pm$ 1.88 & 3.40 $\pm$ 1.02 & 10.33 $\pm$ 8.27 & 6.17 $\pm$ 3.30 \\
GPT-6.1 Sol, coefficients & 1.41 $\pm$ 0.36 & 2.74 $\pm$ 1.67 & 1.81 $\pm$ 0.59 & 2.21 $\pm$ 0.63 & 2.56 $\pm$ 0.86 & 2.35 $\pm$ 0.51 \\
Claude Sonnet 5.5, text & 6.66 $\pm$ 2.35 & 4.29 $\pm$ 2.70 & 6.54 $\pm$ 1.80 & 3.30 $\pm$ 0.93 & 7.85 $\pm$ 5.79 & 5.12 $\pm$ 2.33 \\
Claude Sonnet 5.5, coefficients & 5.92 $\pm$ 1.79 & 4.65 $\pm$ 3.11 & 6.30 $\pm$ 1.68 & 3.05 $\pm$ 0.74 & 5.37 $\pm$ 3.08 & 3.98 $\pm$ 1.32 \\
\bottomrule
\end{tabular}
\end{table}

\paragraph{Search cost against the number of misses.} \Cref{tab:app-misses} gives the search cost for pairs by the number $e$ of members of a best set that the anchor misses, for the predictions of GPT-6.1 Sol and Claude Sonnet 5.5 under both presentations on the rich models. Instances on which no set within the budget has a positive gain are left out. The number $e$ is counted on the anchor that the search uses. For an anchor of at most $s$ families, it is the smallest number of families of a minimal best set that lie outside the anchor. An anchor with more than $s$ families is cut to its first $s$ families in the order of the families, so its number of misses depends on this order. Over the 1,832 combinations of a prediction and a budget $s\in\{2,3\}$, the staged order alone costs one evaluation whenever $e=0$. Whenever the best set has at most as many families as the anchor, its cost is at most the number of queries of stages $0$ to $e$ in every order of the families. For $s=3$, the best set has three families and the anchor two in 193 predictions (\cref{tab:app-larger}). The staged order of \cref{thm:pred} covers groups of the size of the anchor, so it reaches such a best set only in some orders of the families, and JointDW reaches one in every order through $O_2$.

\begin{table}[!htbp]
\caption{Rich models, $s=2$: number of evaluations until a best pair is evaluated, by the number $e$ of its members that the anchor misses, for the staged order $O_1$ alone and for JointDW, with the number of queries of stages $0$ to $e$ of \cref{thm:pred}(a) and the mean cost of the covering order on the same instances. Rows marked cut: anchors with more than two families, with their mean number of misses after the cut in parentheses. Predictions of GPT-6.1 Sol and Claude Sonnet 5.5 under both presentations. Mean $\pm$ standard error of the two-stage bootstrap.}
\label{tab:app-misses}
\centering
\footnotesize
\setlength{\tabcolsep}{5pt}
\begin{tabular}{@{}lclccccc@{}}
\toprule
Class & $k$ & $e$ & predictions & staged order & JointDW & stage queries & covering \\
\midrule
Lot sizing & 9 & 0 & 62 & 1.00 $\pm$ 0.00 & 1.00 $\pm$ 0.00 & 1 & 18.7 \\
 &  & 1 & 4 & 8.57 $\pm$ 0.08 & 12.25 $\pm$ 0.14 & 15 & 18.9 \\
 &  & 2 & 3 & 26.00 $\pm$ 0.11 & 23.44 $\pm$ 0.16 & 36 & 18.4 \\
 &  & cut (0.91) & 27 & 10.54 $\pm$ 0.84 & 12.47 $\pm$ 0.67 & -- & -- \\
 & 12 & 0 & 72 & 1.00 $\pm$ 0.00 & 1.00 $\pm$ 0.00 & 1 & 33.7 \\
 &  & 1 & 5 & 11.47 $\pm$ 0.09 & 17.79 $\pm$ 0.12 & 21 & 33.8 \\
 &  & 2 & 3 & 43.82 $\pm$ 0.37 & 41.13 $\pm$ 0.28 & 66 & 33.4 \\
 &  & cut (1.08) & 40 & 20.71 $\pm$ 1.53 & 23.39 $\pm$ 1.29 & -- & -- \\
 & 16 & 0 & 58 & 1.00 $\pm$ 0.00 & 1.00 $\pm$ 0.00 & 1 & 60.4 \\
 &  & 1 & 3 & 15.41 $\pm$ 0.13 & 25.08 $\pm$ 0.08 & 29 & 59.5 \\
 &  & 2 & 1 & 75.97 $\pm$ 0.00 & 72.26 $\pm$ 0.00 & 120 & 61.4 \\
 &  & cut (1.13) & 57 & 35.72 $\pm$ 2.39 & 39.80 $\pm$ 1.99 & -- & -- \\
Network design & 7 & 0 & 43 & 1.00 $\pm$ 0.00 & 1.00 $\pm$ 0.00 & 1 & 11.0 \\
 &  & 1 & 43 & 6.53 $\pm$ 0.02 & 8.62 $\pm$ 0.04 & 11 & 11.0 \\
 &  & 2 & 1 & 16.60 $\pm$ 0.00 & 14.38 $\pm$ 0.00 & 21 & 11.0 \\
 &  & cut (1.60) & 9 & 12.27 $\pm$ 0.79 & 11.70 $\pm$ 0.49 & -- & -- \\
 & 12 & 0 & 30 & 1.00 $\pm$ 0.00 & 1.00 $\pm$ 0.00 & 1 & 26.0 \\
 &  & 1 & 54 & 10.26 $\pm$ 0.76 & 15.15 $\pm$ 1.13 & 21 & 26.0 \\
 &  & 2 & 2 & 43.83 $\pm$ 0.00 & 41.30 $\pm$ 0.00 & 66 & 33.8 \\
 &  & cut (1.55) & 33 & 26.68 $\pm$ 2.91 & 25.29 $\pm$ 2.65 & -- & -- \\
 & 14 & 0 & 30 & 1.00 $\pm$ 0.00 & 1.00 $\pm$ 0.00 & 1 & 30.3 \\
 &  & 1 & 48 & 10.61 $\pm$ 0.95 & 15.88 $\pm$ 1.43 & 25 & 29.8 \\
 &  & 2 & 10 & 51.41 $\pm$ 3.17 & 45.56 $\pm$ 4.65 & 91 & 35.4 \\
 &  & cut (1.53) & 32 & 32.27 $\pm$ 2.99 & 29.88 $\pm$ 2.86 & -- & -- \\
Facility location & 7 & 0 & 39 & 1.00 $\pm$ 0.00 & 1.00 $\pm$ 0.00 & 1 & 11.0 \\
 &  & 1 & 10 & 6.53 $\pm$ 0.04 & 8.64 $\pm$ 0.05 & 11 & 11.0 \\
 &  & 2 & 14 & 16.54 $\pm$ 0.04 & 14.36 $\pm$ 0.10 & 21 & 11.0 \\
 &  & cut (1.49) & 33 & 11.35 $\pm$ 0.32 & 11.23 $\pm$ 0.26 & -- & -- \\
 & 12 & 0 & 22 & 1.00 $\pm$ 0.00 & 1.00 $\pm$ 0.00 & 1 & 24.3 \\
 &  & 1 & 13 & 11.52 $\pm$ 0.06 & 16.33 $\pm$ 0.05 & 21 & 24.4 \\
 &  & 2 & 8 & 38.20 $\pm$ 0.14 & 32.58 $\pm$ 0.18 & 66 & 24.1 \\
 &  & cut (1.19) & 41 & 18.64 $\pm$ 1.22 & 19.36 $\pm$ 0.78 & -- & -- \\
 & 16 & 0 & 21 & 1.00 $\pm$ 0.00 & 1.00 $\pm$ 0.00 & 1 & 43.3 \\
 &  & 1 & 18 & 14.34 $\pm$ 0.58 & 22.03 $\pm$ 0.89 & 29 & 43.3 \\
 &  & 2 & 11 & 62.25 $\pm$ 1.40 & 55.26 $\pm$ 2.29 & 120 & 42.5 \\
 &  & cut (1.26) & 46 & 31.90 $\pm$ 1.93 & 33.60 $\pm$ 1.81 & -- & -- \\
\bottomrule
\end{tabular}
\end{table}

\begin{table}[!htbp]
\caption{Rich models, $s=3$: predictions whose anchor has two families while the best set has three. Share of the orders of the families in which the staged order $O_1$ alone reaches a best set, number of evaluations of JointDW as mean $\pm$ standard error, and mean cost of the covering order on the same instances.}
\label{tab:app-larger}
\centering
\footnotesize
\setlength{\tabcolsep}{6pt}
\begin{tabular}{@{}lccccc@{}}
\toprule
Class & $k$ & predictions & share of orders & JointDW & covering \\
\midrule
Lot sizing & 9 & 32 & 0.89 & 11.01 $\pm$ 0.75 & 36.0 \\
 & 12 & 43 & 0.93 & 18.70 $\pm$ 3.62 & 106.1 \\
 & 16 & 24 & 0.96 & 14.66 $\pm$ 7.35 & 160.9 \\
Network design & 12 & 37 & 0.98 & 10.91 $\pm$ 2.10 & 56.2 \\
Facility location & 12 & 19 & 0.45 & 64.07 $\pm$ 2.15 & 91.5 \\
 & 16 & 38 & 0.67 & 77.29 $\pm$ 15.05 & 160.3 \\
\bottomrule
\end{tabular}
\end{table}

\paragraph{Cost of a prediction.} \Cref{tab:app-cost} gives, for JointDW interleaved with the five-dual order and $s=2$, the latency and cost of one prediction and the evaluations and net time saved against the covering order and against the five-dual order. It also gives the break-even time of one evaluation against the covering order. The time of one evaluation is the mean over the sets that the orders evaluate: a GCG evaluation in the census setting on the templates and the rich models, the census evaluation on MIPLIB-NL, and a branch-and-price run on the benchmark instances. On the benchmark instances, the net time uses each instance's own mean time, and the five-dual lookahead is charged its measured time, 2.4 seconds on average; on the other collections this time is not recorded and not charged. The latency is charged once per instance, also where instances share a prompt. On the benchmark instances, with an evaluation of 494 seconds on average, GPT-6.1 Sol given every coefficient saves $430\pm81$ seconds net against the covering order, and the break-even time is about 50 seconds.

\begin{table}[!htbp]
\caption{Cost of a prediction for JointDW interleaved with the five-dual lookahead order, $s=2$: latency (s) and cost (USD) of one prediction at list prices, evaluations saved against the covering order and against the five-dual order, net time saved (s) against both, and break-even time of one evaluation (s) against the covering order. Mean $\pm$ standard error of the two-stage bootstrap, with an asterisk when the 95 percent interval excludes zero.}
\label{tab:app-cost}
\centering
\scriptsize
\setlength{\tabcolsep}{2.5pt}
\resizebox{\textwidth}{!}{%
\begin{tabular}{@{}lccccccc@{}}
\toprule
& & & \multicolumn{2}{c}{evaluations saved against} & \multicolumn{2}{c}{net time saved against} & \\
\cmidrule(lr){4-5}\cmidrule(lr){6-7}
Configuration & latency & cost & covering & five duals & covering & five duals & break-even \\
\midrule
\multicolumn{8}{@{}l}{\emph{Templates; one evaluation 0.094 s}} \\
GPT-6.1 Sol, text & 34.9 & 0.011 & 1.53 $\pm$ 0.48$^{*}$ & $-$0.19 $\pm$ 0.11 & $-$34.8 $\pm$ 2.3$^{*}$ & $-$34.9 $\pm$ 2.3$^{*}$ & 22.9 \\
Claude Sonnet 5.5, text & 157.2 & 0.212 & 1.30 $\pm$ 0.52$^{*}$ & $-$0.41 $\pm$ 0.14$^{*}$ & $-$157.0 $\pm$ 18.4$^{*}$ & $-$157.2 $\pm$ 18.4$^{*}$ & 120.5 \\
\multicolumn{8}{@{}l}{\emph{MIPLIB-NL; one evaluation 43.3 s}} \\
GPT-6.1 Sol, text & 35.8 & 0.012 & 1.74 $\pm$ 0.43$^{*}$ & $-$0.17 $\pm$ 0.17 & 39.6 $\pm$ 18.0$^{*}$ & $-$42.9 $\pm$ 7.5$^{*}$ & 20.5 \\
Claude Sonnet 5.5, text & 324.5 & 0.207 & 1.52 $\pm$ 0.43$^{*}$ & $-$0.39 $\pm$ 0.14$^{*}$ & $-$258.7 $\pm$ 88.7$^{*}$ & $-$341.3 $\pm$ 83.4$^{*}$ & 213.4 \\
\multicolumn{8}{@{}l}{\emph{Rich models with 7 to 9 families; one evaluation 0.102 s}} \\
GPT-6.1 Sol, text & 50.4 & 0.019 & 11.13 $\pm$ 1.56$^{*}$ & $-$0.70 $\pm$ 0.06$^{*}$ & $-$49.3 $\pm$ 1.9$^{*}$ & $-$50.5 $\pm$ 1.9$^{*}$ & 4.5 \\
GPT-6.1 Sol, coefficients & 97.8 & 0.039 & 12.54 $\pm$ 0.75$^{*}$ & 0.71 $\pm$ 0.71 & $-$96.5 $\pm$ 4.7$^{*}$ & $-$97.7 $\pm$ 4.8$^{*}$ & 7.8 \\
Claude Sonnet 5.5, text & 1454.8 & 0.205 & 11.40 $\pm$ 1.41$^{*}$ & $-$0.43 $\pm$ 0.23 & $-$1453.7 $\pm$ 359.4$^{*}$ & $-$1454.9 $\pm$ 359.5$^{*}$ & 127.6 \\
Claude Sonnet 5.5, coefficients & 672.7 & 0.209 & 11.65 $\pm$ 0.94$^{*}$ & $-$0.18 $\pm$ 0.32 & $-$671.5 $\pm$ 92.5$^{*}$ & $-$672.7 $\pm$ 92.5$^{*}$ & 57.8 \\
\multicolumn{8}{@{}l}{\emph{Rich models with 12 to 16 families; one evaluation 0.106 s}} \\
GPT-6.1 Sol, text & 31.9 & 0.018 & 26.85 $\pm$ 3.66$^{*}$ & 0.09 $\pm$ 0.37 & $-$29.0 $\pm$ 0.7$^{*}$ & $-$31.9 $\pm$ 0.7$^{*}$ & 1.2 \\
GPT-6.1 Sol, coefficients & 69.6 & 0.036 & 31.93 $\pm$ 2.10$^{*}$ & 5.16 $\pm$ 1.75$^{*}$ & $-$66.2 $\pm$ 1.8$^{*}$ & $-$69.0 $\pm$ 1.7$^{*}$ & 2.2 \\
Claude Sonnet 5.5, text & 475.5 & 0.206 & 27.20 $\pm$ 3.62$^{*}$ & 0.43 $\pm$ 0.51 & $-$472.6 $\pm$ 102.0$^{*}$ & $-$475.4 $\pm$ 102.2$^{*}$ & 17.5 \\
Claude Sonnet 5.5, coefficients & 205.1 & 0.211 & 27.43 $\pm$ 2.68$^{*}$ & 0.67 $\pm$ 0.62 & $-$202.2 $\pm$ 6.7$^{*}$ & $-$205.0 $\pm$ 6.6$^{*}$ & 7.5 \\
\multicolumn{8}{@{}l}{\emph{Benchmark instances; one evaluation 494.5 s}} \\
GPT-6.1 Sol, text & 34.9 & 0.013 & 0.75 $\pm$ 0.18$^{*}$ & 0.44 $\pm$ 0.27$^{*}$ & 452.5 $\pm$ 305.5 & $-$21.5 $\pm$ 12.3 & 49.5 \\
GPT-6.1 Sol, coefficients & 32.6 & 0.046 & 0.65 $\pm$ 0.07$^{*}$ & 0.33 $\pm$ 0.16$^{*}$ & 429.7 $\pm$ 80.9$^{*}$ & $-$44.3 $\pm$ 7.9$^{*}$ & 53.7 \\
Claude Sonnet 5.5, text & 303.4 & 0.206 & 0.74 $\pm$ 0.15$^{*}$ & 0.43 $\pm$ 0.36 & 181.2 $\pm$ 383.7 & $-$292.8 $\pm$ 151.4$^{*}$ & 410.8 \\
Claude Sonnet 5.5, coefficients & 327.2 & 0.356 & 0.30 $\pm$ 0.12$^{*}$ & $-$0.02 $\pm$ 0.11 & 106.1 $\pm$ 75.9 & $-$368.0 $\pm$ 39.5$^{*}$ & 1112.9 \\
\bottomrule
\end{tabular}}
\end{table}

\paragraph{Evaluation time and the number of families.} \Cref{tab:app-growth} gives the median time of one evaluation by the number of convexified families. It also gives the slope of the logarithm of the median time against the logarithm of the number of families, fitted per model with at least three sizes. The medians use the sets that finish within the time limit. On MIPLIB-NL, between 58 and 290 sets of each size reach the limit of 120 seconds, a share that grows with the size, so the medians leave out more of the slow large sets. On the benchmark instances, the root time of a pair is 0.98 times that of a single family at the median, and its pricing time 1.09 times. The total branch-and-price time falls with the number of families (\cref{tab:app-tradeoff}).

\begin{table}[!htbp]
\caption{Median time of one evaluation in seconds by the number of convexified families, over the sets that finish within the time limit, and the slope of the logarithm of the median time against the logarithm of the number of families: number of models with at least three sizes, median and quartiles of their slopes. Census time: the time of the census evaluation; GCG time: the time that GCG reports.}
\label{tab:app-growth}
\centering
\scriptsize
\setlength{\tabcolsep}{3pt}
\begin{tabular}{@{}lccccccccc@{}}
\toprule
& \multicolumn{5}{c}{families convexified} & \multicolumn{4}{c}{slope} \\
\cmidrule(lr){2-6}\cmidrule(l){7-10}
Collection & 1 & 2 & 3 & 4 & 5 & models & median & 25\% & 75\% \\
\midrule
MIPLIB-NL, census time & 1.53 & 4.24 & 16.9 & 18.3 & 17.8 & 41 & 1.65 & 0.66 & 2.26 \\
Templates, census time & 0.0900 & 0.0900 & 0.0890 & 0.0890 & 0.0930 & 4 & 0.02 & 0.02 & 0.03 \\
Templates, GCG time & 0.0145 & 0.0126 & 0.0134 & 0.0128 & 0.0133 & 4 & 0.22 & 0.15 & 0.29 \\
Rich models, census time & 0.0880 & 0.100 & 0.105 & -- & -- & 24 & 0.14 & 0.07 & 0.17 \\
Rich models, GCG time & 0.0158 & 0.0243 & 0.0291 & -- & -- & 24 & 0.50 & 0.33 & 0.62 \\
\bottomrule
\end{tabular}
\end{table}

\begin{table}[!htbp]
\caption{Paired differences in the number of evaluations: JointDW or its interleaving with the lookahead order minus the covering or the lookahead order, mean $\pm$ standard error. Negative values favor JointDW or the interleaved order, and an asterisk marks a difference whose 95 percent bootstrap interval excludes zero, with each cell tested separately.}
\label{tab:app-paired-search}
\centering
\footnotesize
\begin{tabular}{@{}lcccc@{}}
\toprule
& \multicolumn{2}{c}{JointDW minus} & \multicolumn{2}{c}{Interleaved minus} \\
\cmidrule(lr){2-3}\cmidrule(l){4-5}
Configuration & covering & lookahead & covering & lookahead \\
\midrule
\multicolumn{5}{@{}l}{\emph{Templates, $s=2$}} \\
Claude Sonnet 5.5 & $-$0.38 $\pm$ 0.19$^{*}$ & 0.47 $\pm$ 1.06 & $-$0.90 $\pm$ 0.84 & $-$0.04 $\pm$ 0.32 \\
GPT-6.1 Sol & $-$0.79 $\pm$ 0.10$^{*}$ & 0.07 $\pm$ 1.19 & $-$1.65 $\pm$ 0.51$^{*}$ & $-$0.79 $\pm$ 0.82 \\
Grok 4.20 (Reasoning) & $-$0.38 $\pm$ 0.15$^{*}$ & 0.48 $\pm$ 1.10 & $-$0.98 $\pm$ 0.77 & $-$0.12 $\pm$ 0.38 \\
Grok 4.20 (Non-Reasoning) & 0.06 $\pm$ 0.20 & 0.92 $\pm$ 1.06 & $-$0.45 $\pm$ 0.94 & 0.41 $\pm$ 0.23$^{*}$ \\
DeepSeek-V4-Pro (Non-Thinking) & 0.28 $\pm$ 0.11$^{*}$ & 1.14 $\pm$ 1.06 & $-$0.31 $\pm$ 0.93 & 0.55 $\pm$ 0.24$^{*}$ \\
\multicolumn{5}{@{}l}{\emph{Templates, $s=3$}} \\
Claude Sonnet 5.5 & $-$0.22 $\pm$ 0.21 & 0.19 $\pm$ 0.81 & $-$0.40 $\pm$ 0.57 & 0.01 $\pm$ 0.30 \\
GPT-6.1 Sol & $-$0.60 $\pm$ 0.18$^{*}$ & $-$0.20 $\pm$ 0.98 & $-$1.14 $\pm$ 0.22$^{*}$ & $-$0.74 $\pm$ 0.76 \\
Grok 4.20 (Reasoning) & $-$0.22 $\pm$ 0.19 & 0.18 $\pm$ 0.85 & $-$0.50 $\pm$ 0.50 & $-$0.09 $\pm$ 0.36 \\
Grok 4.20 (Non-Reasoning) & 0.36 $\pm$ 0.23 & 0.76 $\pm$ 0.88 & 0.03 $\pm$ 0.67 & 0.44 $\pm$ 0.22$^{*}$ \\
DeepSeek-V4-Pro (Non-Thinking) & 0.36 $\pm$ 0.09$^{*}$ & 0.77 $\pm$ 0.81 & 0.13 $\pm$ 0.65 & 0.53 $\pm$ 0.21$^{*}$ \\
\multicolumn{5}{@{}l}{\emph{Templates, $s=k-1$}} \\
Claude Sonnet 5.5 & $-$0.11 $\pm$ 0.23 & $-$0.21 $\pm$ 0.64 & 0.10 $\pm$ 0.44 & 0.01 $\pm$ 0.30 \\
GPT-6.1 Sol & $-$0.66 $\pm$ 0.15$^{*}$ & $-$0.76 $\pm$ 0.76 & $-$0.72 $\pm$ 0.11$^{*}$ & $-$0.82 $\pm$ 0.74 \\
Grok 4.20 (Reasoning) & $-$0.23 $\pm$ 0.19 & $-$0.33 $\pm$ 0.63 & $-$0.02 $\pm$ 0.37 & $-$0.12 $\pm$ 0.35 \\
Grok 4.20 (Non-Reasoning) & 0.33 $\pm$ 0.22 & 0.24 $\pm$ 0.66 & 0.53 $\pm$ 0.52 & 0.44 $\pm$ 0.22$^{*}$ \\
DeepSeek-V4-Pro (Non-Thinking) & 0.32 $\pm$ 0.09$^{*}$ & 0.22 $\pm$ 0.64 & 0.55 $\pm$ 0.49 & 0.46 $\pm$ 0.20$^{*}$ \\
\multicolumn{5}{@{}l}{\emph{MIPLIB-NL, $s=2$}} \\
Claude Sonnet 5.5 & $-$0.88 $\pm$ 0.38$^{*}$ & 0.03 $\pm$ 0.69 & $-$1.09 $\pm$ 0.34$^{*}$ & $-$0.19 $\pm$ 0.43 \\
GPT-6.1 Sol & $-$1.29 $\pm$ 0.49$^{*}$ & $-$0.39 $\pm$ 0.89 & $-$1.74 $\pm$ 0.43$^{*}$ & $-$0.83 $\pm$ 0.81 \\
Grok 4.20 (Reasoning) & $-$0.66 $\pm$ 0.33$^{*}$ & 0.25 $\pm$ 0.60 & $-$0.93 $\pm$ 0.45$^{*}$ & $-$0.03 $\pm$ 0.34 \\
Grok 4.20 (Non-Reasoning) & $-$0.32 $\pm$ 0.34 & 0.59 $\pm$ 0.82 & $-$1.02 $\pm$ 0.30$^{*}$ & $-$0.11 $\pm$ 0.47 \\
DeepSeek-V4-Pro (Non-Thinking) & $-$0.57 $\pm$ 0.24$^{*}$ & 0.33 $\pm$ 0.69 & $-$0.93 $\pm$ 0.33$^{*}$ & $-$0.02 $\pm$ 0.39 \\
\multicolumn{5}{@{}l}{\emph{MIPLIB-NL, $s=3$}} \\
Claude Sonnet 5.5 & $-$0.50 $\pm$ 0.21$^{*}$ & 0.14 $\pm$ 0.72 & $-$0.84 $\pm$ 0.30$^{*}$ & $-$0.14 $\pm$ 0.42 \\
GPT-6.1 Sol & $-$0.85 $\pm$ 0.40$^{*}$ & $-$0.20 $\pm$ 1.00 & $-$1.55 $\pm$ 0.37$^{*}$ & $-$0.79 $\pm$ 0.81 \\
Grok 4.20 (Reasoning) & $-$0.52 $\pm$ 0.21$^{*}$ & 0.13 $\pm$ 0.77 & $-$0.88 $\pm$ 0.32$^{*}$ & $-$0.16 $\pm$ 0.43 \\
Grok 4.20 (Non-Reasoning) & $-$0.09 $\pm$ 0.35 & 0.56 $\pm$ 0.95 & $-$0.95 $\pm$ 0.27$^{*}$ & $-$0.22 $\pm$ 0.56 \\
DeepSeek-V4-Pro (Non-Thinking) & $-$0.32 $\pm$ 0.20 & 0.33 $\pm$ 0.78 & $-$0.70 $\pm$ 0.31$^{*}$ & 0.01 $\pm$ 0.40 \\
\multicolumn{5}{@{}l}{\emph{MIPLIB-NL, $s=k-1$}} \\
Claude Sonnet 5.5 & $-$0.08 $\pm$ 0.18 & $-$0.38 $\pm$ 1.03 & $-$0.54 $\pm$ 0.11$^{*}$ & $-$0.37 $\pm$ 0.61 \\
GPT-6.1 Sol & $-$0.28 $\pm$ 0.24 & $-$0.59 $\pm$ 1.09 & $-$0.81 $\pm$ 0.07$^{*}$ & $-$0.97 $\pm$ 0.97 \\
Grok 4.20 (Reasoning) & $-$0.15 $\pm$ 0.23 & $-$0.47 $\pm$ 1.09 & $-$0.66 $\pm$ 0.11$^{*}$ & $-$0.46 $\pm$ 0.64 \\
Grok 4.20 (Non-Reasoning) & $-$0.01 $\pm$ 0.21 & $-$0.34 $\pm$ 1.10 & $-$0.52 $\pm$ 0.15$^{*}$ & 0.04 $\pm$ 0.38 \\
DeepSeek-V4-Pro (Non-Thinking) & 0.04 $\pm$ 0.15 & $-$0.29 $\pm$ 1.08 & $-$0.43 $\pm$ 0.11$^{*}$ & 0.15 $\pm$ 0.31 \\
\bottomrule
\end{tabular}
\end{table}

\paragraph{First sets of the language models and of the lookahead.} \Cref{tab:app-first} compares, on the same instances, the first set of a language model with the first set of the five-dual lookahead order. On MIPLIB-NL, at least one of the two names a true group first on 0.55 to 0.73 of the models, against 0.53 for the lookahead alone. On the rich models with 12 to 16 families, the lookahead is right more often than either language model given the text, and at least one of the two is right about a tenth more often than the lookahead. Given every coefficient, GPT-6.1 Sol is right on almost every rich model on which the lookahead is right. In the interleaved order, the five-dual lookahead order takes every second turn and the staged order every fourth. Over all collections, configurations, presentations, and budgets, the interleaved order needs between 0.01 and 2.5 evaluations more than the smaller, for each sample, of the cost of the staged order alone and of the five-dual lookahead order.

\begin{table}[!htbp]
\caption{First set of a language model and first set of the five-dual lookahead order on the instances with a true joint support and, for the benchmark instances, with a best pair: hit@1 of the language model, share of instances on which only the language model, only the lookahead, and at least one of the two names a true group first, mean $\pm$ standard error. The outcome of a sample is averaged over the samples of an instance and then over the instances. On the nine structured templates, the five-dual lookahead names a true group first on every instance.}
\label{tab:app-first}
\centering
\footnotesize
\setlength{\tabcolsep}{5pt}
\begin{tabular}{@{}lcccc@{}}
\toprule
Configuration & hit@1 & only language model & only lookahead & at least one \\
\midrule
\multicolumn{5}{@{}l}{\emph{MIPLIB-NL, 17 models; five-dual lookahead 0.53 $\pm$ 0.12}} \\
Claude Sonnet 5.5 & 0.57 $\pm$ 0.10 & 0.20 $\pm$ 0.09 & 0.16 $\pm$ 0.07 & 0.73 $\pm$ 0.11 \\
GPT-6.1 Sol & 0.53 $\pm$ 0.12 & 0.12 $\pm$ 0.08 & 0.12 $\pm$ 0.08 & 0.65 $\pm$ 0.12 \\
Grok 4.20 (Reasoning) & 0.55 $\pm$ 0.11 & 0.14 $\pm$ 0.07 & 0.12 $\pm$ 0.08 & 0.67 $\pm$ 0.11 \\
Grok 4.20 (Non-Reasoning) & 0.29 $\pm$ 0.10 & 0.06 $\pm$ 0.04 & 0.29 $\pm$ 0.11 & 0.59 $\pm$ 0.11 \\
DeepSeek-V4-Pro (Non-Thinking) & 0.27 $\pm$ 0.08 & 0.02 $\pm$ 0.02 & 0.27 $\pm$ 0.08 & 0.55 $\pm$ 0.12 \\
\multicolumn{5}{@{}l}{\emph{Rich models with 7 to 9 families, 24 models; five-dual lookahead 0.04 $\pm$ 0.05}} \\
GPT-6.1 Sol, text & 0.22 $\pm$ 0.05 & 0.21 $\pm$ 0.06 & 0.03 $\pm$ 0.03 & 0.25 $\pm$ 0.07 \\
GPT-6.1 Sol, coefficients & 0.99 $\pm$ 0.01 & 0.94 $\pm$ 0.04 & 0.00 $\pm$ 0.00 & 0.99 $\pm$ 0.01 \\
Claude Sonnet 5.5, text & 0.32 $\pm$ 0.06 & 0.31 $\pm$ 0.06 & 0.03 $\pm$ 0.03 & 0.35 $\pm$ 0.07 \\
Claude Sonnet 5.5, coefficients & 0.47 $\pm$ 0.08 & 0.47 $\pm$ 0.08 & 0.04 $\pm$ 0.04 & 0.51 $\pm$ 0.08 \\
\multicolumn{5}{@{}l}{\emph{Rich models with 12 to 16 families, 60 models; five-dual lookahead 0.32 $\pm$ 0.07}} \\
GPT-6.1 Sol, text & 0.07 $\pm$ 0.02 & 0.06 $\pm$ 0.02 & 0.32 $\pm$ 0.07 & 0.39 $\pm$ 0.07 \\
GPT-6.1 Sol, coefficients & 0.99 $\pm$ 0.01 & 0.67 $\pm$ 0.06 & 0.01 $\pm$ 0.01 & 0.99 $\pm$ 0.01 \\
Claude Sonnet 5.5, text & 0.13 $\pm$ 0.04 & 0.12 $\pm$ 0.04 & 0.31 $\pm$ 0.07 & 0.45 $\pm$ 0.06 \\
Claude Sonnet 5.5, coefficients & 0.27 $\pm$ 0.04 & 0.24 $\pm$ 0.04 & 0.30 $\pm$ 0.06 & 0.57 $\pm$ 0.05 \\
\multicolumn{5}{@{}l}{\emph{Benchmark, 59 instances; five-dual lookahead 0.76 $\pm$ 0.13}} \\
GPT-6.1 Sol, text & 0.89 $\pm$ 0.08 & 0.17 $\pm$ 0.11 & 0.05 $\pm$ 0.04 & 0.93 $\pm$ 0.05 \\
GPT-6.1 Sol, coefficients & 0.78 $\pm$ 0.04 & 0.13 $\pm$ 0.03 & 0.11 $\pm$ 0.04 & 0.89 $\pm$ 0.03 \\
Claude Sonnet 5.5, text & 0.81 $\pm$ 0.09 & 0.19 $\pm$ 0.11 & 0.14 $\pm$ 0.07 & 0.95 $\pm$ 0.04 \\
Claude Sonnet 5.5, coefficients & 0.56 $\pm$ 0.05 & 0.04 $\pm$ 0.02 & 0.24 $\pm$ 0.04 & 0.80 $\pm$ 0.05 \\
\bottomrule
\end{tabular}
\end{table}

\paragraph{Hidden-group programs with every coefficient.} The text of a hidden-group template gives every family the same description, and in the original presentation no configuration names the hidden group first in any answer. A set of $w$ families drawn at random is the hidden group with probability $1/\binom{k}{w}$, which is 0.11 on average over the four templates. With the coefficient variant of the prompt (\cref{app:prompt}), which has 1318 to 1550 tokens on these programs, Claude Sonnet 5.5 names the hidden group first in all 12 answers, three per template. GPT-6.1 Sol does so in all nine answers on HG a, HG b, and HG c. On HG d, each of its three requests returns an answer without reasoning in all three attempts. Such an answer is not accepted from a configuration with reasoning and counts as a failed prediction, which gives a hit rate of $0.75\pm0.21$ over the four templates. The 21 answers describe the group as an odd cycle or as a triangle with its clique inequality. The hidden group can therefore be read from the constraint matrix, although bound evaluations find it only at the cost stated in \cref{thm:cover}.

\FloatBarrier
\end{document}